%% file: neurips.tex
\documentclass{article}

\input{neurips_preamble}

\input{preamble}
\input{macros}

\title{Counterfactual reasoning: \\ an analysis of in-context emergence}
\author{%
Moritz Miller$^{12}$ \thanks{Correspondence to \texttt{moritz.miller@tuebingen.mpg.de}} \quad Bernhard Schölkopf$^{12}$\quad Siyuan Guo$^{13}$ \\
$^1$Max Planck Institute for Intelligent Systems \quad $^2$ETH Zurich \quad
$^3$ University of Cambridge\\
}

\usepackage{float}
\usepackage[textsize=tiny,textwidth=1.3in]{todonotes}
\begin{document}
\maketitle

\input{chapters/paper-chapters/abstract}
\input{chapters/introduction}
\input{chapters/background}
\input{chapters/methodology}
\input{chapters/paper-chapters/experiments}
\input{chapters/paper-chapters/cyclic}
\input{chapters/discussion}
\input{chapters/conclusion}
\input{chapters/paper-chapters/acknowledgements}

\bibliography{bibliography}

\clearpage 
\input{chapters/checklist}

\appendix

\input{chapters/appendix-posterior-predictive}
\input{chapters/appendix-proofs}
\input{chapters/paper-chapters/appendix-training-details}
\input{chapters/appendix-model-details}
\input{chapters/paper-chapters/appendix-diversity}
\input{chapters/paper-chapters/appendix-cyclic}
\input{chapters/appendix-natural-language}


\end{document}

%% file: neurips_preamble.tex
\usepackage[english]{babel}

\usepackage[final]{neurips_2025}

\usepackage[utf8]{inputenc} 
\usepackage[T1]{fontenc}    
\usepackage{url}            
\usepackage{booktabs}       
\usepackage{amsfonts}       
\usepackage{nicefrac}       
\usepackage{microtype}      

%% file: preamble.tex
\usepackage{amsmath}
\usepackage{graphicx}
\usepackage{hyperref}   
\usepackage{enumitem}
\usepackage{amssymb}
\usepackage{amsthm}
\usepackage{physics}
\usepackage{caption}
\usepackage{subcaption}
\usepackage{tabularx} 
\usepackage{tabularray}
\usepackage{array}    
\usepackage{enumitem} 
\usepackage{wrapfig}
\usepackage{xcolor}         
\usepackage{skak}
\usepackage[inkscapelatex=false]{svg}

\definecolor{ourroyal}{HTML}{0073E6}
\definecolor{ourmagenta}{HTML}{B51963}
\definecolor{ouryellow}{HTML}{F2C14E}
\definecolor{ournavy}{HTML}{011638}
\definecolor{ourgreen}{HTML}{5FAD56}

\newtheorem{theorem}{Theorem}

\theoremstyle{definition}
\newtheorem{definition}{Definition} 
\newtheorem{lemma}{Lemma}           
\newtheorem{proposition}{Proposition}

\newtheorem*{theorem*}{Theorem}

\theoremstyle{remark}

\usepackage{color}
\usepackage{tikz}
\usepackage{titlesec}
\usepackage{pdfpages}
\usepackage[nottoc]{tocbibind}

\usetikzlibrary{shapes,decorations,arrows,calc,arrows.meta,fit,positioning}
\tikzset{
    -Latex,auto,node distance =1 cm and 1 cm,semithick,
    state/.style ={ellipse, draw, minimum width = 0.7 cm},
    point/.style = {circle, draw, inner sep=0.04cm,fill,node contents={}},
    bidirected/.style={Latex-Latex,dashed},
    el/.style = {inner sep=2pt, align=left, sloped}
}

\usepackage{natbib}
\def\ind{\perp\!\!\!\perp}

%% file: macros.tex
\newcommand{\mymacro}[1]{{#1}}

\newcommand{\inv}[1]{{\mymacro{#1^{-1}}}}

\newcommand{\probmeasure}{{\mymacro{ \mathbb{P}}}}

\newcommand{\MSE}{{\mymacro{ \mathrm{MSE}}}}

\newcommand{\ESS}{{\mymacro{ \mathrm{Ess}}}}
\newcommand{\comp}{\mathrm{comp}}

\newcommand{\R}{{\mymacro{ \mathbb{R}}}}

\newcommand{\RE}{{\mymacro{ \R^E}}}

\newcommand{\pbetadot}{p_\beta}
\newcommand{\Beta}{\mathrm{B}}

\newcommand{\XCF}{X^\text{CF}}
\newcommand{\YCF}{Y^\text{CF}}
\newcommand{\xCF}{x^\text{CF}}
\newcommand{\yCF}{y^\text{CF}}
\newcommand{\yCFbhat}{y_{\hat{\beta}}^\text{CF}}

\newcommand{\yCFhat}{\widehat{y^\text{CF}}}
\newcommand{\XiCFdot}[1]{X_{#1}^\text{CF}}
\newcommand{\YiCFdot}[1]{Y_{#1}^\text{CF}}
\newcommand{\mXiCFdot}[1]{\mX_{#1}^\text{CF}}
\newcommand{\mYiCFdot}[1]{\mY_{#1}^\text{CF}}

\newcommand{\softmaxcausal}{\softmax_*}

\newcommand{\outputclass}{\mathcal{Y}}
\newcommand{\noiseclass}{\mathcal{U}}
\newcommand{\adjR}{\text{adjusted }\mathrm{R}^2}
\newcommand{\Pa}{\mathrm{Pa}}

\newcommand{\ubar}[1]{\text{\b{$#1$}}}
\newcommand{\xlower}{\ubar{x}}
\newcommand{\xupper}{\bar{x}}
\newcommand{\ylower}{\ubar{y}}
\newcommand{\yupper}{\bar{y}}
\newcommand{\betaylwr}{\beta \ylower}
\newcommand{\betayupr}{\beta \yupper}
\newcommand{\logxtx}{\log \left( \frac{x_t}{x_0}\right)}
\newcommand{\logyty}{\log \left( \frac{y_t}{y_0}\right)}
\newcommand{\oneovert}{\frac{1}{t}}
\newcommand{\oneoverT}{\frac{1}{T}}
\newcommand{\logxuprx}{\log \left( \frac{\xupper}{x_0}\right)}
\newcommand{\logxlwrx}{\log \left( \frac{\xlower}{x_0}\right)}
\newcommand{\logyupry}{\log \left( \frac{\yupper}{y_0}\right)}
\newcommand{\logylwry}{\log \left( \frac{\ylower}{y_0}\right)}
\newcommand{\loguprxlwrx}{\log \left( \frac{\xupper}{\xlower}\right)}
\newcommand{\loguprylwry}{\log \left( \frac{\yupper}{\ylower}\right)}
\newcommand{\deltaxlwr}{\delta \xlower}
\newcommand{\deltaxupr}{\delta \xupper}

\newcommand{\alphalwr}{\ubar{\alpha}}
\newcommand{\alphaupr}{\bar{\alpha}}
\newcommand{\betalwr}{\ubar{\beta}}
\newcommand{\gammalwr}{\ubar{\gamma}}
\newcommand{\gammaupr}{\bar{\gamma}}
\newcommand{\deltalwr}{\ubar{\delta}}
\newcommand{\alphalwrbf}{\ubar{\rvalpha}}
\newcommand{\alphauprbf}{\bar{\rvalpha}}
\newcommand{\betalwrbf}{\ubar{\rvbeta}}
\newcommand{\gammalwrbf}{\ubar{\rvgamma}}
\newcommand{\gammauprbf}{\bar{\rvgamma}}
\newcommand{\deltalwrbf}{\ubar{\rvdelta}}

\newcommand{\Exp}{\mathrm{Exp}}
\newcommand{\Uniform}{\mathcal{U}}
\newcommand{\Normal}{\mathcal{N}}
\newcommand{\Betadist}{\mathrm{Beta}}

\newcommand\smalldots{\hbox to 1em{.\hss.\hss.}}

\newcommand{\yiCFdot}[1]{\mathbf{y}_{#1}^\text{CF}}

\newcommand{\YZCF}{Y_{Z}^\text{CF}}

\newcommand{\yzCF}{y_{z}^\text{CF}}

\newcommand{\mXCF}{\mX^\text{CF}}

\newcommand{\xz}{x_{z}}

\newcommand{\yz}{y_{z}}

\newcommand{\xiCFdot}[1]{\mathbf{x}_{#1}^\text{CF}}
\newcommand{\xzeroCF}{x_0^\text{CF}}
\newcommand{\yzeroCF}{y_0^\text{CF}}

\newcommand{\deltaYCF}[1]{\delta(\YCF = #1)}

\newcommand{\deltaYZCF}[1]{\delta(\YZCF = #1)}

\newcommand{\contextOBS}{(x_{1},y_{1},...,x_{n},y_{n})}

\newcommand{\Wunem}{W_U}
\newcommand{\Wembd}{W_E}

\newcommand{\tp}[1]{{#1}^\top}

\newcommand{\pos}{\text{pos}}

\newcommand{\Qweight}{W_Q}
\newcommand{\Kweight}{W_K}
\newcommand{\Vweight}{W_V}
\newcommand{\Oweight}{W_O}
\newcommand{\MLP}{\mathrm{MLP}}
\newcommand{\Concat}{\mathrm{Concat}}

\newcommand{\one}[1]{{\mymacro{\mathbf{1}_{#1}}}}

\newcommand{\softmax}{{\mymacro{ \mathrm{softmax}}}}

\newcommand{\negterm}[1]{{\mymacro{ {\raise.17ex\hbox{$\scriptstyle\sim$}} #1}}}

\newcommand{\ignore}[1]{}
\newcommand{\expandLater}[1]{}

\def\1{\mathbf{1}}

\def\rvtheta{{{\mymacro{ \boldsymbol{\theta}}}}}
\def\rvalpha{{{\mymacro{ \boldsymbol{\alpha}}}}}
\def\rvgamma{{{\mymacro{ \boldsymbol{\gamma}}}}}
\def\rvdelta{{{\mymacro{ \boldsymbol{\delta}}}}}

\def\rvbeta{{{\mymacro{ \boldsymbol{\beta}}}}}

\def\mA{{{\mymacro{ \mathbf{A}}}}}

\def\mE{{{\mymacro{ \mathbf{E}}}}}

\def\mI{{{\mymacro{ \mathbf{I}}}}}

\def\mM{{{\mymacro{ \mathbf{M}}}}}

\def\mO{{{\mymacro{ \mathbf{O}}}}}

\def\mR{{{\mymacro{ \mathbf{R}}}}}

\def\mU{{{\mymacro{ \mathbf{U}}}}}

\def\mX{{{\mymacro{ \mathbf{X}}}}}
\def\mY{{{\mymacro{ \mathbf{Y}}}}}

\def\sU{{{\mymacro{ \mathcal{U}}}}}

\newcommand{\E}{{\mymacro{ \mathbb{E}}}}

\newcommand{\Var}{{\mymacro{ \mathrm{Var}}}}

\DeclareMathSymbol{\mlq}{\mathord}{operators}{``} 
\DeclareMathSymbol{\mrq}{\mathord}{operators}{`'} 


%% file: chapters/paper-chapters/abstract.tex
\begin{abstract}
Large-scale neural language models exhibit remarkable performance in in-context learning: the ability to learn and reason about the input context on the fly. This work studies in-context counterfactual reasoning in language models, that is, the ability to predict consequences of a hypothetical scenario. We focus on a well-defined, synthetic linear regression task that requires noise abduction. Accurate prediction is based on (1) inferring an unobserved latent concept and (2) copying contextual noise from factual observations. We show that language models are capable of counterfactual reasoning. Further, we enhance existing identifiability results and reduce counterfactual reasoning for a broad class of functions to a transformation on in-context observations. In Transformers, we find that self-attention, model depth and pre-training data diversity drive performance. Moreover, we provide mechanistic evidence that the latent concept is linearly represented in the residual stream and we introduce designated \textit{noise abduction heads} central to performing counterfactual reasoning. Lastly, our findings extend to counterfactual reasoning under SDE dynamics and reflect that Transformers can perform noise abduction on sequential data, providing preliminary evidence on the potential for counterfactual story generation. Our code is available under \texttt{https://github.com/mrtzmllr/iccr}. \looseness=-1
\end{abstract}

%% file: chapters/introduction.tex
\section{Introduction}
\label{chp:introduction}
\begin{center}
    Thinking is acting in an imagined space.  -- Konrad Lorenz~\citep{lorenz1973spiegel}
\end{center}

Large language models demonstrate remarkable capability in task-agnostic, few-shot performance, such as in-context learning and algorithmic reasoning~\citep{brown2020fewshot}. Despite their success, one of the grand challenges of artificial general intelligence remains the ability to unearth novel knowledge. This requires principled reasoning over factual observations instead of inventing information when uncertain, a phenomenon termed hallucination \citep{achiam2023gpt}. \looseness=-1

\begin{figure}[ht]
    \centering
    \includegraphics[width = \linewidth]{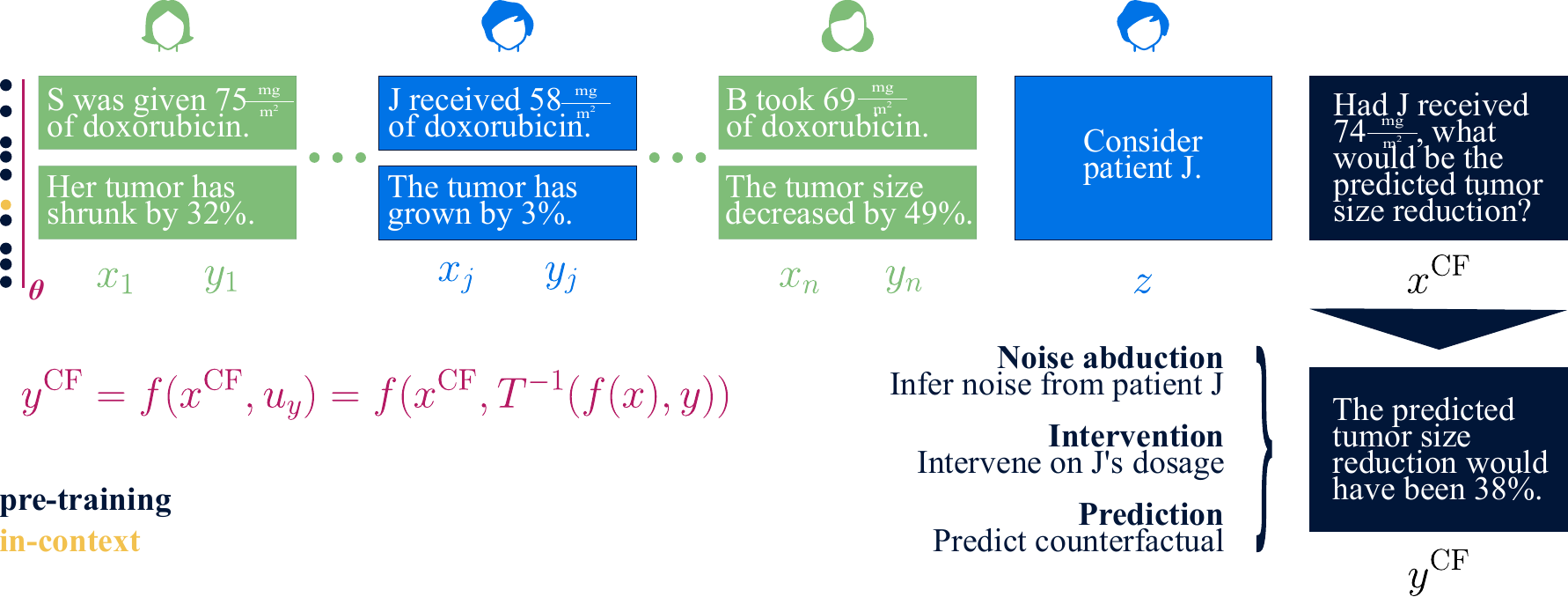}
    \caption{\textbf{In-context counterfactual reasoning.} Training on a corpus of sequences that come from a mixture of distributions (each \textcolor{ournavy}{$\bullet$} on the far left represents a single sequence from a distinct distribution parameterized by \textcolor{ourmagenta}{$\theta$}). Suppose each observation satisfies $y = f(x, u_y)$ for some noise $u_y$. An in-context sequence \textcolor{ouryellow}{$\bullet$} takes the form of $n$ examples. This is concatenated with index token \textcolor{ourroyal}{$z$} referring back to observed factual observation \textcolor{ourroyal}{$(x_j, y_j)$} when \textcolor{ourroyal}{$z=j$} and the hypothetical new information $\xCF$: $(x_1, y_1, \ldots, x_n, y_n, z, \xCF)$. In-context counterfactual reasoning can be measured via accurate prediction on $\yCF$. Accurate prediction requires \textit{noise abduction} from factual observation, that is, to infer $u_y$ consistent with \textcolor{ourroyal}{$(x_j, y_j)$}, and \textit{prediction} based on the \textit{intervention} $\xCF$ and inferred $u_y$. }
    \label{fig:mainfigure}
\end{figure}

Thinking, in Konrad Lorenz's words, is acting in an imagined space. Counterfactual imagination/reasoning, studied in early childhood cognition development \citep{southgate2014belief} and formalized in causality \citep{scholkopf2022causality, pearl2009, Hernan2024-HERCIW}, predicts the consequences of changes in hypothetical scenarios. It answers "what if" questions and predicts potential outcomes that could have occurred had different actions been taken. Such principled thinking enables \textit{fast} and \textit{responsible} learning, e.g., efficient reinforcement learning \citep{mesnard2020counterfactual, lu2020sample}, counterfactual generative networks \citep{sauer2021counterfactual}, responsible decision-making in complex systems \citep{wachter2017counterfactual}, and fairness \citep{kusner2017counterfactual}.

In precision healthcare, one may ask what would have happened had a patient not received treatment to assess the individualized treatment effect. One step further, \textit{in-context} counterfactual reasoning pertains the hypothetical consequence for an individual after observing the effect of different treatments on multiple patients. \autoref{fig:mainfigure} displays a simplified setting where $n$ patients receive neoadjuvant chemotherapy to treat breast cancer. Given different dosages of doxorubicin, the tumor size reduction is recorded as output. A counterfactual reasoner now imagines the tumor size reduction of patient $j$, had she received $45 \frac{\text{mg}}{\text{m}^2}$ of doxorubicin instead of the prescribed dosage of $65 \frac{\text{mg}}{\text{m}^2}$. Augmented with agentic verification, such principled counterfactual reasoning would be a valuable tool for automatic scientific discovery and a guardrail for safe AI deployment.

To concretely understand to what extent language models perform in-context counterfactual reasoning, we use the counterfactual framework of \citet{pearl2009} and study a controlled synthetic setup similar to \citet{linreg}. That is, let $y = f(x, u_y)$ for some function $f \in \mathcal{F}$, for function class $\mathcal{F}$. The model predicts target $\yCF$ given counterfactual query $\xCF$ conditioned on a prompt sequence $(x_1, y_1, \ldots, x_k, y_k, z, \xCF)$ where $z$ is an index token indicating the position of the factual observation that such counterfactual query is based on. Given factual observation $(x, y)$, counterfactual reasoning requires three steps:

\begin{itemize}
    \item noise abduction: infer $u_y$ that is consistent with the factual observation $(x, y)$, \looseness=-1
    \item intervention $do(X=\xCF)$: intervene on $X$ to consider the hypothetical value $\xCF$, \looseness=-1
    \item prediction $\yCF = f(\xCF, u_y)$: predict the effect of changes $\xCF$ by inheriting noise $u_y$ from the factual observation. 
\end{itemize}
For a complete introduction on the basics of causality, see Section \ref{subsec:exch_causality_icl}. In contrast to natural language, under this controlled setup, we can formally ask 
\begin{center}
    \emph{Can language models perform in-context counterfactual reasoning?}
\end{center}
with concrete metrics $\mathbb{E}\left[\ell(\yCF, y^{\text{CF}}_{\text{pred}})\right]$, for model prediction $y^{\text{CF}}_{\text{pred}}$ and squared error function $\ell$.

This work studies counterfactual reasoning in exchangeable data, as we note language models are pre-trained on a mixture of sequences coming from different distributions. The training dynamics allow random permutations over different input sequences, i.e., implicitly assume sequences are exchangeable (see Def. \ref{def:exchangeable_seq}). \Citet{deFinetti1931FunzioneAleatorio} shows that any exchangeable sequence can be modeled as a mixture of conditionally i.i.d.\ sequences. In other words, there exists a latent variable $\theta$ with probability measure $\pi$ such that 
\begin{align}
    P(s_1, \ldots, s_n) = \int \prod_{i=1}^n p(s_i \mid \theta) d\pi(\theta) \label{eqn:definetti}
\end{align}
for $s_i$ the $i^\text{th}$ sequence. Pre-training thus amounts to learning mutual information $\theta$ between sequences and their prior $\pi(\theta)$. At inference time, the model computes the \textit{posterior predictive distribution}: given a sequence $s_i$ with tokens $x_1^i, \ldots, x_t^i$, the next token prediction writes 
\begin{align*}
    p(x^i_t | x^i_{<t})  =  \int_\theta p(x^i_t|x^i_{<t},\theta) \pi(\theta | x^i_{<t}) \dd \theta .
\end{align*}

\citet{causaldefinetti, dofinetti} have established a theoretical foundation to study observational and interventional causality in exchangeable data. We extend this framework to counterfactuals. Building upon previous work, our contributions are:

\begin{itemize}
    \item We relate in-context counterfactual reasoning to transformations on the factual in-context observations (Lemma \ref{lmm:retrieval}). Extending known identifiability results~\citep{nasr2023counterfactualidentifiability}, we provide theoretical guarantees for our exchangeable setup (\autoref{thm:identifiability}).
    \item We empirically show that in-context counterfactual reasoning emerges in Transformers in form of a designated \textit{noise abduction head} (Section \ref{subsec:emergentbehavior}). Causal probing reveals that the residual stream linearly encodes the latent $\theta$ (Section \ref{subsec:attn_depth}).
    \item We find that data diversity in pre-training, self-attention and model depth are key for Transformers' performance (Sections \ref{subsec:attn_depth} and \ref{subsec:data_diversity_ood}). More interestingly, our findings transfer to cyclic sequential data (Section \ref{chp:dynamical}), demonstrating concrete preliminary evidence that language models can perform counterfactual story generation in sequential data.  \looseness=-1
\end{itemize}

%% file: chapters/background.tex
\section{Background}
\label{chp:preliminaries}

\subsection{Exchangeability and causality}
\label{subsec:exch_causality_icl}

\begin{definition}[Exchangeable sequence]
\label{def:exchangeable_seq}
    A sequence of random variables is exchangeable if for any finite permutation $\sigma$ of its indices,
    \begin{align*}
        \probmeasure(S_1,...,S_n) = \probmeasure(S_{\sigma{(1)}},...,S_{\sigma{(n)}}).
    \end{align*}
Here we treat each random variable $S_i$ as a sequence with bounded context length $t$.
\end{definition}

\textbf{Causality fundamentals.} When $X$ causes $Y$, the structural causal model (SCM) is determined by functional assignments $f_X, f_Y$ and exogenous variables $U_X, U_Y$, i.e., $X := f_X(U_X), Y:= f_Y(X, U_Y)$. An intervention performed on variable $X$ is represented as $\text{do}(X=x)$. Thus, one replaces $X$ with value $x$ and allows $Y$ to inherit the replaced value, i.e., $X := x, Y:= f_Y(x, U_Y)$. A \textit{counterfactual statement} of the form "$Y$ would be $y$ had $X$ been $x$ in situation $U=u$" is often denoted as $Y_x(u) = y$. For explicit representation, we use $\yCF, \xCF$ for values intended for counterfactual predictions. In contrast to the interventional setting, the counterfactual entity inherits noise $u_y$ consistent with the factual observation. \looseness=-1

\subsection{Transformer architecture}
\label{chp:transformerarchitecture}
Transformers~\citep{transformer} operate on a sequence of input embeddings by passing them to blocks consisting of attention and a multi-layer perceptron $(\mathrm{MLP})$. Mimicking present-day large language models, this work focuses on the decoder-only, autoregressive GPT-2~\citep{radford2019language} architecture for next-token prediction. The softmax operation with causal masking is denoted by $\softmaxcausal$. Given a sequence of input embeddings $\mE \in \R^{T \times E}$ of context length $T$ and embedding dimension $E$, the model first projects each token into $D$-dimensional hidden embeddings via $\mX_0 = \mE \Wembd +\pos(\mE) W_P$, for embedding matrix $\Wembd \in \R^{E \times D}$, positional embedding $W_P \in \R^{T \times D}$ with absolute positional encoding. Given an input sequence of embeddings, multi-head attention at layer $l$ passes the embeddings to query, key, value weight matrices $\Qweight^h,\Kweight^h,\Vweight^h \in \R^{D \times D}$ and computes attention per head as $\mA_l^h = \softmaxcausal \left(\mX_{l-1} \Qweight^h \tp{(\mX_{l-1} \Kweight^h)} \right) $. Then the model concatenates the multi-head output and transforms it via output weight matrix $\Oweight \in \R^{H\cdot D \times D}$ as $\mM_l = \Concat(\mA_l^1,...,\mA_l^{H}) \Oweight $. The output is added into the residual stream as $\mR_l = \mX_{l-1} + \mM_l$ and passed to an $\MLP$ $\mX_{l} = \MLP(\mR_l) +\mR_l$. After the last Transformer layer, the embeddings are mapped back to logits through the unembedding matrix: $\mO = \mX_L \Wunem$. We ignore layer normalization. 

%% file: chapters/methodology.tex
\section{In-context counterfactual reasoning}
\label{chp:copying}

We study a linear regression task that requires noise abduction. The structural causal model (SCM) considered is $f_X(U_X) := U_X, f_Y(X, U_Y) := \beta X + U_Y$, where $U_X, U_Y$ are independent exogenous variables. We are interested in accurate counterfactual prediction on $\yCF$ given new information $\xCF$ and factual observation $(x,y) = (u_x, \beta u_x + u_y)$, where $\yCF = \beta \xCF + u_y$. 

\textbf{Pre-training sequence generation.} Our pre-training corpus consists of a mixture of sequences coming from different distributions that share the same causal structure but different functional classes. Throughout, we use the terms \textit{training} and \textit{pre-training} interchangeably. To generate a single sequence, we randomly draw a latent variable $\theta$ and parameterize both the regression coefficient $\beta$ and noise variables $U_X, U_Y$ based on $\theta$. Each sequence takes the form $(x_1, y_1, \ldots, x_{n_i}, y_{n_i}, z, \xCF, \yCF)$, where $n_i$ in-context examples are sampled from the SCM parameterized via $\theta$ and $z$ denotes the index token for that factual observation which the counterfactual query would be based on. We vary the number of in-context examples $n_i$ observed at each sequence to avoid prediction based on information from positional encoding only. In particular, we respect the causal attention mask by ordering observed variables in topological order in a sequence format.

\textbf{Training objective.} We minimize $\mathbb{E}\left[\ell(\yCF, y^{\text{CF}}_{\text{pred}})\right]$, for $y^{\text{CF}}_{\text{pred}}$ the model's predicted outputs and $\yCF$ the true counterfactual completion. We choose $\ell$ to be the mean squared error $(\MSE)$.

In-context counterfactual reasoning can be represented via the \textit{posterior predictive distribution}
\begin{equation}
\label{eqn:posteriorpredictive}
    p(\yCF | x_1, y_1, \ldots, x_n, y_n, z, \xCF) = \int_\Theta \int_\Beta \deltaYCF{\beta \xCF + \yz - \beta \xz} \pbetadot(\beta | \mathbf{x}, \theta) \pi(\theta | \mathbf{x} ) \dd \beta \dd \theta.
\end{equation}
The model is required to learn how to compute posteriors $p_\beta, \pi$ given a query at pre-training. In-context completion reduces to identifying $z$, inserting query elements $(\xz, \yz, \xCF)$ and weighting the resulting $\yzCF$ with the posterior of $\beta$ given $\mathbf{x}$. 

We first present Lemma \ref{lmm:retrieval} and show how accurate prediction on counterfactual values $\yCF$ can be reduced to a transformation on in-context observations. Lemma \ref{lmm:retrieval} incorporates a broad class of functions, including additive noise models (ANMs) \citep{peters2011causal,peters2014causal}, multiplicative noise models and exponential noise models. \looseness=-1
\begin{lemma}[Counterfactual reasoning as transformation on observed values]

\label{lmm:retrieval}
 Suppose $y = T(f(x), u)$ for some function $T: \outputclass \times \noiseclass \longrightarrow \outputclass$. 
Assume for any fixed $f(x) \in \outputclass$, the inverse $\inv{T}(f(x), \cdot)$ exists for all $y$, i.e., $u = \inv{T}(f(x),y)$. Then, counterfactual reasoning  reduces to learning a transformation $h$ on observed factual observations $(x, y)$ and counterfactual information $\xCF$ with
    \begin{align}
        \yCF = h \left(f(\xCF),f(x),y \right),
    \end{align}
    where $h(f(\xCF),f(x),y) = T(f(\xCF),T^{-1}(f(x), y))$ operates on elements of $\outputclass$ only.
\end{lemma}

Lemma \ref{lmm:retrieval} shows that given the existing information in the query, in-context counterfactual reasoning is no more than estimating the transformed function $T, f, T^{-1}$ on the fly -- a task Transformers are known to be capable of performing \citep{linreg, akyürek2023what, vonoswald2023transformerslearn}. The assumptions in Lemma \ref{lmm:retrieval} cover a broad class of functions, for example, 
\begin{itemize}
    \item \textbf{Additive noise models (ANMs).} Functions taking the form $Y = f(X) + U$ for arbitrary function $f$ can be equivalently represented as $y := T(f(x),u) = f(x) + u$. For fixed $f(x)$, the inverse $\inv{T}(f(x), \cdot )$ exists and reads $u = \inv{T}(f(x),y) = y - f(x)$.
    \item \textbf{Multiplicative noise models.} Functions taking the form $Y = f(X) \cdot U$ for arbitrary function $f$. For fixed $f(x)$, the inverse $\inv{T}$ exists and reads $u = \inv{T}(f(x),y) = \frac{y}{f(x)}$.
    \item \textbf{Exponential noise models.} Functions taking the form $Y = \exp(f(X)+U)$ with inverse $u = \inv{T}(f(x),y) = \log(y) - f(x)$.
    \looseness=-1 
\end{itemize}

On top of that, \citet{nasr2023counterfactualidentifiability} provide three cases under which the class of bijective generation mechanisms (BGMs) is counterfactually identifiable. Under a set of assumptions, one can disentangle the causal mechanisms solely by knowing the causal graph and observational data. Our setup can be subsumed under that framework of BGMs. We therefore extend the result in the Markovian case \citep{nasr2023counterfactualidentifiability} to our exchangeable setup and show that the transformation on observed values is counterfactually identifiable. Details are deferred to Appendix~\ref{app:pathexp}.

\begin{theorem}[Counterfactual identifiability under exchangeability]
    Let $X, Y, U, \theta$ scalar random variables with $X \ind U | \theta$. Take $T: \outputclass \times \noiseclass \longrightarrow \outputclass$ with $y = T(f(x),u)$. Assume $\forall f(x) \in \mathcal{Y}$, the inverse $T^{-1}(f(x),\cdot)$ exists for all $y$, i.e., $u = T^{-1}(f(x),y)$, and $\forall f(x) \in \mathcal{Y}$, $T(f(x), \cdot)$ is continuous. Suppose further that $f(x)$ continuous $\forall x$, and strictly monotonic in $x$. Then, set $Y = T(f(X),U)$. Given the joint law $\mathbb{P}_{X,Y|\theta}$, $T$ is counterfactually identifiable.
    \label{thm:identifiability}
\end{theorem}

%% file: chapters/paper-chapters/experiments.tex
\input{chapters/experiments}
\input{chapters/emergence}

%% file: chapters/experiments.tex
\section{Experiments}
\label{chp:experiments}

Based on the training setup described in Section \ref{chp:copying} with details in Appendix \ref{app:trainingdetails}, we choose a controlled synthetic setup to allow concrete evaluation on in-context counterfactual reasoning.

\subsection{Language models can in context perform counterfactual reasoning on linear functions}
\label{subsec:can_cf_reasoning}
We assess the impact of model architecture choice on in-context counterfactual reasoning performance. In contrast to natural language tasks, the performance in our task is clearly defined and can be measured as: \looseness=-1
\begin{itemize}
    \item \textit{accuracy} in prediction: $\MSE(\yCF, \yCFhat) $,
    \item \textit{learning efficiency}: number of in-context examples seen to achieve satisfactory prediction.  \looseness=-1
\end{itemize} 
We compare the \textsc{standard} decoder-only GPT-2 Transformer architecture against several autoregressive baselines. We use the terms \textsc{standard} and GPT-2 interchangeably. The models considered are:
\begin{itemize}
    \item \textsc{GPT-2}~\citep{radford2019language}: 12 layers, 8 attention heads, hidden dimension 256, \looseness=-1
    \item \textsc{LSTM}~\citep{hochreiter1997lstm}: 2 layers, hidden dimension 256,
    \item \textsc{GRU}~\citep{cho2014learning}: 2 layers, hidden dimension 256,
    \item \textsc{Elman RNN}~\citep{elman1990finding}: 2 layers, hidden dimension 256.
\end{itemize}

We evaluate each model on unseen sequences sampled in-distribution and report results averaged over $6400$ sequences. Our synthetic generation yields $\mathbb{E}[\YCF] = 0$ and $\log(\mathrm{std}(\YCF)) = 2.56$. Appendices \ref{app:trainingdetails} and \ref{app:modeldetails} include data generation particularities as well as experimental and model details.

\begin{figure}[h]
    \centering
    \includegraphics[width=0.6\linewidth]{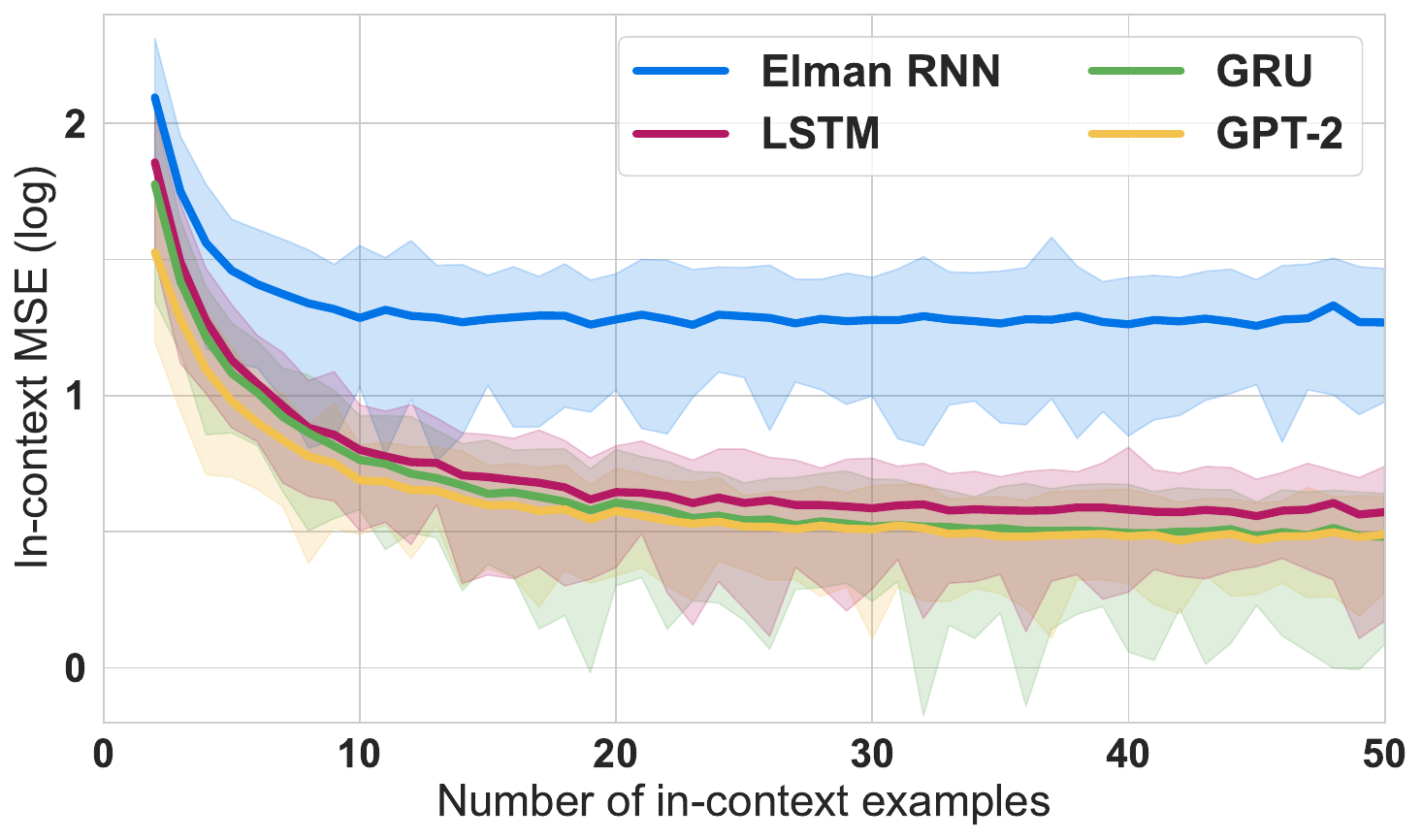}
    \caption{\textbf{Model comparisons for in-context counterfactual reasoning.} In-context counterfactual prediction accuracy measured via log-transformed $\MSE$ averaged over $6400$ sequences versus the number of in-context examples observed in a prompt. We compare GPT-2 (\textsc{standard}), LSTM, GRU, and Elman RNN. Though not significant, GPT-2 achieves lowest error and fastest convergence rate for a small number of in-context examples. For more than $14$ in-context examples, the Elman RNN obtains significantly~\citep{efron1979basicbootstrap} higher in-context $\MSE$ than the three other architectures. \looseness=-1}
    \label{fig:performanceregression}
\end{figure}
\autoref{fig:performanceregression} compares the impact of different model architectures on in-context counterfactual reasoning. We plot the log-transformed in-context $\MSE$ against the number of in-context examples observed in a prompt. We hypothesize that more observed examples lead to higher counterfactual prediction accuracy, and that better models require less in-context examples for accurate prediction. For contexts of over $14$ in-context examples, the Elman RNN achieves significantly higher in-context $\MSE$ than the rest. Across LSTM, GRU and \textsc{standard} Transformer, we observe no significant performance difference. In Appendix~\ref{app:modeldetails}, we reference literature discussing the RNN architectures. For now, we confine ourselves to autoregressive Transformers such that all future statements refer to this setup only. \looseness=-1

\subsection{Counterfactual reasoning emerges in self-attention}
\label{subsec:attn_depth}
\textbf{Self-attention.} Lemma \ref{lmm:retrieval} shows the reliance of counterfactual reasoning on copying in-context observed values. Self-attention has been shown to perform copying by implementing induction heads~\citep{olsson2022incontextlearninginductionheads,akyürek2024incontextlanguagelearningarchitectures}. We thus hypothesize that counterfactual reasoning emerges in the attention sublayers of the Transformer.

\begin{figure}[h]
    \centering
    \begin{subfigure}[t]{0.49\linewidth}
         \centering
         \includegraphics[width=\textwidth]{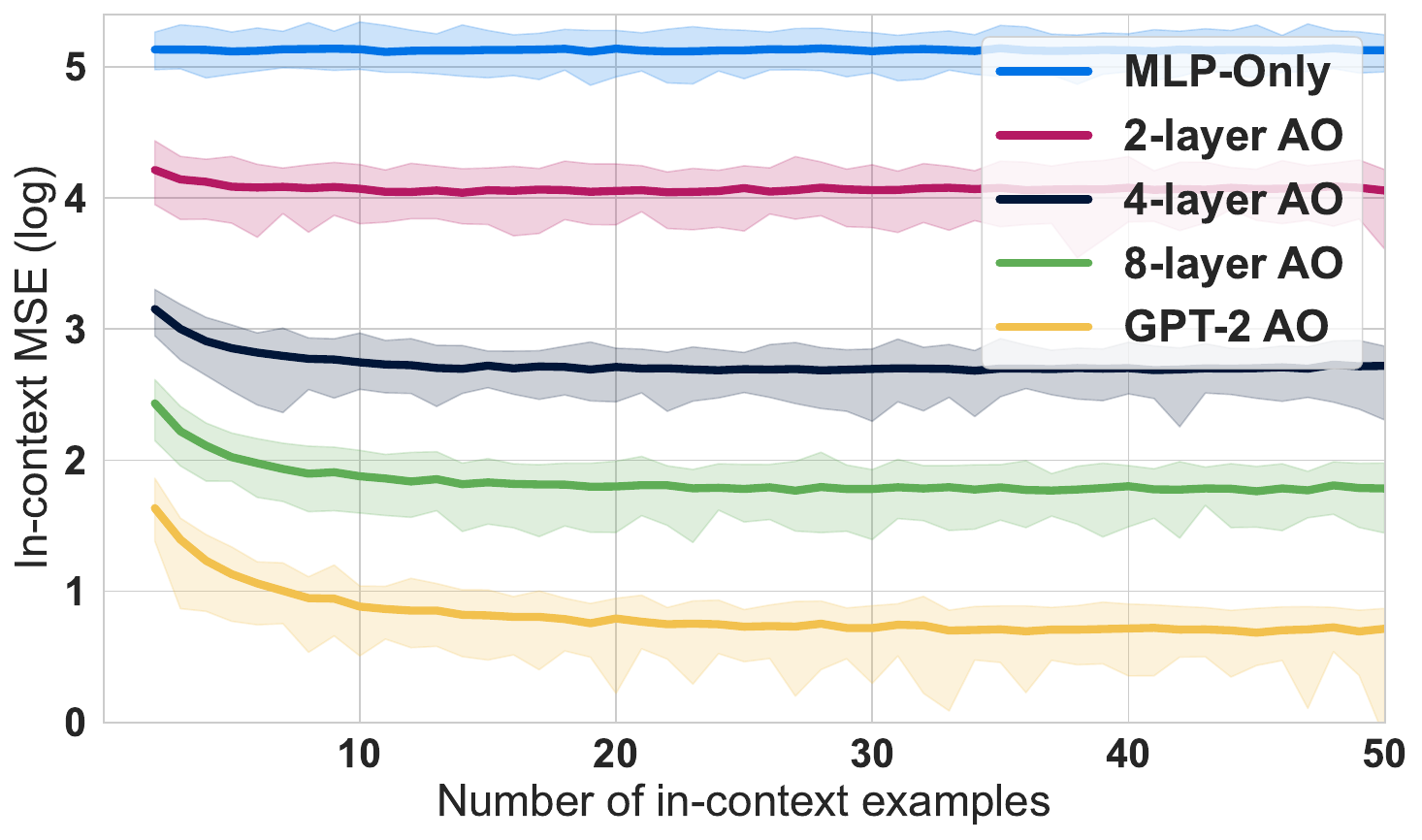}
         \caption{\textbf{Switching attention.} In-context $\MSE$ stagnates with increasing number of in-context examples observed for the \textbf{MLP-Only} model. We compare to attention-only (\textbf{AO}) Transformers of $2$, $4$, $8$ layers with $1$ head each as well as the \textsc{standard} setup. For all Transformers, in-context $\MSE$ decreases as more in-context examples are observed. \looseness=-1 }
         \label{fig:mse_attention_only}
     \end{subfigure}
     \hfill
     \begin{subfigure}[t]{0.49\linewidth}
         \centering
         \includegraphics[width=\textwidth]{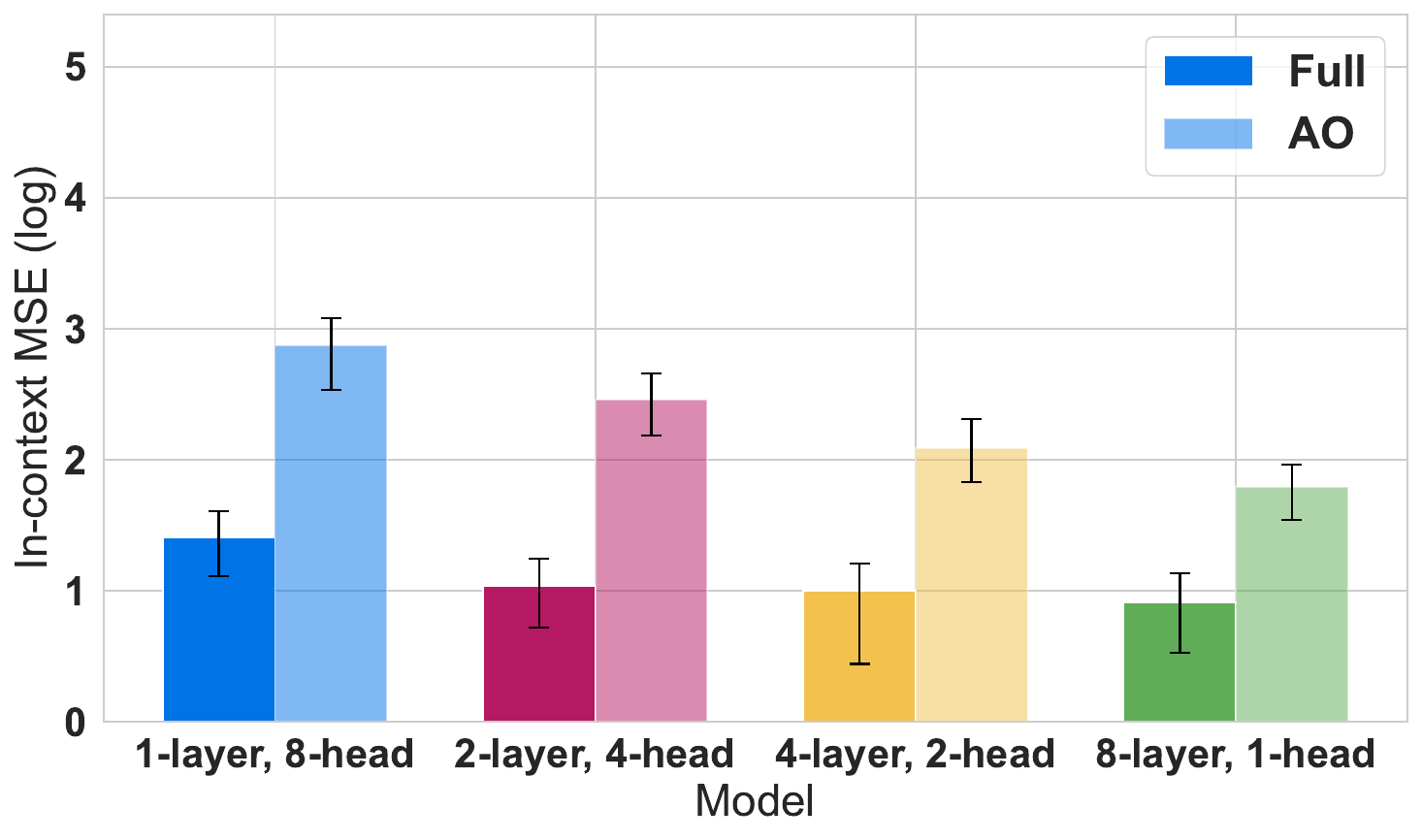}
         \caption{\textbf{Varying depth.} Keeping the number of attention heads constant at $8$, we compare \textbf{Full} and attention-only (\textbf{AO}) Transformers with $1$ layer, $8$ heads; $2$ layers, $4$ heads; $4$ layers, $2$ heads; $8$ layers, $1$ head. We observe a decrease in in-context loss as model depth increases. We evaluate on $6400$ sequences of $35$ in-context examples each.}
         \label{fig:bar_chart_depth}
    \end{subfigure}
    \caption{\textbf{Attention and model depth matter.}}
    \label{fig:depth}
\end{figure}

We conduct an experiment that switches on and off attention layers in the Transformer and separately train each model on our task. \autoref{fig:mse_attention_only} shows the in-context $\MSE$ for \textbf{MLP-Only} and \textbf{AO} (attention-only) with $2$, $4$, $8$ layers and the \textsc{standard} setup. At training time, we observe that \textbf{AO} models mimic the loss curve of the \textsc{standard} setup, while the \textbf{MLP-Only}'s loss remains constant at the level of the theoretical variance, $\Var(\YCF)$. \autoref{fig:mse_attention_only} shows that counterfactual reasoning performance improves for all considered Transformers with increasing context length. Although at a higher loss than the \textsc{standard} Transformer, smaller models do perform better on longer contexts. Oblivious of context length, \textbf{MLP-Only} does not appear to reason counterfactually.

\textbf{Model depth.}
\citet{elhage2021transformercircuits} show that self-attention performs a copying task. Lemma~\ref{lmm:retrieval} relates counterfactual reasoning to copying multiple values and learning transformations. Such a task requires the composition of attention heads to pass information across layers. We hypothesize that model depth plays an important role. Fixing the total number of attention heads at $8$, we train $4$ Transformer-based models with increasing depth: from $1$-layer, $8$-head Transformers to models with $8$ layers of $1$ head each. We expect that if depth does not influence performance, given the same number of attention heads, all models should have comparable in-context $\MSE$. Figure \ref{fig:bar_chart_depth} represents the loss on $6400$ sequences of $35$ in-context examples for both \textbf{Full} and \textbf{AO} models. Additional results can be found in appendix~\ref{app:modeldetails}. With increasing depth, loss declines for both setups. In particular, the \textbf{Full} $8$-layer, $1$-head Transformer yields lowest MSE, while the $4$-layer, $2$-head designs achieve competitive results. 

\begin{figure}[h]
    \centering
    \begin{subfigure}[t]{0.49\linewidth}
         \centering
         \includegraphics[width=\textwidth]{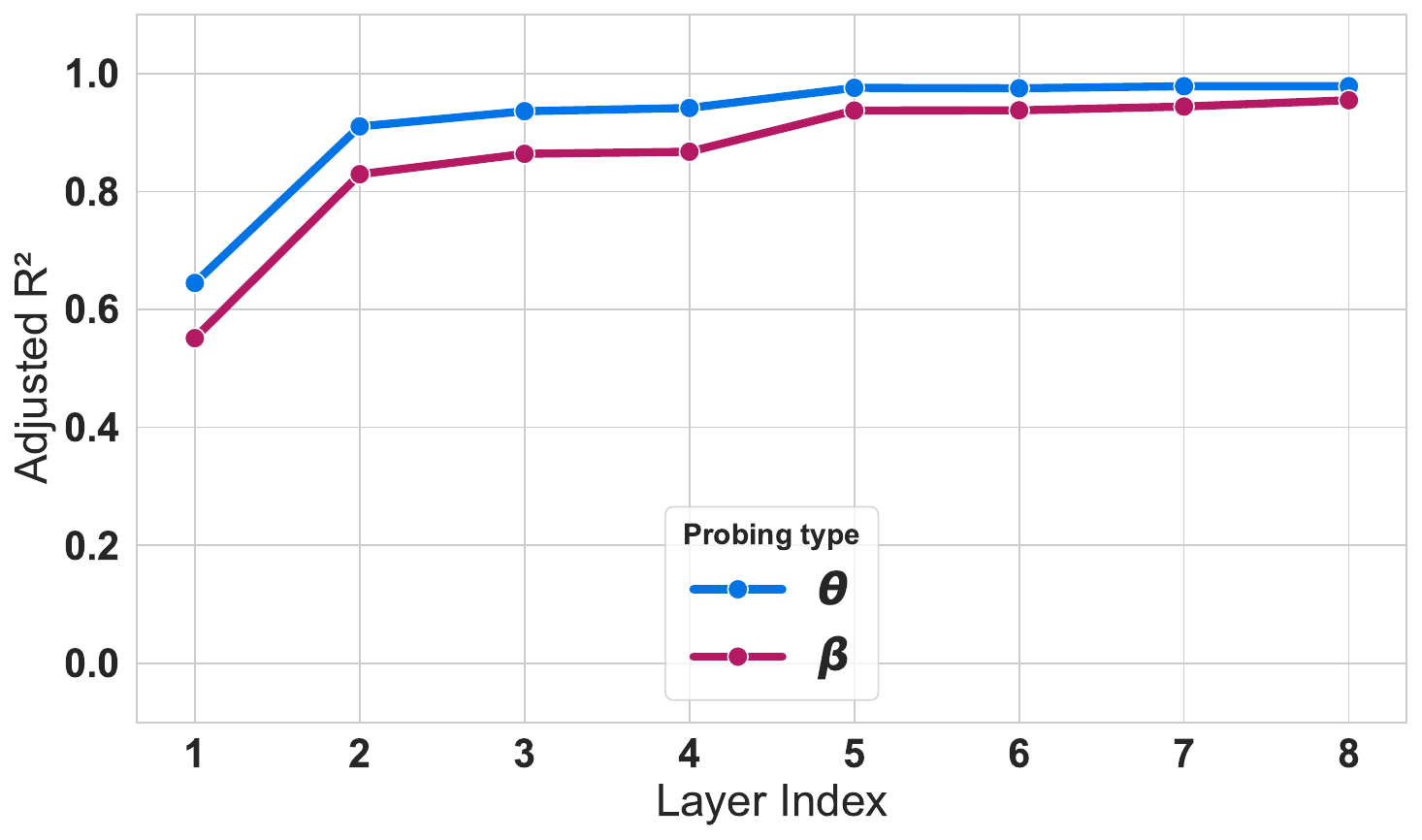}
         \caption{\textbf{Relevance at every layer.} We plot the $\adjR$ after every layer against the layer indices. Starting at the second layer, we observe $\adjR > 0.9$ for predicting $\theta$, indicating that the latent is encoded linearly in the residual stream. \looseness=-1 }
         \label{fig:probing_all_layers}
     \end{subfigure}
     \hfill
     \begin{subfigure}[t]{0.49\linewidth}
         \centering
         \includegraphics[width=\textwidth]{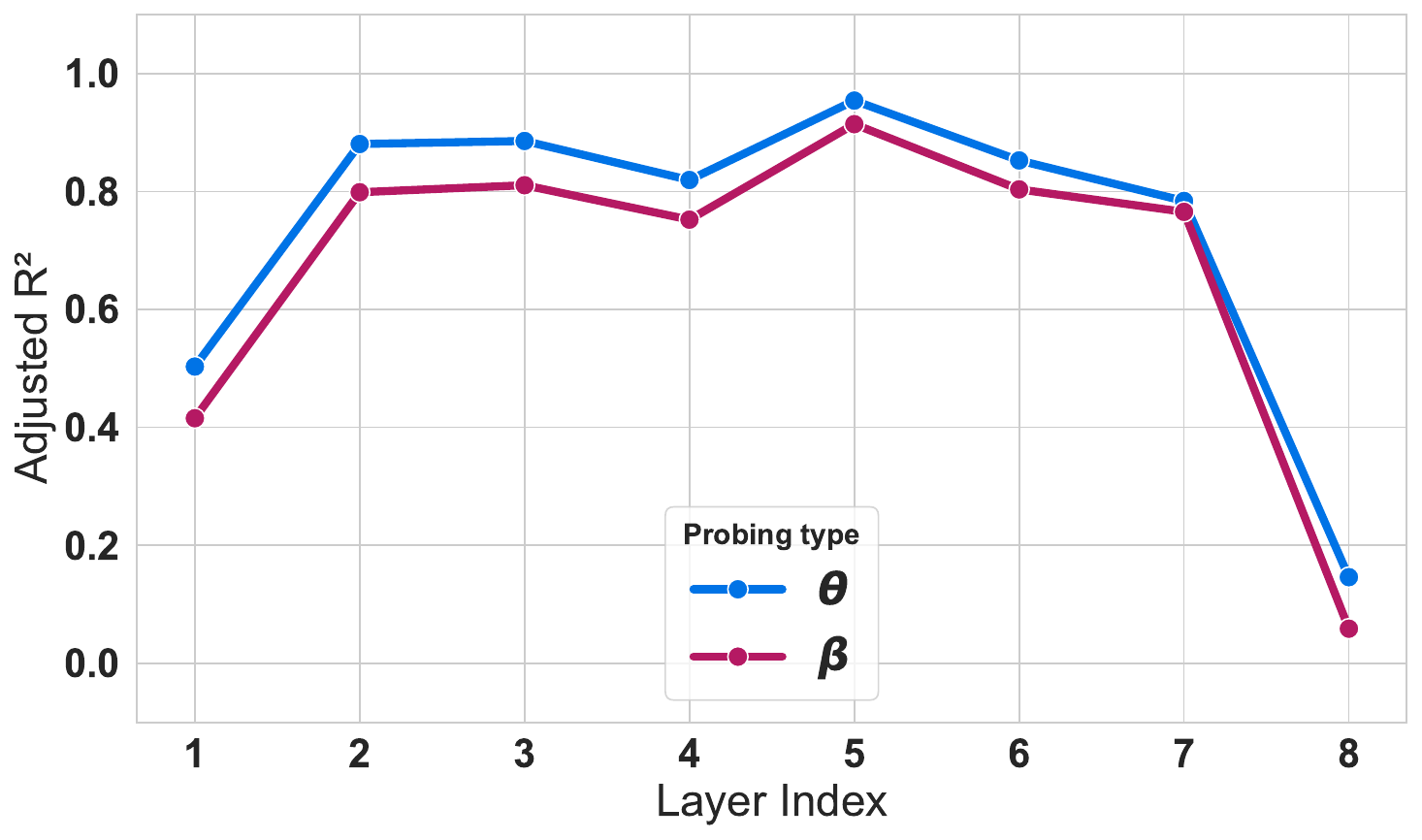}
         \caption{\textbf{Additional relevance of this layer.} We subtract the residual streams before each layer from the residual stream after and report the $\adjR$. Layers $2$ and $5$ are especially relevant while layer $8$ does not add substantial information. \looseness=-1 }
         \label{fig:probing_diff}
    \end{subfigure}
    \caption{\textbf{The Transformer linearly encodes the latent parameter.} We train a linear probe on $6400$ prompts from a fresh evaluation set after every layer. We evaluate on $1280$ sequences. All layers after the first one encode relevant information for predicting $\theta$ from the residual stream only.}
    \label{fig:probing}
\end{figure}

\textbf{Latent parameter.}
A final way to confirm the relevance of self-attention is by probing the residual stream for the latent $\theta$. As $\theta$ governs the distribution of the in-context sequence, it is natural to check for that signal. By doing so, we find that the $8$-layer \textbf{AO} Transformer learns the latent parameter early on. \autoref{fig:probing_all_layers} shows that a linear model predicting $\theta$ from the residual stream after layer $2$ does so at $\adjR \approx 0.9104$. After this initial spike, all future layers absorb that information as the $\adjR$ stays above $0.9$. In addition we train separate probes on the regression parameter $\beta$ and find similar results to probing for $\theta$. As $\beta \sim \Normal(\theta, 1)$, this is in line with our intuition. Moreover, \autoref{fig:probing_diff} checks whether the difference of residual streams after and before every layer carries latent information. In fact, we have additional evidence that layers $2$ and $5$ are central in learning $\theta$ (and $\beta$), while the final layer's contribution is marginal at $\adjR < 0.2$. By running this standard experiment, we find that self-attention alone suffices to linearly encode the latent theme. We thus provide additional support for the Linear Representation Hypothesis~\citep{park2024lrh}.

\subsection{Data diversity and out-of-distribution (\textit{OOD}) generalization}
\label{subsec:data_diversity_ood}

Causal structure identification, a previously deemed impossible task from i.i.d.\ observational data alone \citep{pearl2009}, has been shown to be feasible in multi-domain data \citep{causaldefinetti}, a natural setting for exchangeable data. We hypothesize that (counterfactual) reasoning is similarly only feasible under diverse pre-training data. 
As a measure of pre-training data diversity, we use the effective support size \citep{grendar2006effectivesupportsize}, defined as
\begin{align*}
    \ESS(\theta) := \exp\left( -\sum_{\vartheta \in \Theta_0} p(\vartheta) \log p(\vartheta) \right) = \exp \left(H(\theta) \right),
\end{align*}
where  $\Theta_0$ is the space of latent $\theta$ sampled at pre-training, and $H$ the Shannon entropy~\citep{shannon1948entropy}. We further compare the pre-training data diversity's impact on simple \textit{OOD} generalization. We pre-train on both \textsc{uniform} and \textsc{normal} sampling of latent $\theta$ and evaluate on \textsc{uniform} sampled test data. \autoref{fig:mse_vs_ess} shows a clear trend that diverse pre-training data, measured as a higher effective support size, yields lower in-context $\MSE$ for both in-distribution and \textit{OOD} cases. Appendix~\ref{app:ood-diversity} shows that such generalization ability persists when evaluated on \textsc{normal}, and that in-context loss declines with increasing data diversity across a changing number of in-context examples. We also discuss how this relates to existing research. \looseness=-1

%% file: chapters/emergence.tex
\subsection{Emergent model behavior}
\label{subsec:emergentbehavior}
\begin{figure}[h]
    \centering
    \begin{subfigure}[t]{0.49\linewidth}
        \centering
        \includegraphics[width=\linewidth]{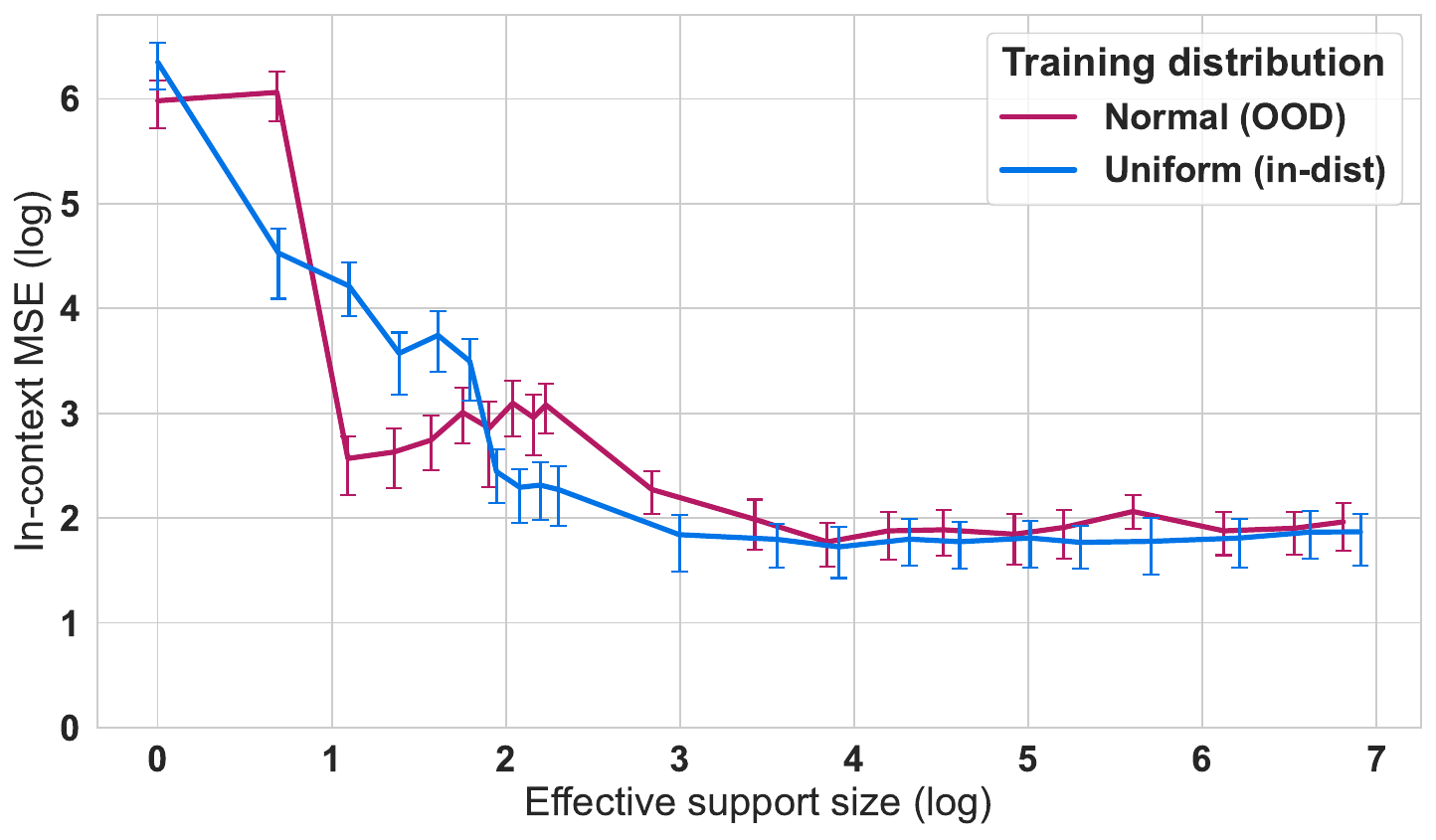}
        \caption{\textbf{Data diversity in pre-training.} In-context $\MSE$ (log-scaled) averaged over $6400$ prompts at $35$ in-context examples against log-scaled effective support size. Each point represents one fully pre-trained model on either the \textsc{uniform} sampled or \textsc{normal} sampled $\theta$. Models are evaluated on \textsc{uniform}. \looseness=-1}
        \label{fig:mse_vs_ess}
     \end{subfigure}
     \hfill
     \begin{subfigure}[t]{0.49\linewidth}
         \centering
         \includegraphics[width=\textwidth]{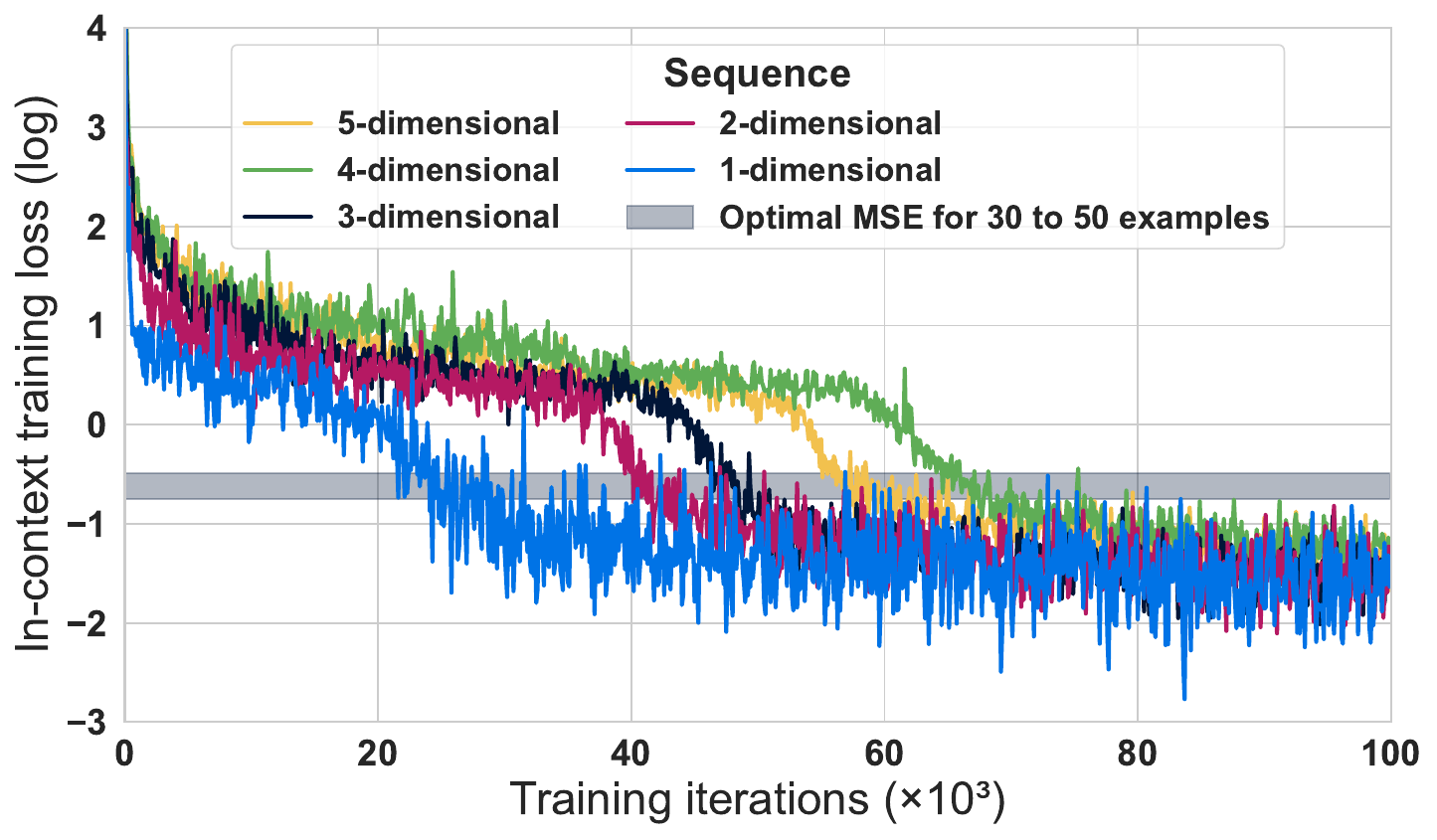}
         \caption{\textbf{Phase transitions during training.} In-context training $\MSE$ against the number of training steps. When training on $100'000$ iterations, we observe that the model with $E = 5$ reaches a phase transition after more than $50'000$ steps. Depending on embedding dimension $E$, the phase transition occurs sooner or later. \looseness=-1 }
         \label{fig:phase_transitions}
    \end{subfigure}
    \caption{\textbf{Data diversity and emergence.} In-context counterfactual reasoning \textit{emerges} at training. To generalize to unknown $\theta$, the model is to be trained on a sufficiently diverse pre-training corpus. \looseness=-1}
    \label{fig:emergence}
\end{figure}

The ordinary least squares estimator for the regression parameter $\beta$ writes
\begin{align*}
    \hat{\beta} = \frac{\sum_{i = 1}^n (x_i - \bar{x})(y_i - \bar{y})}{\sum_{i = 1}^n (x_i - \bar{x})^2}
\end{align*}
which leads to the fitted value $\yCFbhat = \hat{\beta} (\xCF - x_z) + y_z$, and the baseline $\MSE$ between the linear regression implementation $\yCFbhat$ and the ground truth counterfactual, $\MSE \left( \yCFbhat, \yCF \right)$. We refer to this as the \textit{(estimated) optimal $\MSE$} and note that it lies between $0.6132$ and $0.4739$ for $30$ to $50$ in-context examples. We therefore compare models trained on data with embedding dimension $E \in \{1,2,3,4,5\}$ to this estimated optimal $\MSE$. Here, we fix $z = 14$ to enable faster learning relative to the setup with variable index token. \autoref{fig:phase_transitions} displays the training progression for \textbf{Full} GPT-2 Transformers trained for $100'000$ steps. We observe that the ability to attain the estimated optimal $\MSE$ \textit{emerges} as training proceeds~\citep{wei2022emergent,nanda2023progress}. Given that we primarily train on $50'000$ training steps, we observe that at $E = 5$, the counterfactual reasoning behavior is yet to emerge. For sufficient training length, therefore, the model implements a solution performing better than the estimated optimal $\MSE$. In Appendix~\ref{app:ood-diversity}, we confirm this behavior for variable $z$ by introducing a designated \textit{noise abduction head} which emerges at the $7^\text{th}$ layer of the \textbf{Full} GPT-2 architecture. We also move beyond linear regression, and we compare our results to models trained on the setup in~\citet{linreg}. \looseness=-1

%% file: chapters/paper-chapters/cyclic.tex
\input{chapters/cyclic-body-1}

\input{chapters/cyclic-uniform}

\input{chapters/cyclic-body-2}
\input{chapters/cyclic-paper}

%% file: chapters/cyclic-body-1.tex
\section{Cyclic sequential dynamical systems}
\label{chp:dynamical}

Language is sequential. Sentences within a story depend on each other. Counterfactual story generation naturally occurs by prompting the model with a factual story and querying it to complete the story under a hypothetical scenario. This is done while keeping the noise unchanged. Although difficult to evaluate in language, our key insight is that such behavior can be mimicked by ordinary differential equations modeling the underlying causal mechanism~\citep{MooJanSch13,peters2020dynamicalsystems, lorch2024sde}. Given an initial condition $(x_0,y_0)$, a dynamical system determines the state $(x_t,y_t)$, for $t \geq 0$. We model causal dependencies using Itô stochastic differential equations (SDEs), \looseness=-1
\begin{align}
    \begin{split}
        \dd Y_t &:= f(X_t, Y_t) \dd t + \sigma_Y \cdot \dd W_t \\
        \dd X_t &:= g(X_t, Y_t) \dd t + \sigma_X \cdot \dd U_t \\
        X_0 &:= \xi_0 \text{ and } Y_0 := \upsilon_0,
    \end{split}
\label{eqn:sdes}
\end{align}
for initial condition $(\xi_0, \upsilon_0)$. Here, $W_t, U_t$ denote independent Brownian motions with constant diffusion by $\sigma_X,\sigma_Y$, and $f, g$ represent the drift coefficients for $X$, $Y$, respectively.

%% file: chapters/cyclic-uniform.tex
Autoregressive models are discrete.
To address the challenge of adapting continuous SDE data generation to discrete observations, we draw $n$ event times from a uniform distribution over a bounded interval. We evaluate the above SDE at time $t_m$ for all $m \leq n$ and input observations to the model. Instead of deterministically slicing the interval into equivalent regions, we can capture the complete bounded interval as the randomly sampled set of event times, independent across iterations. Noise abduction is thus required over both the realization of $u$ and of event times $t_m \leq t_n$. 

%% file: chapters/cyclic-body-2.tex
In summary, we provide context $(x_{t_1},y_{t_1},...,x_{t_n},y_{t_n})$, a counterfactual token indicating the start of counterfactual story generation, and the start of the counterfactual sequence $(x_{t_1}^\text{CF},y_{t_1}^\text{CF})$. Then, we train the model on completing the full counterfactual story. We thus ask the model to generate $(x_{t_2}^\text{CF},y_{t_2}^\text{CF},...,x_{t_n}^\text{CF},y_{t_n}^\text{CF})$. Models are evaluated on the $\MSE$ across all predicted tokens. Contrary to above, we assess performance not just on the final token, but over the entire counterfactual continuation consisting of $(n-1) \cdot 2$ tokens. Additionally, the full context corresponds to a \textit{single} instance of a factual story. Given this story, the model must infer the noise associated with all observational examples, rather than just a single example pair. We thus have deterministic functional assignments $f,g$, and copy the noise $W_t,U_t$ from the observational sequence. With constant diffusion, we invoke Lemma \ref{lmm:retrieval}, which connects the cyclic extension to the regression setup,
\begin{align}
    \YiCFdot{t} &= \int_0^t f(\XiCFdot{s},\YiCFdot{s}) \dd s - \int_0^t f(X_{s},Y_s) \dd s + Y_t.
\label{eqn:transformationsde}
\end{align}
We conduct preliminary experiments on data generated from the Lotka-Volterra model~\citep{lotka1910volterra} describing predator-prey dynamics,
\begin{align}
    f(X_t, Y_t) &:= \alpha X_t - \beta X_tY_t \label{eqn:preyconcentration} \\ 
    g(X_t, Y_t) &:= -\gamma Y_t + \delta X_t Y_t \label{eqn:predatorconcentration} 
\end{align}

%% file: chapters/cyclic-paper.tex
\begin{figure}[h]
    \centering
    \begin{subfigure}[t]{0.49\linewidth}
         \centering
         \includegraphics[width=\textwidth]{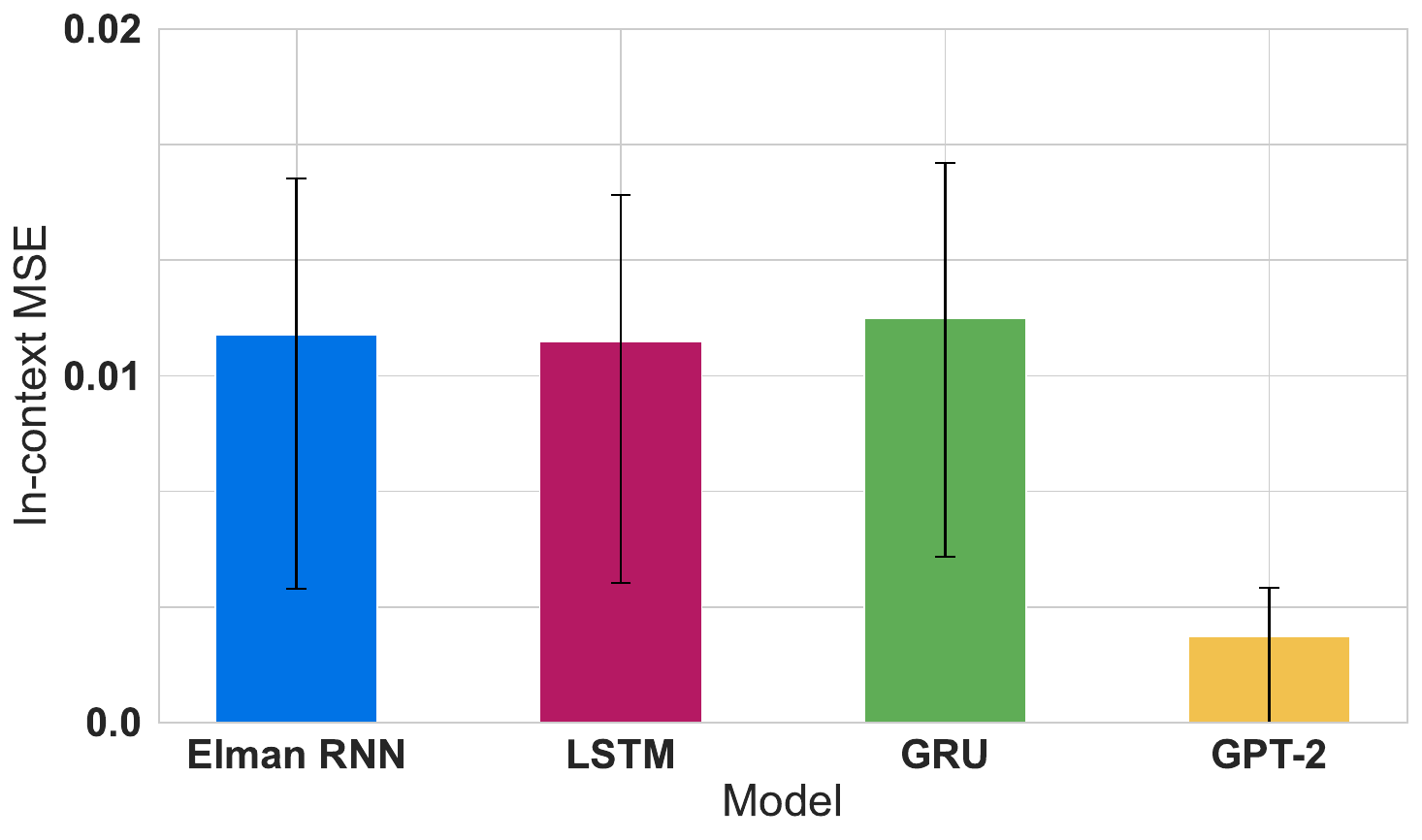}
        \caption{\textbf{In-context evaluation on Lotka-Volterra SDEs.} In-context $\MSE$ of four models queried with data on $20$ distinct time points. The $\MSE$ is competitive for Elman RNN, LSTM, and GRU at over $0.1$. For the \textbf{Full} GPT-2 Transformer, we observe $\MSE \approx 0.0025$, which is significantly below the performance of the former three architectures.}
        \label{fig:cyclicperformance}
     \end{subfigure}
     \hfill
     \begin{subfigure}[t]{0.49\linewidth}
         \centering
         \includegraphics[width=\textwidth]{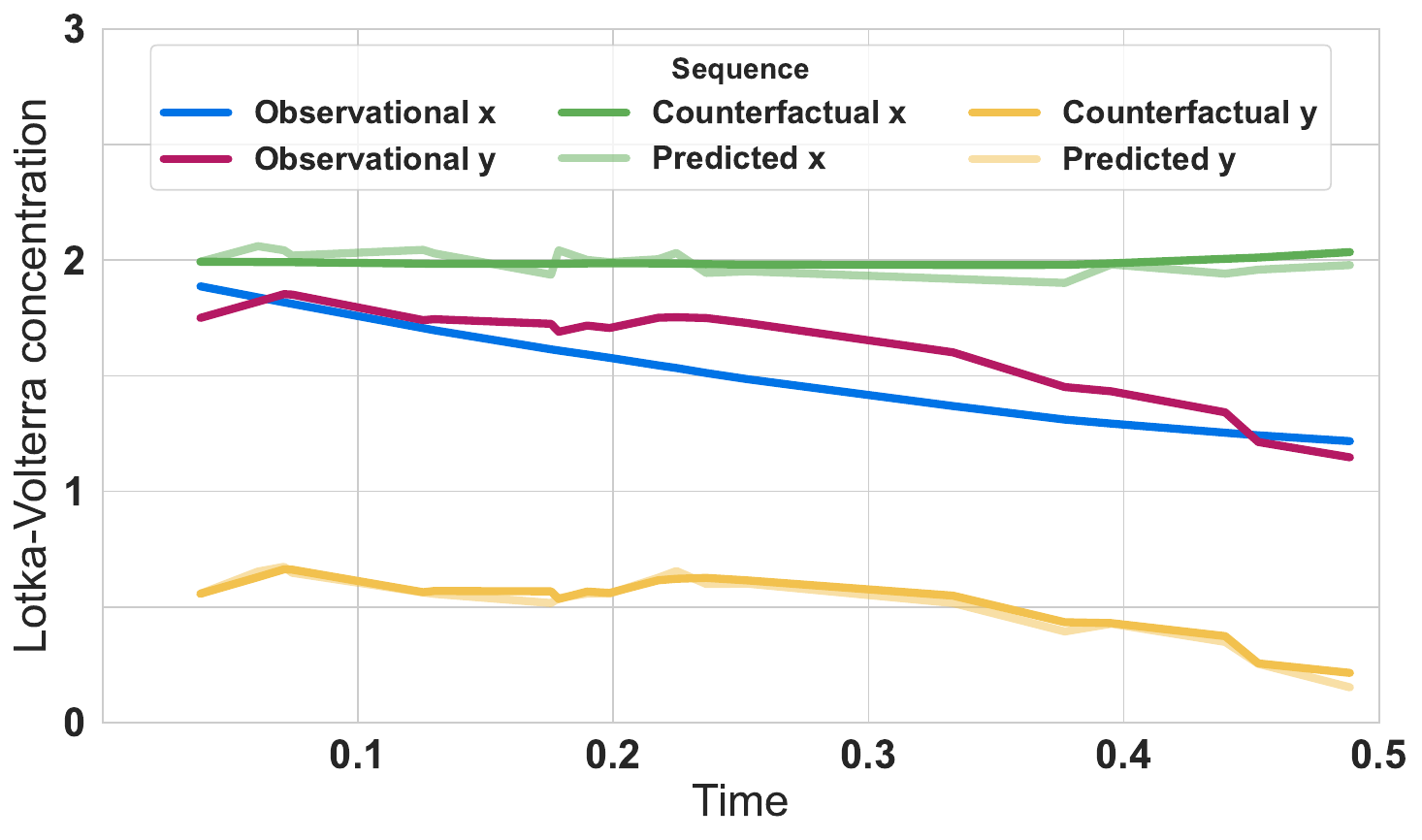}
         \caption{\textbf{Prediction of Lotka-Volterra SDEs in-context.} The query consists of the observational interwoven $(\textcolor{ourroyal}{x},\textcolor{ourmagenta}{y})$ data. In-context, the model learns the dynamics governed by the latent $\theta$ and completes the counterfactual trajectory for all $t_m > 0$. Although observational $\textcolor{ourmagenta}{y}$ spans $[1.1,1.9]$, the model infers that counterfactual \textcolor{ouryellow}{$\yCF$} is constrained to $[0.1,0.7]$.}
         \label{fig:prediction_lv}
    \end{subfigure}
    \caption{\textbf{Cyclic causal relationship.}}
\label{fig:cyclic}
\end{figure}

with $\alpha,\beta,\gamma,\delta \geq 0$. Appendix \ref{app:cyclic} specifies training and data generation details and analyzes the attention behavior in the Transformer. \autoref{fig:cyclicperformance} shows that the \textbf{Full} GPT-2 Transformer obtains the lowest in-context $\MSE$ across the four studied models. Error bars~\citep{efron1979basicbootstrap} point to competitive performance across RNN-type architectures. We evaluate the models on contexts of $n = 20$ realizations of in-context pairs. Recall that these are not \textit{conditionally i.i.d.} in-context examples as above, but rather follow a sequential pattern. \autoref{fig:prediction_lv} illustrates this sequentiality. Given a query consisting of interleaved observations of the prey, predator concentrations $(\textcolor{ourroyal}{x}, \textcolor{ourmagenta}{y})$, the model infers the underlying latent $\theta$ governing the system. It then infers the full counterfactual trajectory beyond $(x_{t_1}^\text{CF},y_{t_1}^\text{CF})$. Notably, while the observed $\textcolor{ourmagenta}{y} \in [1.1, 1.9]$, the model correctly infers that the counterfactual is realized at a lower level, $\textcolor{ouryellow}{\yCF} \in [0.1,0.7]$. This indicates that the model internalizes the underlying dynamics and uses them to generate coherent predictions.

%% file: chapters/discussion.tex
\section{Discussion and related work}
\label{chp:discussion}

\textbf{Counterfactual reasoning in natural language.}  
Prior investigation into \textit{counterfactual story generation} has been mainly focused on empirical evaluation on off-the-shelf language models. \citet{tandon2019wiqa,li2023counterfactualreasoning} study lexical associations between factual statements and the hypothetical \textit{"what if"} scenario; \citet{qin2019counterfactualstory} provide a dataset for counterfactual rewriting evaluation. \citet{bottou2025fiction} link the ability to hypothesize about counterfactuals to generating meaningful text beyond the training distribution. \citet{ravfogel2025gumbel} study language directly as structural causal acyclic models and \citet{yan2023counterfactual} focus on representation identification for adaptive counterfactual generation. This work respects the sequential nature of language and performs controlled pre-training and concrete evaluations that demonstrate preliminary and promising evidence that language models are capable of counterfactual story generation.

\textbf{In-context learning.}
Language models~\citep{brown2020fewshot}
have shown surprising in-context reasoning ability demonstrating the potential for practical user interactions in the form of chatbots. Transformers, the backbone architecture for modern language models, have been shown to in-context learn different function classes~\citep{linreg}. Some studies investigate the models' ability to perform gradient descent over in-context examples~\citep{dai2023why,deutch-etal-2024-context,vonoswald2023transformerslearn} while others discuss implicit Bayesian inference~\citep{xie2022explanationincontextlearningimplicit,pmlr-v235-falck24a,ye2024exchangeablesequencemodelsquantify} and \textit{induction heads}~\citep{olsson2022incontextlearninginductionheads,akyürek2024incontextlanguagelearningarchitectures}. A recent line of work considers guarantees for in-context learning algorithms on linear models~\citep{akyürek2023what}. \looseness=-1

\textbf{Causality and exchangeability.}
\citet{pearl2009, Hernan2024-HERCIW} formalize the notion of counterfactuals in causality. \citet{peters2020dynamicalsystems, lorch2024sde} study cyclic causal models via differential equations. Mixture of i.i.d.\ data or multi-domain data have been the foundation of modern pre-training corpora. \citet{causaldefinetti} demonstrate that inferring causality is feasible in such a mixture of i.i.d.\ data, providing the potential for language models to exhibit causal reasoning capabilities. Recent work has continued to investigate connections between causality and exchangeability from intervention \citep{dofinetti} to representation learning \citep{reizinger2024identifiable}. Concurrently, interest in exchangeability and language models has resurfaced. \citet{zhang2023deepfinettirecoveringtopic} study topic recovery from language models and \citet{pmlr-v235-falck24a,ye2024exchangeablesequencemodelsquantify, xie2022explanationincontextlearningimplicit} discuss in-context learning as performing implicit Bayesian inference in the manner of de Finetti. This provides potential for \textit{causal} de Finetti, which allows structured in-context learning with the potential for principled reasoning and fast inference. This work serves as a preliminary study on in-context counterfactual reasoning in the bivariate case and demonstrates its potential. 

\textbf{Limitations.} Our study is focused on controlled synthetic datasets to circumvent evaluation difficulties inherent to natural language. We therefore assess reasoning from a causal perspective~\citep{elements,mondorf2024beyond} as algebraic manipulation of existing knowledge~\citep{bottou2011reasoning}. We only train on small-scale GPT-2 type models to demonstrate preliminary evidence on counterfactual story generation. Furthermore, we assume no observed confounders, which is difficult to satisfy in the real world, as observations are finite and contain missing information which requires \textit{out-of-variable} generalization~\citep{guoout}. Realistic benchmarks on natural language evaluation and larger-scale pre-training that provide principled reasoning for real-world impact, such as scientific reasoning, represent potential avenues for future research. 

\textbf{Broader impacts.}
Since our models are not trained on natural language, the typical societal concerns associated with counterfactual story generation do not directly apply. We view our current work as a scientific exploration of desirable properties for the next generation of AI models. 

%% file: chapters/conclusion.tex
\section{Conclusion}
We discuss in-context counterfactual reasoning in both linear regression tasks and cyclic sequential data. We observe that counterfactual reasoning can emerge in language models. We find that Transformers, in particular, rely on self-attention and model depth for in-context counterfactual reasoning. Data diversity in pre-training is essential for the emergence of reasoning and \textit{OOD} generalization. We provide insights on how counterfactual reasoning can be interpreted as a copying and function estimation task, with theoretical and empirical evidence over a broad class of invertible functions~\eqref{lmm:retrieval}. We mechanistically support our findings by introducing a designated \textit{noise abduction head} and probing the residual stream for the latent concept. We develop a sequential extension grounded in the Lotka-Volterra model of predator-prey dynamics~\citep{lotka1910volterra}. We show that both Transformers and RNNs can in context learn such complex dynamics, and thus provide preliminary but concrete evidence on the potential for \textit{counterfactual story generation} that is important to scientific discovery and AI safety.

%% file: chapters/paper-chapters/acknowledgements.tex
\section*{Acknowledgements}
\label{chp:ack}
MM thanks Jonas Peters for valuable discussions on the concept of counterfactual \textit{reasoning} and the role of noise abduction. MM also thanks Hsiao-Ru Pan and Florent Draye for their input on identifiability and causal probing. MM acknowledges financial support from the \textit{Konrad-Adenauer-Stiftung}. \looseness=-1

%% file: chapters/checklist.tex
\section*{NeurIPS Paper Checklist}

\begin{enumerate}

\item {\bf Claims}
    \item[] Question: Do the main claims made in the abstract and introduction accurately reflect the paper's contributions and scope?
    \item[] Answer: \answerYes{} 
    \item[] Justification: The introduction provides a bullet point list of contributions with direct reference to the corresponding claims in the paper.
    \item[] Guidelines:
    \begin{itemize}
        \item The answer NA means that the abstract and introduction do not include the claims made in the paper.
        \item The abstract and/or introduction should clearly state the claims made, including the contributions made in the paper and important assumptions and limitations. A No or NA answer to this question will not be perceived well by the reviewers. 
        \item The claims made should match theoretical and experimental results, and reflect how much the results can be expected to generalize to other settings. 
        \item It is fine to include aspirational goals as motivation as long as it is clear that these goals are not attained by the paper. 
    \end{itemize}

\item {\bf Limitations}
    \item[] Question: Does the paper discuss the limitations of the work performed by the authors?
    \item[] Answer: \answerYes{} 
    \item[] Justification: Limitations are discussed in the final section.
    \item[] Guidelines:
    \begin{itemize}
        \item The answer NA means that the paper has no limitation while the answer No means that the paper has limitations, but those are not discussed in the paper. 
        \item The authors are encouraged to create a separate "Limitations" section in their paper.
        \item The paper should point out any strong assumptions and how robust the results are to violations of these assumptions (e.g., independence assumptions, noiseless settings, model well-specification, asymptotic approximations only holding locally). The authors should reflect on how these assumptions might be violated in practice and what the implications would be.
        \item The authors should reflect on the scope of the claims made, e.g., if the approach was only tested on a few datasets or with a few runs. In general, empirical results often depend on implicit assumptions, which should be articulated.
        \item The authors should reflect on the factors that influence the performance of the approach. For example, a facial recognition algorithm may perform poorly when image resolution is low or images are taken in low lighting. Or a speech-to-text system might not be used reliably to provide closed captions for online lectures because it fails to handle technical jargon.
        \item The authors should discuss the computational efficiency of the proposed algorithms and how they scale with dataset size.
        \item If applicable, the authors should discuss possible limitations of their approach to address problems of privacy and fairness.
        \item While the authors might fear that complete honesty about limitations might be used by reviewers as grounds for rejection, a worse outcome might be that reviewers discover limitations that aren't acknowledged in the paper. The authors should use their best judgment and recognize that individual actions in favor of transparency play an important role in developing norms that preserve the integrity of the community. Reviewers will be specifically instructed to not penalize honesty concerning limitations.
    \end{itemize}

\item {\bf Theory assumptions and proofs}
    \item[] Question: For each theoretical result, does the paper provide the full set of assumptions and a complete (and correct) proof?
    \item[] Answer: \answerYes{} 
    \item[] Justification: All statements in the main text are proven in the appendix. Lemmata in the appendix are either proven directly or accompanied by a reference to a proof.
    \item[] Guidelines:
    \begin{itemize}
        \item The answer NA means that the paper does not include theoretical results. 
        \item All the theorems, formulas, and proofs in the paper should be numbered and cross-referenced.
        \item All assumptions should be clearly stated or referenced in the statement of any theorems.
        \item The proofs can either appear in the main paper or the supplemental material, but if they appear in the supplemental material, the authors are encouraged to provide a short proof sketch to provide intuition. 
        \item Inversely, any informal proof provided in the core of the paper should be complemented by formal proofs provided in appendix or supplemental material.
        \item Theorems and Lemmas that the proof relies upon should be properly referenced. 
    \end{itemize}

    \item {\bf Experimental result reproducibility}
    \item[] Question: Does the paper fully disclose all the information needed to reproduce the main experimental results of the paper to the extent that it affects the main claims and/or conclusions of the paper (regardless of whether the code and data are provided or not)?
    \item[] Answer: \answerYes{} 
    \item[] Justification: Multiple appendix sections discuss model setup, training details and the synthetic dataset.
    \item[] Guidelines:
    \begin{itemize}
        \item The answer NA means that the paper does not include experiments.
        \item If the paper includes experiments, a No answer to this question will not be perceived well by the reviewers: Making the paper reproducible is important, regardless of whether the code and data are provided or not.
        \item If the contribution is a dataset and/or model, the authors should describe the steps taken to make their results reproducible or verifiable. 
        \item Depending on the contribution, reproducibility can be accomplished in various ways. For example, if the contribution is a novel architecture, describing the architecture fully might suffice, or if the contribution is a specific model and empirical evaluation, it may be necessary to either make it possible for others to replicate the model with the same dataset, or provide access to the model. In general. releasing code and data is often one good way to accomplish this, but reproducibility can also be provided via detailed instructions for how to replicate the results, access to a hosted model (e.g., in the case of a large language model), releasing of a model checkpoint, or other means that are appropriate to the research performed.
        \item While NeurIPS does not require releasing code, the conference does require all submissions to provide some reasonable avenue for reproducibility, which may depend on the nature of the contribution. For example
        \begin{enumerate}
            \item If the contribution is primarily a new algorithm, the paper should make it clear how to reproduce that algorithm.
            \item If the contribution is primarily a new model architecture, the paper should describe the architecture clearly and fully.
            \item If the contribution is a new model (e.g., a large language model), then there should either be a way to access this model for reproducing the results or a way to reproduce the model (e.g., with an open-source dataset or instructions for how to construct the dataset).
            \item We recognize that reproducibility may be tricky in some cases, in which case authors are welcome to describe the particular way they provide for reproducibility. In the case of closed-source models, it may be that access to the model is limited in some way (e.g., to registered users), but it should be possible for other researchers to have some path to reproducing or verifying the results.
        \end{enumerate}
    \end{itemize}

\item {\bf Open access to data and code}
    \item[] Question: Does the paper provide open access to the data and code, with sufficient instructions to faithfully reproduce the main experimental results, as described in supplemental material?
    \item[] Answer: \answerYes{} 
    \item[] Justification: Anonymized code is provided in the supplemental material.
    \item[] Guidelines:
    \begin{itemize}
        \item The answer NA means that paper does not include experiments requiring code.
        \item Please see the NeurIPS code and data submission guidelines (\url{https://nips.cc/public/guides/CodeSubmissionPolicy}) for more details.
        \item While we encourage the release of code and data, we understand that this might not be possible, so “No” is an acceptable answer. Papers cannot be rejected simply for not including code, unless this is central to the contribution (e.g., for a new open-source benchmark).
        \item The instructions should contain the exact command and environment needed to run to reproduce the results. See the NeurIPS code and data submission guidelines (\url{https://nips.cc/public/guides/CodeSubmissionPolicy}) for more details.
        \item The authors should provide instructions on data access and preparation, including how to access the raw data, preprocessed data, intermediate data, and generated data, etc.
        \item The authors should provide scripts to reproduce all experimental results for the new proposed method and baselines. If only a subset of experiments are reproducible, they should state which ones are omitted from the script and why.
        \item At submission time, to preserve anonymity, the authors should release anonymized versions (if applicable).
        \item Providing as much information as possible in supplemental material (appended to the paper) is recommended, but including URLs to data and code is permitted.
    \end{itemize}

\item {\bf Experimental setting/details}
    \item[] Question: Does the paper specify all the training and test details (e.g., data splits, hyperparameters, how they were chosen, type of optimizer, etc.) necessary to understand the results?
    \item[] Answer: \answerYes{} 
    \item[] Justification: Technical details and experimental setup for both settings are discussed in the appendix.
    \item[] Guidelines:
    \begin{itemize}
        \item The answer NA means that the paper does not include experiments.
        \item The experimental setting should be presented in the core of the paper to a level of detail that is necessary to appreciate the results and make sense of them.
        \item The full details can be provided either with the code, in appendix, or as supplemental material.
    \end{itemize}

\item {\bf Experiment statistical significance}
    \item[] Question: Does the paper report error bars suitably and correctly defined or other appropriate information about the statistical significance of the experiments?
    \item[] Answer: \answerYes{} 
    \item[] Justification: Where applicable, all plots are supported by basic bootstrap errorbars. For readability, some plots containing error bars are deferred to the appendix, with the figure in the main text missing error bars. The reader is explicitly referred to the appendix in these cases.
    \item[] Guidelines:
    \begin{itemize}
        \item The answer NA means that the paper does not include experiments.
        \item The authors should answer "Yes" if the results are accompanied by error bars, confidence intervals, or statistical significance tests, at least for the experiments that support the main claims of the paper.
        \item The factors of variability that the error bars are capturing should be clearly stated (for example, train/test split, initialization, random drawing of some parameter, or overall run with given experimental conditions).
        \item The method for calculating the error bars should be explained (closed form formula, call to a library function, bootstrap, etc.)
        \item The assumptions made should be given (e.g., Normally distributed errors).
        \item It should be clear whether the error bar is the standard deviation or the standard error of the mean.
        \item It is OK to report 1-sigma error bars, but one should state it. The authors should preferably report a 2-sigma error bar than state that they have a 96\% CI, if the hypothesis of Normality of errors is not verified.
        \item For asymmetric distributions, the authors should be careful not to show in tables or figures symmetric error bars that would yield results that are out of range (e.g. negative error rates).
        \item If error bars are reported in tables or plots, The authors should explain in the text how they were calculated and reference the corresponding figures or tables in the text.
    \end{itemize}

\item {\bf Experiments compute resources}
    \item[] Question: For each experiment, does the paper provide sufficient information on the computer resources (type of compute workers, memory, time of execution) needed to reproduce the experiments?
    \item[] Answer: \answerYes{} 
    \item[] Justification: This can be found under the training and model details in the appendix.
    \item[] Guidelines:
    \begin{itemize}
        \item The answer NA means that the paper does not include experiments.
        \item The paper should indicate the type of compute workers CPU or GPU, internal cluster, or cloud provider, including relevant memory and storage.
        \item The paper should provide the amount of compute required for each of the individual experimental runs as well as estimate the total compute. 
        \item The paper should disclose whether the full research project required more compute than the experiments reported in the paper (e.g., preliminary or failed experiments that didn't make it into the paper). 
    \end{itemize}
    
\item {\bf Code of ethics}
    \item[] Question: Does the research conducted in the paper conform, in every respect, with the NeurIPS Code of Ethics \url{https://neurips.cc/public/EthicsGuidelines}?
    \item[] Answer: \answerYes{} 
    \item[] Justification: This study does not involve human research subjects and experiments are run on a synthetically generated dataset. Therefore, data security, copyright or biases are not imperiled here.
    \item[] Guidelines:
    \begin{itemize}
        \item The answer NA means that the authors have not reviewed the NeurIPS Code of Ethics.
        \item If the authors answer No, they should explain the special circumstances that require a deviation from the Code of Ethics.
        \item The authors should make sure to preserve anonymity (e.g., if there is a special consideration due to laws or regulations in their jurisdiction).
    \end{itemize}

\item {\bf Broader impacts}
    \item[] Question: Does the paper discuss both potential positive societal impacts and negative societal impacts of the work performed?
    \item[] Answer: \answerYes{} 
    \item[] Justification: This paper contains a \textit{Broader impacts} paragraph discussing the consequences of potentially malicious use.
    \item[] Guidelines:
    \begin{itemize}
        \item The answer NA means that there is no societal impact of the work performed.
        \item If the authors answer NA or No, they should explain why their work has no societal impact or why the paper does not address societal impact.
        \item Examples of negative societal impacts include potential malicious or unintended uses (e.g., disinformation, generating fake profiles, surveillance), fairness considerations (e.g., deployment of technologies that could make decisions that unfairly impact specific groups), privacy considerations, and security considerations.
        \item The conference expects that many papers will be foundational research and not tied to particular applications, let alone deployments. However, if there is a direct path to any negative applications, the authors should point it out. For example, it is legitimate to point out that an improvement in the quality of generative models could be used to generate deepfakes for disinformation. On the other hand, it is not needed to point out that a generic algorithm for optimizing neural networks could enable people to train models that generate Deepfakes faster.
        \item The authors should consider possible harms that could arise when the technology is being used as intended and functioning correctly, harms that could arise when the technology is being used as intended but gives incorrect results, and harms following from (intentional or unintentional) misuse of the technology.
        \item If there are negative societal impacts, the authors could also discuss possible mitigation strategies (e.g., gated release of models, providing defenses in addition to attacks, mechanisms for monitoring misuse, mechanisms to monitor how a system learns from feedback over time, improving the efficiency and accessibility of ML).
    \end{itemize}
    
\item {\bf Safeguards}
    \item[] Question: Does the paper describe safeguards that have been put in place for responsible release of data or models that have a high risk for misuse (e.g., pre-trained language models, image generators, or scraped datasets)?
    \item[] Answer: \answerNA{} 
    \item[] Justification: This paper uses a synthetically generated dataset.
    \item[] Guidelines:
    \begin{itemize}
        \item The answer NA means that the paper poses no such risks.
        \item Released models that have a high risk for misuse or dual-use should be released with necessary safeguards to allow for controlled use of the model, for example by requiring that users adhere to usage guidelines or restrictions to access the model or implementing safety filters. 
        \item Datasets that have been scraped from the Internet could pose safety risks. The authors should describe how they avoided releasing unsafe images.
        \item We recognize that providing effective safeguards is challenging, and many papers do not require this, but we encourage authors to take this into account and make a best faith effort.
    \end{itemize}

\item {\bf Licenses for existing assets}
    \item[] Question: Are the creators or original owners of assets (e.g., code, data, models), used in the paper, properly credited and are the license and terms of use explicitly mentioned and properly respected?
    \item[] Answer: \answerYes{} 
    \item[] Justification: The existing asset used is an established open-source codebase. The paper recognizes the authors of that codebase.
    \item[] Guidelines:
    \begin{itemize}
        \item The answer NA means that the paper does not use existing assets.
        \item The authors should cite the original paper that produced the code package or dataset.
        \item The authors should state which version of the asset is used and, if possible, include a URL.
        \item The name of the license (e.g., CC-BY 4.0) should be included for each asset.
        \item For scraped data from a particular source (e.g., website), the copyright and terms of service of that source should be provided.
        \item If assets are released, the license, copyright information, and terms of use in the package should be provided. For popular datasets, \url{paperswithcode.com/datasets} has curated licenses for some datasets. Their licensing guide can help determine the license of a dataset.
        \item For existing datasets that are re-packaged, both the original license and the license of the derived asset (if it has changed) should be provided.
        \item If this information is not available online, the authors are encouraged to reach out to the asset's creators.
    \end{itemize}

\item {\bf New assets}
    \item[] Question: Are new assets introduced in the paper well documented and is the documentation provided alongside the assets?
    \item[] Answer: \answerNA{} 
    \item[] Justification: This is an experimental study on a synthetic dataset.
    \item[] Guidelines:
    \begin{itemize}
        \item The answer NA means that the paper does not release new assets.
        \item Researchers should communicate the details of the dataset/code/model as part of their submissions via structured templates. This includes details about training, license, limitations, etc. 
        \item The paper should discuss whether and how consent was obtained from people whose asset is used.
        \item At submission time, remember to anonymize your assets (if applicable). You can either create an anonymized URL or include an anonymized zip file.
    \end{itemize}

\item {\bf Crowdsourcing and research with human subjects}
    \item[] Question: For crowdsourcing experiments and research with human subjects, does the paper include the full text of instructions given to participants and screenshots, if applicable, as well as details about compensation (if any)? 
    \item[] Answer: \answerNA{} 
    \item[] Justification: The paper does not involve crowdsourcing nor research with human subjects.
    \item[] Guidelines:
    \begin{itemize}
        \item The answer NA means that the paper does not involve crowdsourcing nor research with human subjects.
        \item Including this information in the supplemental material is fine, but if the main contribution of the paper involves human subjects, then as much detail as possible should be included in the main paper. 
        \item According to the NeurIPS Code of Ethics, workers involved in data collection, curation, or other labor should be paid at least the minimum wage in the country of the data collector. 
    \end{itemize}

\item {\bf Institutional review board (IRB) approvals or equivalent for research with human subjects}
    \item[] Question: Does the paper describe potential risks incurred by study participants, whether such risks were disclosed to the subjects, and whether Institutional Review Board (IRB) approvals (or an equivalent approval/review based on the requirements of your country or institution) were obtained?
    \item[] Answer: \answerNA{} 
    \item[] Justification: The paper does not involve crowdsourcing nor research with human subjects.
    \item[] Guidelines:
    \begin{itemize}
        \item The answer NA means that the paper does not involve crowdsourcing nor research with human subjects.
        \item Depending on the country in which research is conducted, IRB approval (or equivalent) may be required for any human subjects research. If you obtained IRB approval, you should clearly state this in the paper. 
        \item We recognize that the procedures for this may vary significantly between institutions and locations, and we expect authors to adhere to the NeurIPS Code of Ethics and the guidelines for their institution. 
        \item For initial submissions, do not include any information that would break anonymity (if applicable), such as the institution conducting the review.
    \end{itemize}

\item {\bf Declaration of LLM usage}
    \item[] Question: Does the paper describe the usage of LLMs if it is an important, original, or non-standard component of the core methods in this research? Note that if the LLM is used only for writing, editing, or formatting purposes and does not impact the core methodology, scientific rigorousness, or originality of the research, declaration is not required.
    \item[] Answer: \answerNA{} 
    \item[] Justification: No LLMs are used for research-related purposes.
    \item[] Guidelines:
    \begin{itemize}
        \item The answer NA means that the core method development in this research does not involve LLMs as any important, original, or non-standard components.
        \item Please refer to our LLM policy (\url{https://neurips.cc/Conferences/2025/LLM}) for what should or should not be described.
    \end{itemize}

\end{enumerate}

\clearpage 

%% file: chapters/appendix-posterior-predictive.tex
\section{Posterior predictive distribution}
\begin{proof}[Proof of \textit{posterior predictive distribution}]
Let $\Theta \subset \R, \Beta \subseteq \R$ and $\beta \in \Beta$. Define the observational data $\mathbf{x} = \contextOBS$ for fixed $n$. Then, with 
$\mathrm{query} := (x_1,y_1,...,x_n,y_n,z,x^\text{CF})$,
    \begin{align}
        &p_{\YZCF}(\yzCF | \mathrm{query}) \nonumber \\
        &= \int_\Theta p_{\YZCF}(\yzCF | \mathrm{query}, \theta) \pi(\theta | \mathrm{query} ) \dd \theta \nonumber \\
        &= \int_\Theta p_{\YZCF}(\yzCF | \mathbf{x}, \theta, \xCF,z) \pi(\theta | \mathbf{x}, \xCF,z ) \dd \theta \nonumber\\
        &\overset{\ind}{=} \int_\Theta p_{\YZCF}(\yzCF | \mathbf{x}, \theta, \xCF,z) \pi(\theta | \mathbf{x}) \dd \theta \label{equiv:independence} \\
        &= \int_\Theta \int_\Beta p_{\YZCF}(\beta (\xCF - \xz) + \yz | \mathbf{x}, \theta, \xCF,z, \beta) \pbetadot(\beta | \mathbf{x}, \theta, \xCF, z) \pi(\theta | \mathbf{x}) \dd \beta \dd \theta \nonumber \\
        &\overset{\text{dF}}{=} \int_\Theta \int_\Beta p_{\YZCF}(\beta (\xCF - \xz) + \yz | \xz, \yz, \theta, \xCF,z, \beta) \pbetadot(\beta | \mathbf{x}, \theta) \pi(\theta | \mathbf{x}) \dd \beta \dd \theta \label{equiv:exch} \\
        &= \int_\Theta \int_\Beta \deltaYZCF{\beta \xCF + \yz - \beta \xz} \pbetadot(\beta | \mathbf{x}, \theta) \pi(\theta | \mathbf{x} ) \dd \beta \dd \theta, \nonumber
    \end{align}
where \eqref{equiv:independence} follows from independence, as $\theta \ind (\XCF,Z)$, and \eqref{equiv:exch} by de Finetti~\citep{deFinetti1931FunzioneAleatorio,klenke2008}, as each sequence $\{(x_1,y_1),...,(x_n,y_n)\}$ forms an exchangeable unit.
\end{proof}

\begin{proof}[Mean and variance in linear additive framework]
Both follow immediately by the tower law $\left(\symrook\right)$ of iterated expectations,
    \begin{align*}
        \E[\YCF] &= \E[\beta \XCF + U_Y] \\
        &\overset{\symrook}{=} \E\left[\E\left[\beta|\theta\right]\right] \E[\XCF] + \E \left[\E \left[U_Y|\theta \right] \right] \\
        &= \E[\theta] = 0 \\
        \Var(\YCF) &= \E[(\YCF)^2] = \E\left[\left(\beta \XCF + U_Y \right)^2 \right] \\
        &= \E\left[\left(\beta \XCF \right)^2 + 2 \beta \XCF U_Y + U_Y^2 \right] \\
        &\overset{\ind}{=} \E\left[\beta^2\right] \E\left[(\XCF)^2\right] + 2 \E\left[\beta U_Y\right] \E\left[\XCF\right]+ \E\left[U_Y^2 \right] \\
        &\overset{\symrook}{=} \E\left[\E\left[\beta^2|\theta\right]\right] \Var\left(\XCF\right) + \E\left[\E\left[U_Y^2 | \theta \right] \right] \\
        &= \E\left[1 + \theta^2\right] (\Var \left(\XCF \right) + 1) \\
        &= (1 + \Var(\theta)) (12 + 1) = 13^2 = 169.
    \end{align*}
The proof for the variance under the multiplicative extension, $\Var(\beta \XCF U_Y)$, is analogous.
\end{proof}

%% file: chapters/appendix-proofs.tex
\section{Transformation lemma and identifiability}
\label{app:pathexp}

\subsection{Transformation lemma}
\begin{proof}[Proof of Lemma \ref{lmm:retrieval}]
As $T$ is invertible in $u$ given $f(x)$,
\begin{align*}
    \yCF = T(f(\xCF), u) = T \left(f(\xCF), \inv{T}(f(x), y) \right).
\end{align*}
We require injectivity in $u$ to uniquely determine $u$ from $(f(x),y)$. Else, $\yCF = T(f(\xCF),u)$ is ambiguous and counterfactuals are ill-defined. Thus, the counterfactual completion writes
\begin{align*}
    \YCF = h \left( f(\xCF), f(x), y \right)
\end{align*}
for $h: \outputclass \times \outputclass \times \outputclass \longrightarrow \outputclass$.
\end{proof}

\subsection{Counterfactual identifiability of bijective causal models}
To prove Theorem \ref{thm:identifiability}, we restate Definition 6.1, Proposition 6.2, Lemma B.2 and Theorem 5.1 from~\citet{nasr2023counterfactualidentifiability}. For brevity, the proofs are omitted. We begin by introducing the \textit{Bijective Generation Mechanism (BGM)} in our notation. Let $\{X_1,...,X_d\}$ be a sequence of $d$ endogenous random variables. Each $X_j, j \in [d]$ is generated by
\begin{align*}
    X_j := f_j \left( \Pa(X_j), U_j \right).
\end{align*}
Importantly, $f_j$ is a bijective mapping from $U_j$ to $X_j$ for each realization of $\Pa(X_j)$ with $\Pa(X_j)$ denoting the causal parents of $X_j$. In the bivariate case, we write
\begin{align*}
    Y := f \left( \Pa(Y), U_Y \right) = f (X, U_Y)
\end{align*}
for $f$ bijective, as described in Lemma~\ref{lmm:retrieval}. Thus, with causal graph $X \longrightarrow Y$, we have $\Pa(Y) = \{X\}, \Pa(X) = \emptyset$.

\begin{definition}[Definition 6.1, Equivalence,~\citet{nasr2023counterfactualidentifiability}]
\label{def:nasrD6.1}
BGMs $f_1$ and $f_2$ are equivalent iff there exists an invertible function $g(\cdot)$ such that
\begin{align*}
    \forall x,y:& f_1^{-1}(x, y) = g\Big(f_2^{-1}(x, y)\Big)\text{ or alternatively}\\
    \forall x,u:& f_1(x, u) = f_2\Big(x, g^{-1}(u)\Big).
\end{align*}
\end{definition}

\begin{proposition}[Proposition 6.2,~\citet{nasr2023counterfactualidentifiability}]
\label{prop:6.2}
    BGMs $f_1, f_2$ produce the same counterfactuals $\iff f_1, f_2$ are equivalent BGMs.
\end{proposition}

\begin{lemma}[Lemma B.2,~\citet{nasr2023counterfactualidentifiability}]
\label{lmm:nasrB.2}
    BGMs $f$ and $\hat{f}$ that produce the same distribution $P_{\mathcal{D}}(X, Y)$ are equivalent if
    \begin{enumerate}
        \item (Markovian) $U \ind X$ and $\hat{U} \ind X$.
        \item for all $x$, $f(x, \cdot)$ and $\hat{f}(x, \cdot)$ are either strictly increasing or strictly decreasing functions.
    \end{enumerate}
\end{lemma}

\begin{theorem}[Theorem 5.1,~\citet{nasr2023counterfactualidentifiability}]
\label{thm:nasr5.1}
    BGM $f$ is counterfactually identifiable given $\mathbb{P}_{X, Y}$ if
    \begin{enumerate}
        \item (Markovian) $U \ind X$.
        \item for all $x$, $f(x, \cdot)$ is either a strictly increasing or a strictly decreasing function.
    \end{enumerate}
\end{theorem}

\subsection{Prove counterfactual identifiability under exchangeability}
\begin{theorem*}[Counterfactual identifiability under exchangeability,~\autoref{thm:identifiability}]
    Let $X, Y, U, \theta$ scalar random variables with $X \ind U | \theta$. Take $T: \outputclass \times \noiseclass \longrightarrow \outputclass$ with $y = T(f(x),u)$. Assume $\forall f(x) \in \mathcal{Y}$, the inverse $T^{-1}(f(x),\cdot)$ exists for all $y$, i.e., $u = T^{-1}(f(x),y)$, and $\forall f(x) \in \mathcal{Y}$, $T(f(x), \cdot)$ is continuous. Suppose further that $f(x)$ continuous $\forall x$, and strictly monotonic in $x$. Then, set $Y = T(f(X),U)$. Given the joint law $\mathbb{P}_{X,Y|\theta}$, $T$ is counterfactually identifiable.
\end{theorem*}

\begin{proof}[Proof of Theorem \ref{thm:identifiability}]
    We apply Lemma~\ref{lmm:nasrB.2} and Theorem~\ref{thm:nasr5.1}~\citep{nasr2023counterfactualidentifiability} to the case of conditional independence. The proof is analogous by taking $\mathbb{Q}_{X,Y} := \mathbb{P}_{X,Y|\theta}$. Clearly, the joint satisfies Markov conditional on the latent $\theta$ by design. Next by the inverse function theorem, $T(f(x),\cdot)$ is strictly monotonic, $\forall f(x)$, satisfying condition 2 in Theorem~\ref{thm:nasr5.1}. With $f(x)$ strictly monotonic, continuous, $T^\prime(x,\cdot) := T(f(x), \cdot)$ satisfies condition 2 for $x$.
\end{proof}

%% file: chapters/paper-chapters/appendix-training-details.tex
\input{chapters/appendix-training-details}

%% file: chapters/appendix-training-details.tex
\section{Training details}
\label{app:trainingdetails}
To start, we introduce notation $[\cdot] := \{1,...,\cdot\}$, for instance, $[n] := \{1,...,n\}$. We construct synthetic datasets with fixed noise. Conditional on a uniformly distributed latent parameter $\rvtheta_i \in \Theta$, we draw $n_i \sim \mathcal{U}(\{2,...,50\})$ observational data points with noise $\mU_X^{i;j}|\rvtheta_i, \mU_Y^{i;j}|\rvtheta_i \in \RE, j \in [n_i]$ and weights $\rvbeta_i|\rvtheta_i \in \RE, i \in [N \cdot B]$ to enforce \textit{non-i.i.d.} data. We choose $E = 5$. Next, we sample $N \cdot B$ data points with 
\begin{equation}
    \begin{aligned}
        \rvtheta &\sim \mathcal{U}([-6,6]^E) & \rvbeta | \rvtheta &\sim \mathcal{N}(\rvtheta,\mI_E) \\
        \mU|\rvtheta &\sim \mathcal{N}(\rvtheta,\mI_E) & \mXCF &\sim \sU([-6,6]^E)
    \end{aligned}
\label{eqn:dataset}
\end{equation}
for $N = 50'000$ training steps of batch size $B = 64$. We thus have sample size $N \cdot B$. We set
\begin{align*}
    \mY = \rvbeta \odot \mU_X + \mU_Y
\end{align*} 
with $\odot$ denoting element-wise products to have input and output embedding dimension agree without zero-padding. Including the indicator token $Z_b = z_b \cdot \one{E}, z_b \in \{1,...,n_b\}$, the corpus of queries then consists of $N$ batches of the form
\begin{align*}
    \{\left(\mathbf{x}_{b;1},\mathbf{y}_{b;1},...,\mathbf{x}_{b;n_b},\mathbf{y}_{b;n_b},\mathbf{z}_b,\xiCFdot{b}\right)\}_{b \in [B]}
\end{align*}
on which the model has to predict $\yiCFdot{b;z_b}, b \in [B]$. Our target $\YCF$ is zero-mean with $\Var(\YCF) = 169$ and $\log(\Var(\YCF)) = 5.1299$.

We train on minimizing the per-batch mean squared error $(\MSE)$ between the counterfactual prediction $\widehat{\yiCFdot{[B]}}$ and the ground truth $\yiCFdot{[B]}$,
\begin{align*}
    \MSE \left(\widehat{\yiCFdot{[B]}}, \yiCFdot{[B]} \right) = \frac{1}{B \cdot E} \sum_{b = 1}^B \left\|\widehat{\yiCFdot{b;z_b}} - \yiCFdot{b;z_b}\right\|_2^2
\end{align*}
and evaluate on an unseen test set following \eqref{eqn:dataset}.

Our code is based on the repository by \citet{linreg}. We therefore adopt the learning rate of $10^{-4}$ for all function classes and models but use the AdamW optimizer \citep{loshchilov2018decoupled} instead of Adam \citep{kingma2015adam}. We implement the experiments in \texttt{pytorch}~\citep{paszke2019pytorch} and use one NVIDIA GeForce RTX $3090$ GPU for training. The conducted experiments require between $5$ minutes and $2$ hours of training depending on model setup and task complexity.

%% file: chapters/appendix-model-details.tex
\section{Model details}
\label{app:modeldetails}
The Transformer architecture is described in Section \ref{chp:transformerarchitecture}. Here we lay out the details on the three recurrent models used and investigate the relevance of model depth in Transformers more closely.

\subsection{Model details on RNN architectures}

The Elman RNN~\citep{elman1990finding} is a foundational recurrent neural network architecture. It consists of an input layer, a hidden layer with recurrent connections, and an output layer. The hidden state at time step $t$, denoted by $h_t$, is computed as
\begin{align*}
    h_t &= \tanh(W_{ih}x_t + W_{hh}h_{t-1} + b_h) \\
    y_t &= W_{ho}h_t + b_o,
\end{align*}
where $x_t$ is the input at time $t$, and $\tanh$ denotes the \textit{tangens hyperbolicus}, a non-linear activation function. The output is given by $y_t$ and $W_\cdot$, $b_\cdot$ represent learnable weight matrices and biases.

The Long Short-Term Memory (LSTM) network~\citep{hochreiter1997lstm} addresses the vanishing gradient problem by incorporating a memory cell and three gates: input $(i_t)$, forget $(f_t)$, and output $(o_t)$. These gates regulate the flow of information and enable the network to retain long-term dependencies. The LSTM cell is defined by the following equations,
\begin{align*}
    f_t &= \sigma(W_f x_t + U_f h_{t-1} + b_f) \\
    i_t &= \sigma(W_i x_t + U_i h_{t-1} + b_i) \\
    o_t &= \sigma(W_o x_t + U_o h_{t-1} + b_o) \\
    g_t &= \tanh(W_g x_t + U_g h_{t-1} + b_g) \\
    c_t &= f_t \odot c_{t-1} + i_t \odot g_t \\
    h_t &= o_t \odot \tanh(c_t).
\end{align*}
In this formulation, $c_t$ is the cell state and $h_t$ the hidden output state. Activation vectors of the three gates are given by $i_t, f_t, o_t$ and $g_t$ is the cell input activation vector. By $\sigma$, we denote the sigmoid function and element-wise multiplication by $\odot$. $U_\cdot$ represents another weight matrix.

The Gated Recurrent Unit (GRU)~\citep{cho2014learning} introduces gating mechanisms to better control the flow of information. It simplifies the LSTM architecture by combining the forget and input gates into a single update gate, $z_t$. The GRU computes its hidden state using the following equations,
\begin{align*}
    r_t &= \sigma(W_r x_t + U_r h_{t-1} + b_z) \\
    z_t &= \sigma(W_z x_t + U_z h_{t-1} + b_r) \\
    n_t &= \tanh(W_n x_t + U_n (r_t \odot h_{t-1}) + b_n) \\
    h_t &= (1 - z_t) \odot n_t + z_t \odot h_{t-1}.
\end{align*}
Here, $r_t$ is the reset gate, $z_t$ the update gate, $n_t$ the new gate, the candidate update vector. Note that the dependence of the hidden state on the update gate varies across documentations. We follow the definition in the \texttt{pytorch} package. 

We perform a hyperparameter sweep on hidden size $D$ and the number of layers $L$ over the grid $D \in \{64, 128, 256\}, L \in \{1, 2\}$. We select the configuration with hidden size $256$ and $2$ layers for each of the three architectures.

\subsection{Copying in RNN architectures}
In general, we study whether neural sequence models are able to perform counterfactual prediction. We observe that Transformers do so, but also RNN-type sequence models achieve low error. \autoref{fig:performanceregression} displays this finding for \textsc{standard} GPT-2 Transformers as well as LSTMs, GRUs, and Elman RNNs. In their seminal paper, \citet{hochreiter1997lstm} lay out the copy memory task for their developed method, the LSTM. \citet{arjovsky2015unitary} address vanishing and exploding gradients by modifying RNNs and \citet{henaff2016recurrent} consider LSTMs and RNNs under longer contexts. Finally, \citet{feng2024rnnsneeded} test parallelizable modifications of LSTMs and GRUs on the Selective Copying Task~\citep{gu2024mambalineartimesequencemodeling}. Given this extensive literature, we restrict most of our analysis to autoregressive Transformers, as these represent the state-of-the-art architecture.

\subsection{Varying depth for additional values}
\begin{figure}[h]
    \centering
    \begin{subfigure}[t]{0.49\linewidth}
         \centering
         \includegraphics[width=\textwidth]{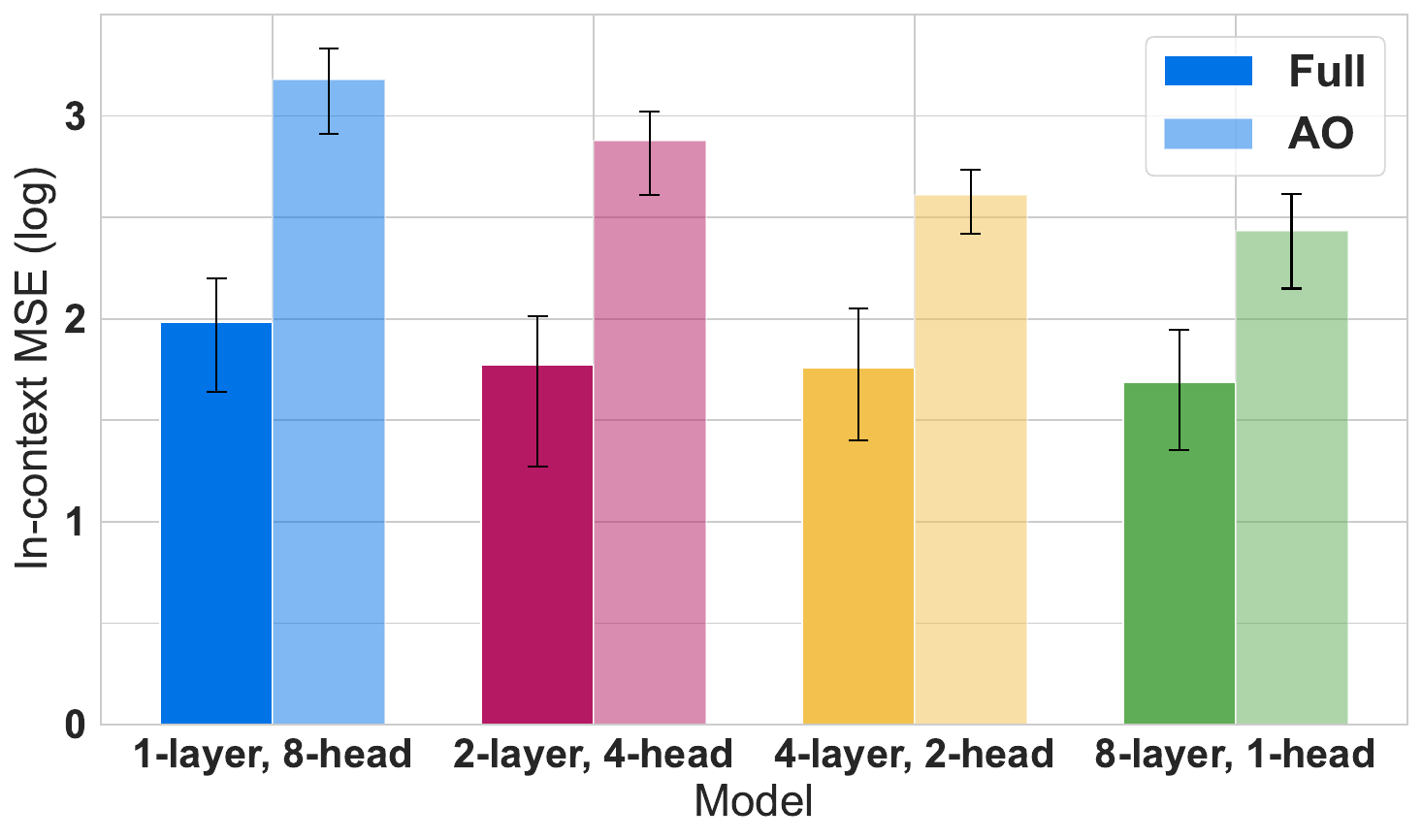}
         \caption{\textbf{Varying depth for two in-context examples.} \looseness=-1 }
         \label{fig:bar_chart_1_app}
     \end{subfigure}
     \hfill
     \begin{subfigure}[t]{0.49\linewidth}
         \centering
         \includegraphics[width=\textwidth]{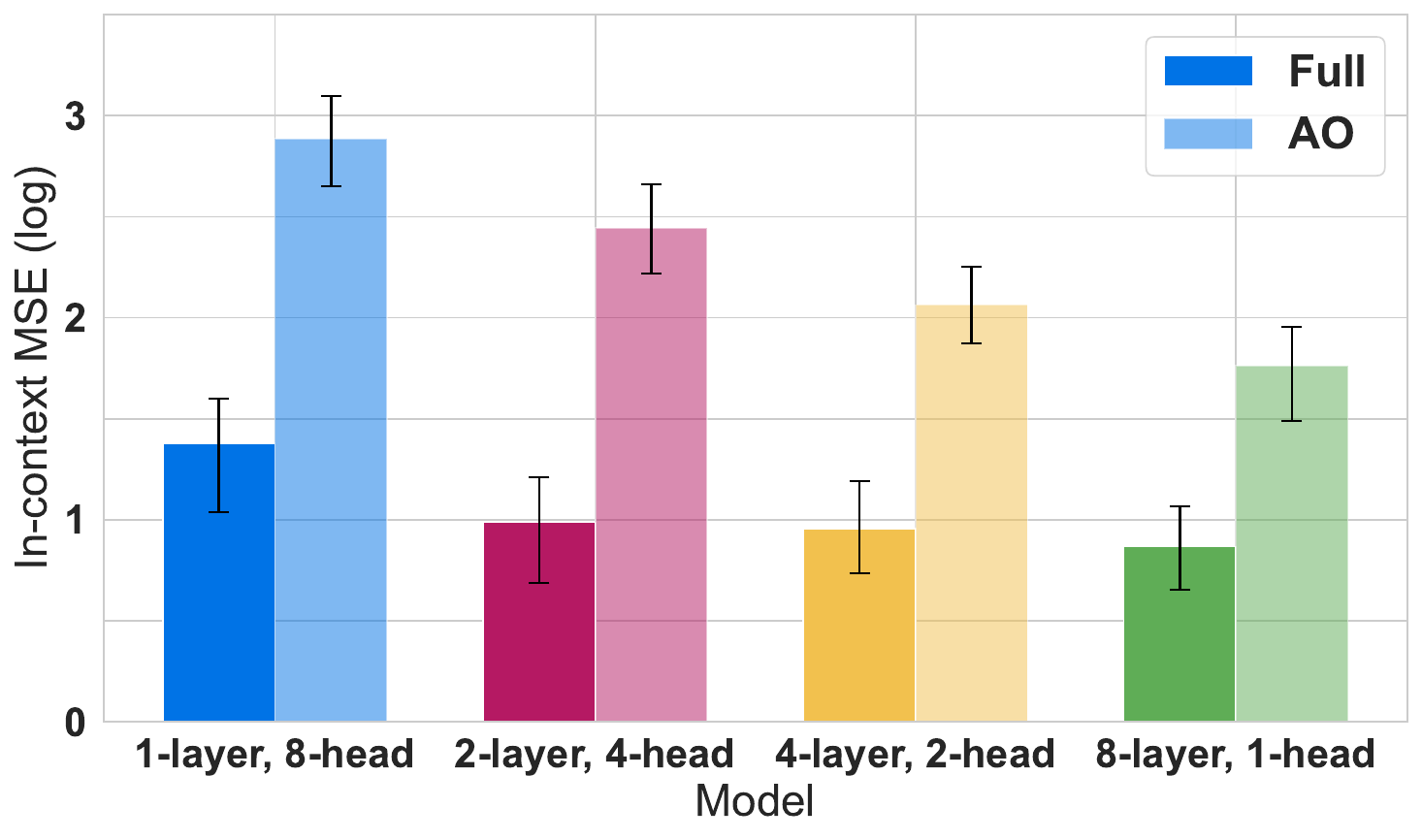}
         \caption{\textbf{Varying depth for forty-five in-context examples.}}
         \label{fig:bar_chart_50_app}
    \end{subfigure}
    \caption{\textbf{Varying depth more closely.} Building on top of the results displayed in \autoref{fig:bar_chart_depth}, we compare \textbf{Full} and \textbf{AO} Transformers at a constant number of $8$ attention heads. We observe a decrease in in-context loss as model depth increases across shorter and longer contexts of $2$ and $45$ in-context examples, respectively. We evaluate on $6400$ sequences.}
    \label{fig:bar_chart_2_45_app}
\end{figure}
Analogous to \autoref{fig:bar_chart_depth}, we note that increasing model depth leads to declining in-context $\MSE$. This holds for contexts both shorter and longer than the $35$-example version considered above. \autoref{fig:bar_chart_2_45_app} illustrates the in-context $\MSE$ for varying depth, evaluated at $2$ and $45$ in-context examples, respectively. We note that the loss of the $1$-layer, $8$-head \textbf{AO} Transformer exhibits no significant differences between $2$ in-context examples and $45$. We interpret this as indication that the model is unable to infer information on the latent $\rvtheta$ from the context. Reinforcing our findings that the $8$-layer, $1$-head Transformers achieves the lowest in-context $\MSE$, this pattern serves as evidence that in-context inference occurs across layers. A contrasting notion may be that the Transformer maps different junks of information to different subspaces of the embedding space. Acting on each subspace individually, each attention head would then focus on a different task before the $\MLP$ would combine the information and output the final prediction. Instead, the model appears to benefit from the interplay between layers, which supports the claim that attention heads across layers communicate through the residual stream~\citep{elhage2021transformercircuits}.

%% file: chapters/paper-chapters/appendix-diversity.tex
\input{chapters/appendix-diversity}
\input{chapters/appendix-garg}

%% file: chapters/appendix-diversity.tex
\section{Data diversity and robustness}
\label{app:ood-diversity}
\subsection{Data diversity adjustments and out-of-distribution}
In Section \ref{subsec:data_diversity_ood}, we discuss data diversity and generalizability to \textit{OOD}. The \textsc{uniform} setup largely follows the overall setup described in~\eqref{eqn:dataset} with the following adjustment: instead of $N \cdot B \cdot E$, we sample $d \ll N$ realizations of the latent and compute the effective support size~\citep{grendar2006effectivesupportsize}. Note that we control the absolute number of one-dimensional realizations as our setup effectively analyzes $E$ independent examples at every turn. Thus, we draw $d$ realizations and allocate across iterations and embedding dimensions. The parameter $d$ corresponds to the \texttt{diversity} argument in our code.

\begin{figure}[h]
    \centering
    \begin{subfigure}[t]{0.49\linewidth}
         \centering
         \includegraphics[width=\textwidth]{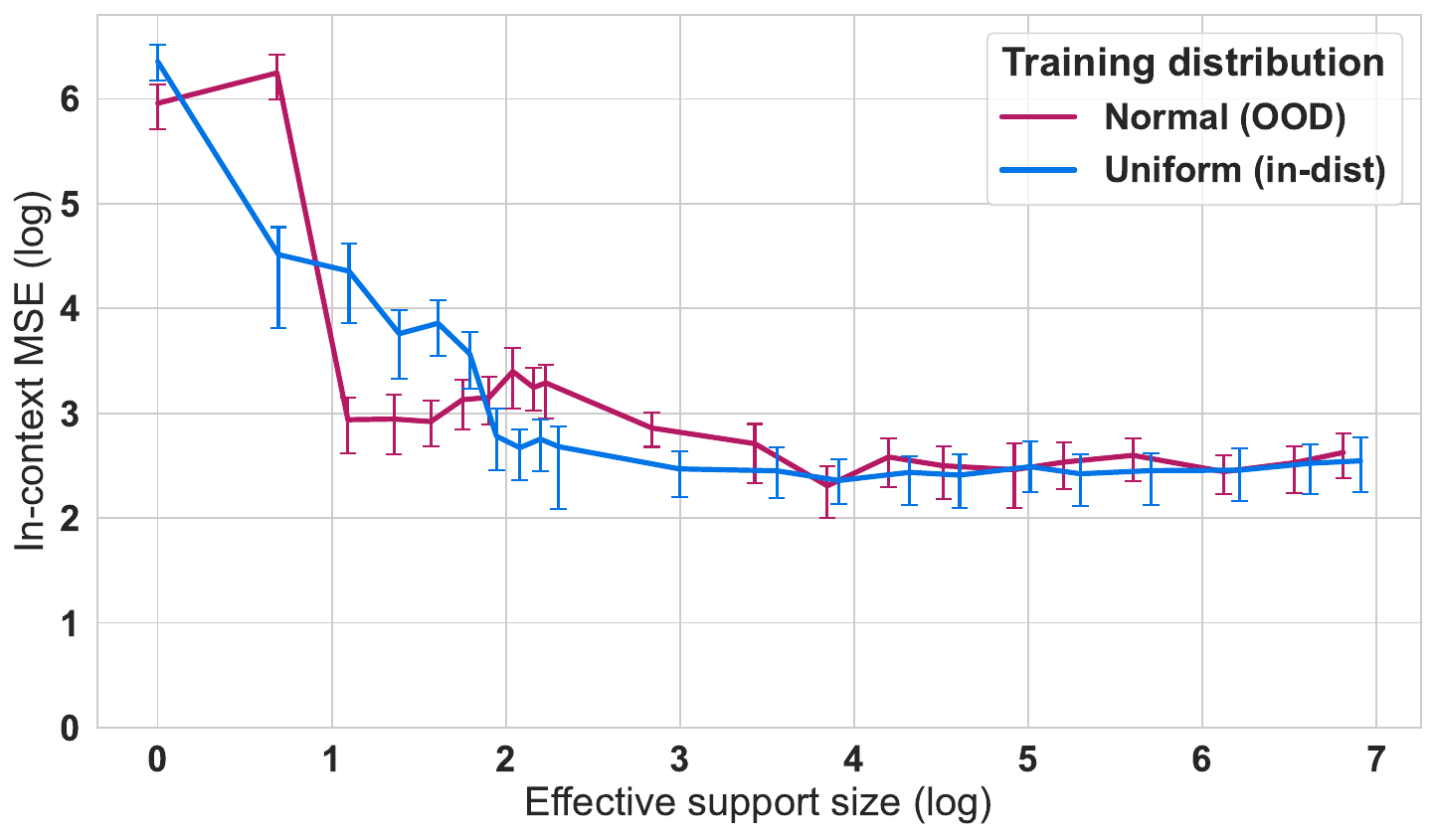}
         \caption{\textbf{Data diversity for evaluation on \textsc{uniform}.}}
         \label{fig:diversity_uniform_1}
     \end{subfigure}
     \hfill
     \begin{subfigure}[t]{0.49\linewidth}
         \centering
         \includegraphics[width=\textwidth]{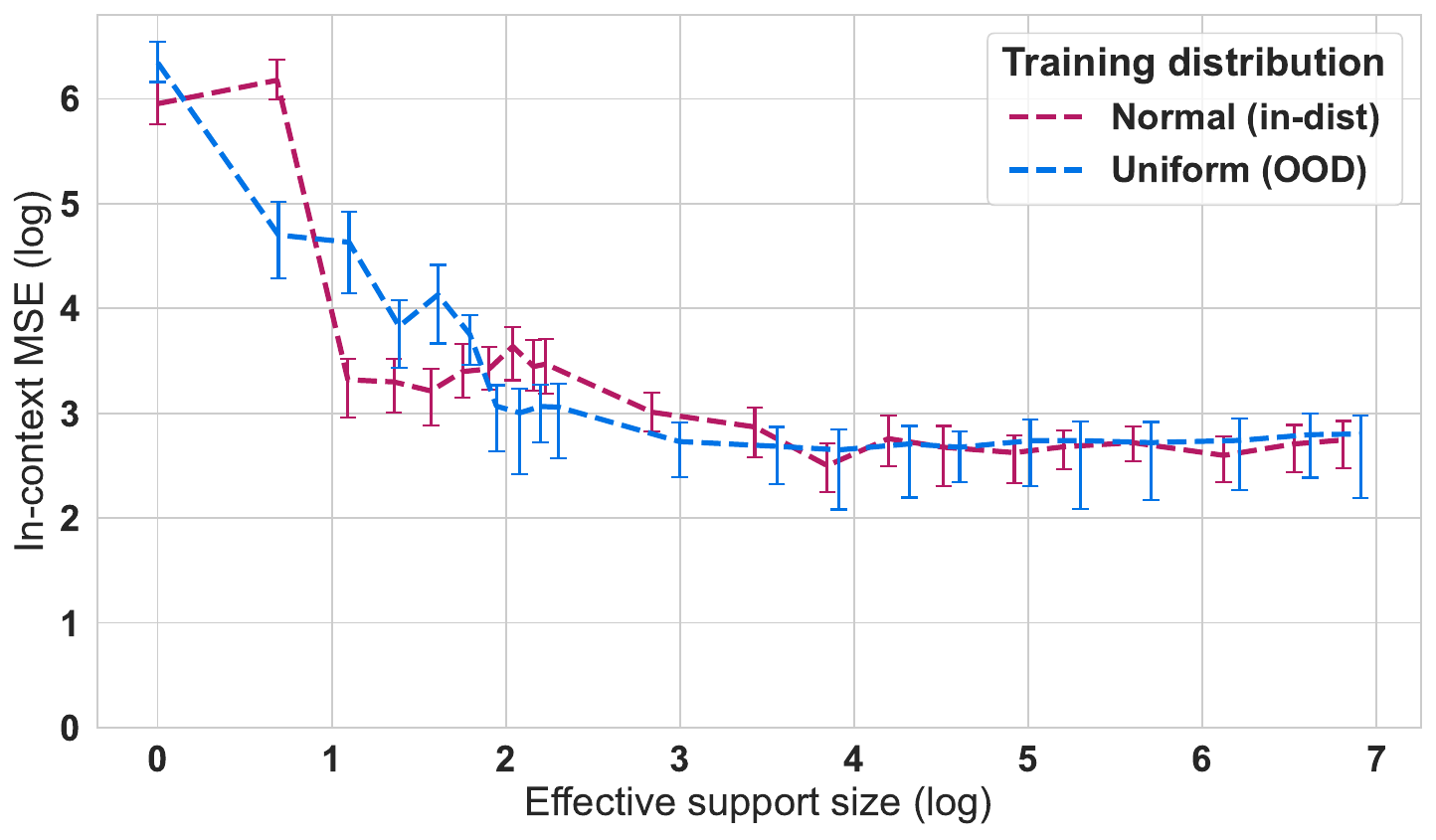}
         \caption{\textbf{Data diversity for evaluation on \textsc{normal}.}}
         \label{fig:diversity_normal_1}
    \end{subfigure}
    \caption{\textbf{Data diversity for pre-training on two in-context examples.} We measure in-context $\MSE$ (log-scaled) averaged over $6400$ prompts at two in-context examples against log-scaled effective support size. Each point represents one fully pre-trained model on either the \textsc{uniform} sampled or \textsc{normal} sampled $\theta$. We train the models for $50'000$ training steps. All models are evaluated on both datasets. Error bars are the $95\%$ basic bootstrap confidence intervals.}
    \label{fig:diversity_app}
\end{figure}

The \textsc{normal} setup follows the above structure. Compared to \eqref{eqn:dataset}, we sample the latent from $\rvtheta \sim \mathcal{N}(\mathbf{0},\sqrt{12} \cdot \mI_E)$ such that the theoretical variance is equivalent across setups. Everything else is analogous. Note that the effective support size is equivalent to the support size under the \textsc{uniform} setup. This can be seen by interpreting $\ESS$ as the logarithm of the Shannon entropy~\citep{shannon1948entropy}. Indeed, $p(\vartheta)$ is uniformly distributed over discrete $\Theta_0$ if $\vartheta$ denotes a realization of the latent $\theta$, \looseness=-1
\begin{align*}
    H_\text{\textsc{uniform}}(\theta) = - \sum_{\vartheta \in \Theta_0} \log p(\vartheta).
\end{align*}
Therefore, we allocate the $d = |\Theta_0|$ realizations uniformly across iterations and embedding dimensions. Under the \textsc{normal} setup, $p(\vartheta)$ depends on the distance of $\vartheta$ to $0$. Here, we sample $d$ Gaussian realizations. We then allocate them proportional to $p(\vartheta)$ across iterations and embedding dimensions. In analogy to~\autoref{fig:mse_vs_ess}, we plot the in-context $\MSE$ against $\ESS$ for pre-training on the \textsc{normal} setting and evaluating both in-distribution and on \textsc{uniform}. \autoref{fig:diversity_app} illustrates the in-context $\MSE$ relative to effective support size on contexts of two examples. Here we train all models for $50'000$ training steps. Although the loss is at a higher level than for $35$ in-context examples, we observe that the $\MSE$ declines with increasing effective support size. Moreover, when evaluated under the \textsc{normal} setup, the model converges more rapidly than under the \textsc{uniform} evaluation scheme. At $d = 3$, the Transformer already achieves performance comparable to that obtained when training on $1000$ unique realizations. For the latter case, the results remain consistent with the findings above, indicating that an effective pre-training support size of approximately $10$ is sufficient.

\subsection{Non-linear and non-additive extensions}
On top of the linear regression setting, we extend in-context counterfactual reasoning to non-linear, non-additive functions which are invertible in $u$. The general dataset setup remains the same with adjustments made only to the computation of $y, \yCF$. We evaluate the $8$-layer, $1$-head \textbf{Full} and \textbf{AO} Transformers. \autoref{tab:robustness} reports the in-context $\MSE$ on $6400$ queries with $35$ in-context examples each. Depending on the exact configuration of architecture and extension, the in-context $\MSE$ is at least one tenth of the empirical variance of the test set.

\begin{table}[H]
    \centering
    \begin{tblr}{|X[-1,l]|X[1.5,l]|X[1,c]X[1,c]X[1,c]|}
        \hline
          & $f(x,u)$ & Empirical variance & $\MSE$ of \textbf{AO} Transformer & $\MSE$ of \textbf{Full} Transformer \\[3pt]
         \hline
         Tanh & $\mathrm{tanh}\left(\tau(\beta x+u)\right)$ & 0.3404 & 0.0132 & 0.0058 \\[3pt]
         Sigmoid $(\sigma)$ & $\frac{1}{1+\exp\left(-\tau(\beta x + u)\right)}$ & 0.0387 & 0.0019 & 0.0009 \\[3pt]
         Multiplicative & $\frac{1}{\sqrt{\Var(\beta \XCF U_Y)}} \beta x  u$ & 0.9274 & 0.0668 & 0.0364 \\
         \hline
    \end{tblr}
    \caption{\textbf{Robustness to different function classes for regression tasks.} We report in-context $\MSE$ averaged over $6400$ sequences for the $8$-layer \textbf{AO} and \textbf{Full} Transformers under invertible nonlinear activation functions and multiplicative noise. We compute the empirical variance of target $\YCF$ on the test set. We observe for both architectures that the $\MSE$ is $10$ to $60$ times lower than the empirical variance, suggesting both can in-context learn more complex function classes beyond linear regression and complete the counterfactual query. All functions are applied element-wise in one dimension. $\tanh$ and $\sigma$ are scaled by $\tau = \frac{1}{13}$ to counteract congestion around $\{-1,1\}$ and $\{0,1\}$, respectively. To guarantee theoretical variance of $1$, we divide the multiplicative term by $\sqrt{\Var(\beta \XCF U_Y)} = \sqrt{3410.4}$.}
    \label{tab:robustness}
\end{table}

Note that under multiplicative noise, counterfactual reasoning reduces to a pure copying task as
\begin{align*}
    \YCF = \frac{1}{\sqrt{\Var(\YCF)}} \beta \XCF U_Y = \frac{1}{\sqrt{\Var(\YCF)}}\beta \XCF \frac{Y\sqrt{\Var(\YCF)}}{\beta X} = \frac{\XCF}{X}Y,
\end{align*}
where estimation of $\beta$ is not required for counterfactual completion.

\subsection{Data diversity in the literature}
Recent theoretical research on data diversity aligns closely with our observations. \citet{xiao2025rolediversityincontextlearning} develop a formal framework showing that diversity in example selection improves in‑context inference. With respect to model depth, \citet{petty2024impact} show that at least two attention layers are required to support compositional generalization beyond memorization. This is in line with our findings. Figures \ref{fig:bar_chart_depth} and \ref{fig:bar_chart_50_app} indicate that the additional benefit of deeper models decreases after having trained models of depth $2$. Improvements are especially high for $2$ layers relative to $1$. Intuitively, we relate this to the multi-step process of performing counterfactual prediction. The model retrieves the relevant pair, $(x_j, y_j)$ for $z = j$, infers noise, and predicts the counterfactual $\yCF$ given $\xCF$ (\autoref{fig:mainfigure}).

%% file: chapters/appendix-garg.tex
\subsection{Emergent learning of variable $z$ token}
\begin{figure}[h]
    \centering
    \begin{subfigure}{\linewidth}
        \centering
        \includegraphics[width=\linewidth]{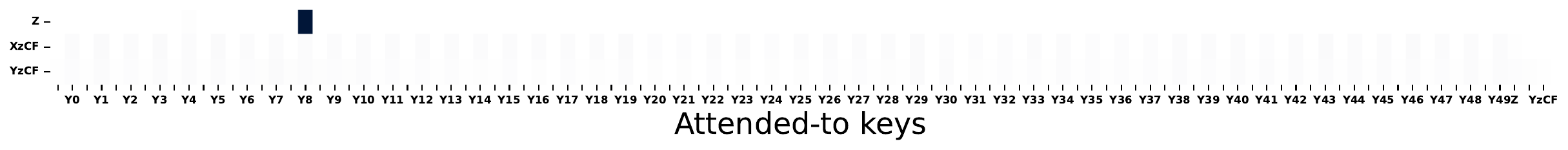}
    \end{subfigure}
    \vspace{1em}
    \begin{subfigure}{\linewidth}
        \centering
        \includegraphics[width=\linewidth]{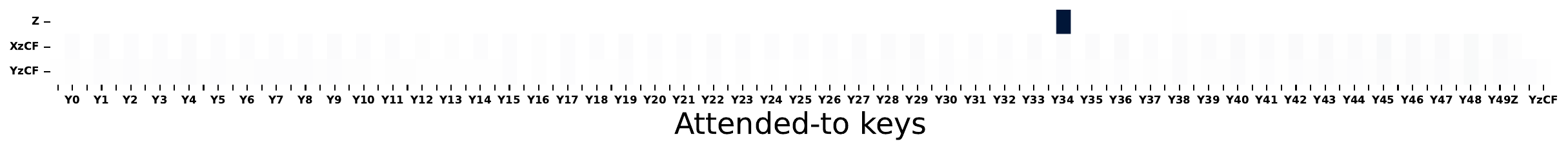}
    \end{subfigure}
    \vspace{1em}
    \begin{subfigure}{\linewidth}
        \centering
        \includegraphics[width=\linewidth]{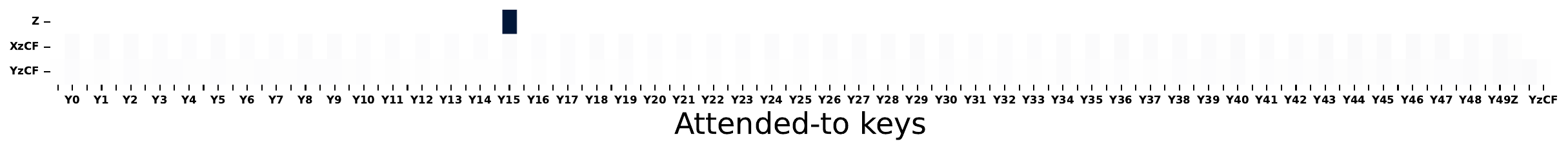}
    \end{subfigure}
    \vspace{1em}
    \begin{subfigure}{\linewidth}
        \centering
        \includegraphics[width=\linewidth]{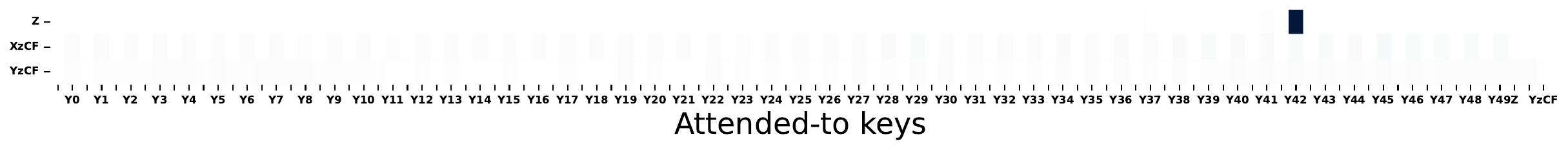}
    \end{subfigure}
    \vspace{1em}
    \begin{subfigure}[b]{\linewidth}
         \centering
         \includegraphics[width=\linewidth]{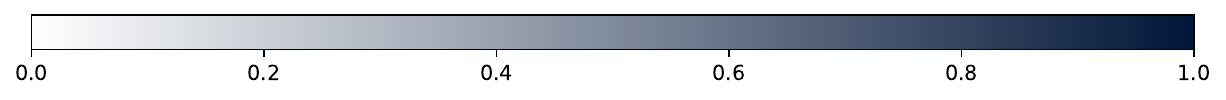}
    \end{subfigure}
    \caption{\textbf{Noise abduction head emerges in GPT-2 Transformer.} We plot $4$ representative noise abduction heads evaluated on sequences with $z = 8, z = 34, z = 15, z = 42$, respectively. Each graph represents the average over $64$ sequences with identical $z$ index and $50$ in-context pairs. We find that when processing embedding token $Z$, the model adapts to different values of $z$. \looseness=-1}
    \label{fig:attentionszindex}
\end{figure}

We confirm in~\autoref{fig:probing} that the model is capable of learning the latent parameter $\theta$ in context. We therefore find evidence that the model can learn the prior $\pi$ over the latent information as hypothesized in~\eqref{eqn:posteriorpredictive}. The final part toward in-context counterfactual reasoning pertains \textit{noise abduction}. Our setup is special as we include an embedding token $z$ which indicates the relevant in-context pair $(x_z, y_z)$ to abduct noise from. Only if the model can learn this connection on the fly, we can guarantee the three-step process of noise abduction, intervention, prediction. Indeed, we find that the \textbf{Full} GPT-2 Transformer with $12$ layers and $8$ heads trained for $1'000'000$ steps on $E = 1$ implements a \textit{noise abduction head} at the $7^\text{th}$ layer. We define a noise abduction head as one which attends to the relevant input-output pair $(x_z,y_z)$ when processing the $z$ token. This noise abduction head emerges at the sixth head of the seventh layer, and it attends to $y_z$ for variable realization of $z$. Visually inspecting the attention behavior of $100$ batches averaged over $64$ in-context sequences, we find that this arises in every single batch. Note that the $z$ token is constant within one batch. In~\autoref{fig:attentionszindex} we show the behavior for four different $z$ indices. For swift comparability, each prompt consists of $50$ in-context examples. We conclude that noise abduction, central to counterfactual reasoning, emerges in a designated attention head.

\subsection{Comparison to non-counterfactual regression}
\begin{figure}[h]
    \centering
    \begin{subfigure}[t]{0.49\linewidth}
         \centering
         \includegraphics[width=\textwidth]{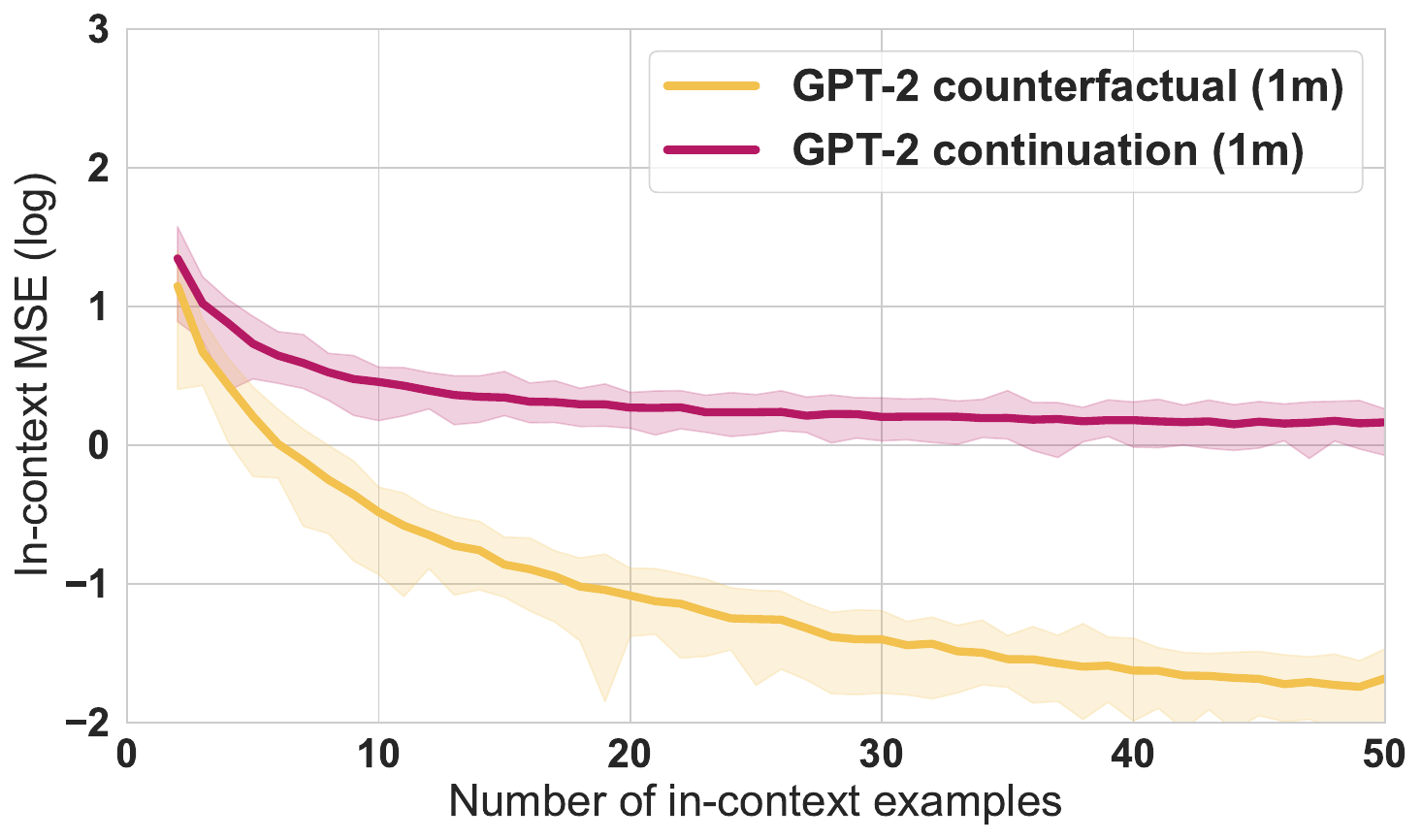}
         \caption{\textbf{GPT-2 performance.}}
         \label{fig:contmse}
     \end{subfigure}
     \hfill
     \begin{subfigure}[t]{0.49\linewidth}
         \centering
         \includegraphics[width=\textwidth]{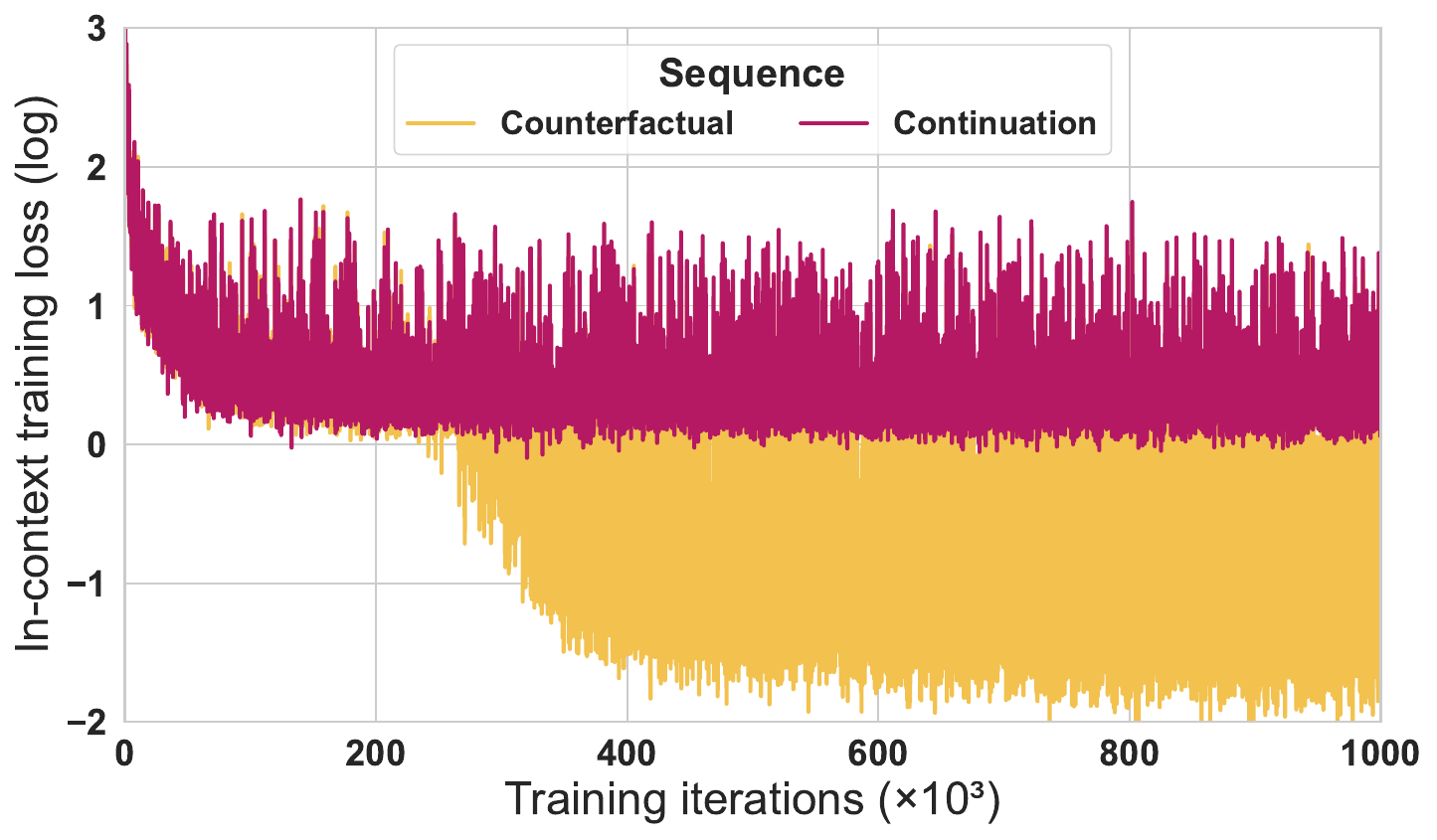}
         \caption{\textbf{{Phase transitions}.}}
         \label{fig:contphases}
    \end{subfigure}
    \caption{\textbf{Distinction between noise abduction and observational continuation.} We measure in-context $\MSE$ (log-scaled) averaged over $6400$ prompts for the \textbf{Full} GPT-2 Transformer. We observe significant improvements for contexts longer than $5$ examples. This we understand as indication of effective noise abduction in the model. We train the model for $1'000'000$ training steps and observe that the phase transition for the model trained on counterfactuals occurs after over $300'000$ steps. The logarithmic in-context training loss for the continuation framework remains at a higher level of above $0$. \looseness=-1}
    \label{fig:cont1mio}
\end{figure}

After reviewer feedback, we include a brief comparison of our results with training models on an observational continuation of the regression setup. Linear regression, as studied by~\citet{linreg}, does not require the ability to perform noise abduction. We take one step further by requiring the model to learn the functional form $f$ in context and to abduct the unobserved noise. To underscore our point, we train a \textbf{Full} GPT-2 architecture on $1'000'000$ training steps.~\autoref{fig:contmse} represents the relative performance improvement of the counterfactual sequence relative to the observational setup. We find that for over $5$ in-context examples the counterfactual $\MSE$ lies significantly below the estimated $\MSE$ of the observational sequence. In~\autoref{fig:contphases} we observe the phase transition to a significantly lower in-context training $\MSE$ occurs after $300'000$ steps. This is in line with our hypothesis that in-context counterfactual reasoning, in principle, leads to a lower prediction error than the standard regression setup~\citep{linreg} as the noise is fixed and can be inferred directly.

%% file: chapters/paper-chapters/appendix-cyclic.tex
\input{chapters/appendix-cyclic-details}
\input{chapters/appendix-cyclic-paper}

%% file: chapters/appendix-cyclic-details.tex
\section{Model details for cyclic sequential dynamical systems}
\label{app:cyclic}

\subsection{Lotka-Volterra model}
The Lotka-Volterra framework models the temporal evolution of the concentrations of two interacting species: predators and prey. In the absence of predators, the prey concentration grows continuously over time. Conversely, the predator concentration declines unless sustained by consuming prey. Predation enables predator growth, which simultaneously reduces prey concentration. We model the four individual dynamics by real, nonnegative parameters. The dynamics of the prey concentration~\eqref{eqn:preyconcentration} is parameterized by the intrinsic growth rate $\alpha$ and the rate at which the prey concentration decreases with predator concentration, $\beta$. For predator concentration~\eqref{eqn:predatorconcentration}, we capture the species-inherent decreasing concentration by $\gamma$ and the rate at which predators feed off prey by $\delta$. In the context of biological species, \textit{concentration} can be thought of as the population density.

\subsection{Bounds on Lotka-Volterra parameters}
Similar to the regression setup, we draw the model parameters conditional on the latent $\theta$. In order to overcome exploding concentrations, we adopt the variation-of-constants method to bound the parameters by controlling the maximum and minimum concentrations. This ODE-based approach we utilize as a guiding principle throughout. As stochasticity may lead to a case where the bounds are violated, however, we \textit{a posteriori} ensure evaluation on positive concentrations only. In the following, we adopt the ODE notation and write \eqref{eqn:preyconcentration} and \eqref{eqn:predatorconcentration} with lower-case variables as
\begin{align}
    \dv{x_t}{t} &= \alpha x_t - \beta x_t y_t, \label{eqn:lvx} \\
    \dv{y_t}{t} &= - \gamma y_t + \delta x_t y_t \label{eqn:lvy}
\end{align}
and observe that
\begin{align*}
    \dv{x_t}{t} - (\alpha - \beta y_t) x_t &= 0, \\
    \dv{y_t}{t} - (- \gamma + \delta x_t) y_t &= 0.
\end{align*}
For now, we confine ourselves to \eqref{eqn:lvx}. Dividing by $x_t$, we obtain
\begin{align}
\frac{\dd x_t / \dd t}{x_t} - (\alpha - \beta y_t) &= 0 \label{eqn:derivative0}
\end{align}
with $\frac{\dd x_t / \dd t}{x_t} = \dv{}{t} \log (x_t)$. By the fundamental theorem of calculus, integrating \eqref{eqn:derivative0} over $[0,t]$ yields
\begin{align*}
     \int_0^t \left( \alpha - \beta y_s \right) \dd s = \int_0^t \dv{}{s} \log (x_s) \dd s = \log (x_t) - \log (x_0).
\end{align*}
By similar reasoning the result follows for \eqref{eqn:lvy} such that we can reorganize
\begin{align}
    x_t &= x_0 \exp \left( \alpha t - \beta \int_0^t y_s \dd s \right), \label{eqn:Xtexp} \\
    y_t &= y_0 \exp \left( - \gamma t + \delta \int_0^t x_s \dd s \right). \label{eqn:Ytexp}
\end{align}

Now, we set $(\xlower, \xupper), (\ylower, \yupper)$ to be lower and upper bounds of $x_t, y_t$, respectively, $t \in [0, T]$. Then,
\begin{align*}
    \alpha t - \betaylwr t \geq \logxtx \geq \alpha t - \betayupr t
\end{align*}
such that 
\begin{equation*}
    \begin{aligned}
        \alpha &\geq \oneovert \logxtx + \betaylwr, & \oneovert \logxtx + \betayupr &\geq \alpha.
    \end{aligned}
\end{equation*}
As $\logxuprx \geq \logxtx \geq \logxlwrx, \forall t$,
\begin{align*}
    \alpha \in \left[ \oneovert \logxuprx + \betaylwr, \oneovert \logxlwrx + \betayupr \right].
\end{align*}
For $t \in [0, T]$, the bound for $ \lim_{t \searrow 0} \frac{1}{t} $ is trivial. We therefore decide to bound the final value at $T$ in $[\xlower, \xupper]$ such that
\begin{align*}
    \alpha \in \left[ \oneoverT \logxuprx + \betaylwr, \oneoverT \logxlwrx + \betayupr \right].
\end{align*}

To guarantee that the two bounds are in fact lower and upper bound, we can bound $\beta$,
\begin{align*}
    \oneoverT \logxuprx + \betaylwr &\leq \oneoverT \logxlwrx + \betayupr \\
    \oneoverT \loguprxlwrx &\leq \beta \left( \yupper - \ylower \right) \\
    \frac{1}{T \left( \yupper - \ylower \right)} \loguprxlwrx &\leq \beta.
\end{align*}
Note that stating an upper bound for $\beta$ violates the model assumption $\beta \geq 0$.

Similarly, we write for \eqref{eqn:Ytexp},
\begin{align*}
    - \gamma t + \deltaxlwr t &\leq \logyty \leq -\gamma t + \deltaxupr t \\
    - \gamma + \deltaxlwr \leq \oneovert \logylwry &\leq \oneovert \logyupry \leq - \gamma + \deltaxupr
\end{align*}
such that for $T > 0$,
\begin{align*}
    \gamma \in \left[ \deltaxlwr - \oneoverT \logylwry, \deltaxupr - \oneoverT \logyupry \right].
\end{align*}
We, too, establish the bound by enforcing that lower and upper bounds are in correct order, i.e.,
\begin{align*}
    \deltaxlwr - \oneoverT \logylwry &\leq \deltaxupr - \oneoverT \logyupry \\
    \oneoverT \left[ \logyupry - \logylwry \right] &\leq \delta \left( \xupper - \xlower \right) \\
    \frac{1}{T \left( \xupper - \xlower \right)} \loguprylwry &\leq \delta.
\end{align*}

Put together, we have the bounds
\begin{equation*}
    \begin{aligned}        
        \alphalwr &:= \oneoverT \logxuprx + \betaylwr &\leq \alpha &\leq \oneoverT \logxlwrx + \betayupr &=: \alphaupr, \\
        \betalwr &:= \frac{1}{T \left( \yupper - \ylower \right)} \loguprxlwrx &\leq \beta, & & \\
        \gammalwr &:= \deltaxlwr - \oneoverT \logylwry &\leq \gamma &\leq \deltaxupr - \oneoverT \logyupry &=: \gammaupr, \\
        \deltalwr &:= \frac{1}{T \left( \xupper - \xlower \right)} \loguprylwry &\leq \delta. & &
    \end{aligned}
\end{equation*}

\subsection{SDE Parameterization}
In accordance with our exchangeable setting, we construct the dataset with
\begin{equation}
    \begin{aligned}
        \rvtheta &\sim \mathcal{U}([1,2]^E), & 
        \mU|\rvtheta &\sim \mathcal{N}(\rvtheta,\mI_E), & \mXiCFdot{0}, \mYiCFdot{0} &\sim \sU([1,2]^E)
    \end{aligned}
\label{eqn:sdedataset}
\end{equation}
and bounds on the variables $(\xlower,\xupper) = (\ylower, \yupper) = (0.5, 2)$. We parameterize the Lotka-Volterra model,
\begin{equation*}
    \begin{aligned}
        \rvbeta - \betalwrbf &\sim \Exp(\rvtheta) &
        \rvdelta - \deltalwrbf &\sim \Exp(\rvtheta) \\
        \rvalpha &\sim \Uniform \left( [\alphalwrbf, \alphauprbf] \right) &
        \rvgamma &\sim \Uniform \left( [\gammalwrbf, \gammauprbf] \right). \\
    \end{aligned}
\end{equation*}
Finally, we sample the initial condition $(\mX_0, \mY_0)$ from a $\Betadist \left( \frac{1}{2 - \rvtheta}, 2 \right)$ distribution. Recall that the one-dimensional $\Betadist \left(\kappa, \lambda \right)$ on the measurable space $([0,1], \mathfrak{B}([0,1]))$ attains its theoretical mode at $\frac{\kappa - 1}{\kappa + \lambda - 2}$ for $\kappa, \lambda > 1$, $\mathfrak{B}$ the $\mathrm{Borel}$ $\sigma$-field. Here we deviate from the conventional parameterization $\mathrm{Beta}(\alpha, \beta)$ to avoid conflict with the Lotka–Volterra parameters. To enforce an initial condition depending on the latent concept, we choose $(\kappa, \lambda)$ such that the theoretical mode equals $\theta - 1$. Equip the measurable space $(\Theta, \mathfrak{B}(\Theta))$ with probability measure $\pi$ with $\mathrm{supp}(\pi) = [1,2]$. This is analogous to \eqref{eqn:definetti}. To align the support of the $\Betadist$ law with that of $\pi$, we require
\begin{align*}
    \theta - 1 &= \frac{\kappa - 1}{\kappa + \lambda - 2} \\
    (\theta - 1) (\kappa + \lambda - 2) &= \kappa - 1 \\
    \theta (\lambda - 2) + 3 - \lambda &= (2 - \theta) \kappa \\
    \frac{\theta (\lambda - 2) - (\lambda - 3)}{2 - \theta} &= \kappa > 1.
\end{align*}
Now, $\lambda > 1$ holds automatically such that we can set arbitrary $\lambda > 1$ and parameterize $\kappa = \frac{\theta (\lambda - 2) - (\lambda - 3)}{2 - \theta}$. For straightforward implementation, we choose $\lambda = 2$ such that $\kappa = \frac{1}{2 - \theta}$. To conclude, the initial value problem as well as the Lotka-Volterra parameterization depend on the latent $\theta$. \looseness=-1

%% file: chapters/appendix-cyclic-paper.tex
\subsection{Training details}
We sample $n = 20$ distinct time steps $t_m \sim \Uniform \left( [0,0.5] \right), m \in [n]$. In addition, we parameterize the Lotka-Volterra bounds defined above with $T = 1$. We again consider $E$ independent sequences at once by stacking independent one-dimensional SDEs. Within one example, all SDEs are evaluated at the same distinct time steps. At any $i$, we then provide the observational sequence as
\begin{align*}
    \left( \mathbf{x}_{t_1},\mathbf{y}_{t_1},...,\mathbf{x}_{t_{n}},\mathbf{y}_{t_{n}},\mathbf{z},\xiCFdot{t_1},\yiCFdot{t_1}\right)
\end{align*}
for $\mathbf{z}$ a counterfactual delimiter token, identical across $i$. Given the query, we ask the model to predict the \textit{full} completion of the counterfactual sequence, $\comp_i := (\xiCFdot{t_2},\yiCFdot{t_2},...,\xiCFdot{t_{n}},\yiCFdot{t_{n}})$, autoregressively. We again evaluate the model on the per-batch $\MSE$ of the predicted completion, $\widehat{\comp_{[B]}}$, and the underlying ground truth sequence, $\comp_{[B]}$,
\begin{align*}
    \MSE \left(\widehat{\comp_{[B]}}, \comp_{[B]} \right) = \frac{1}{B \cdot E} \sum_{b = 1}^B \left\|\widehat{\comp_{b}} - \comp_{b}\right\|_2^2
\end{align*}
and evaluate on an unseen test set following~\eqref{eqn:sdedataset}.

\input{chapters/appendix-cyclic-training-details}

\subsection{Attention behavior}
\begin{figure}[h]
    \begin{subfigure}[b]{0.49\linewidth}
         \centering
         \includegraphics[width=\textwidth]{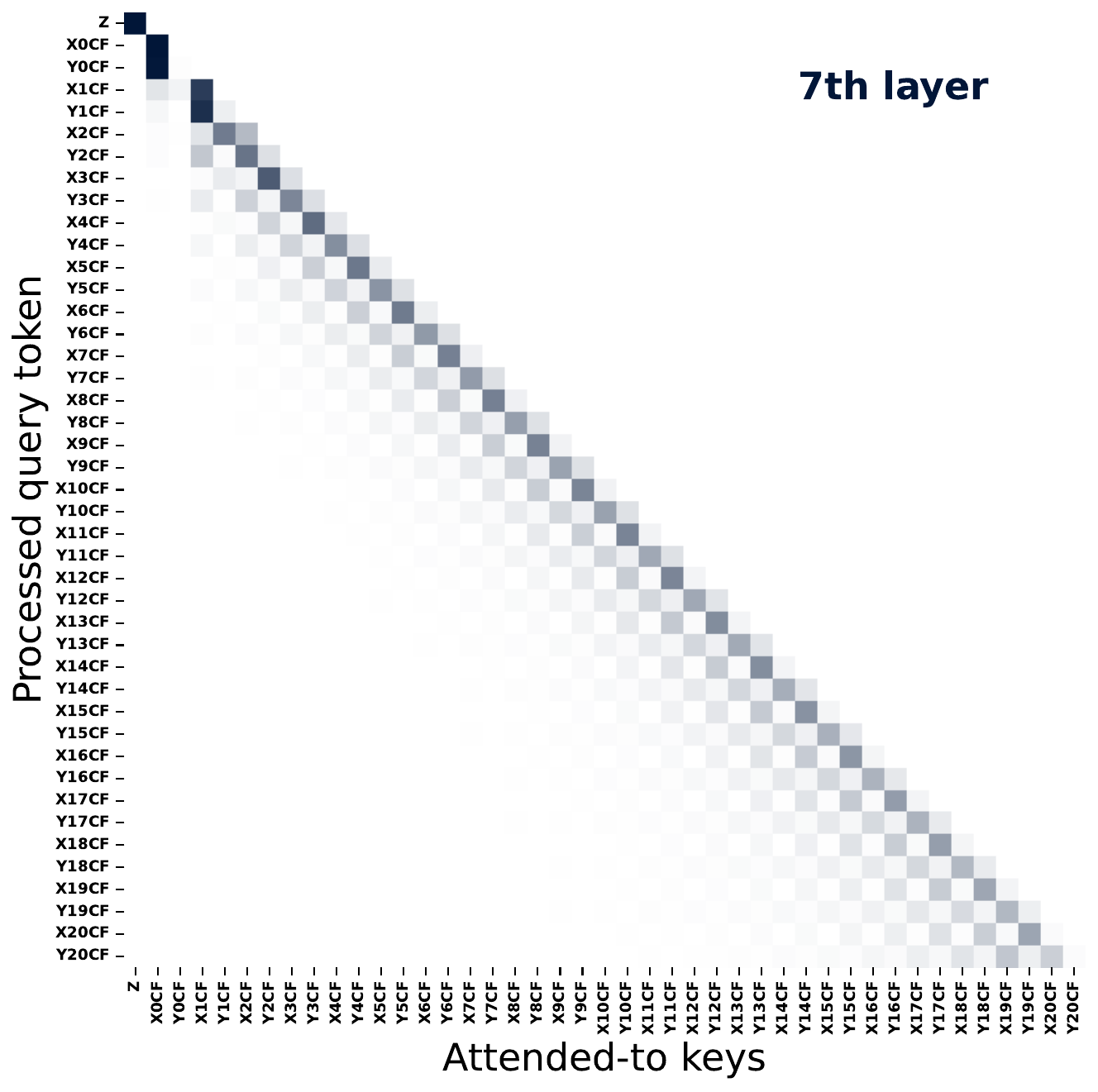}
    \end{subfigure}
    \hfill
    \begin{subfigure}[b]{0.49\linewidth}
         \centering
         \includegraphics[width=\textwidth]{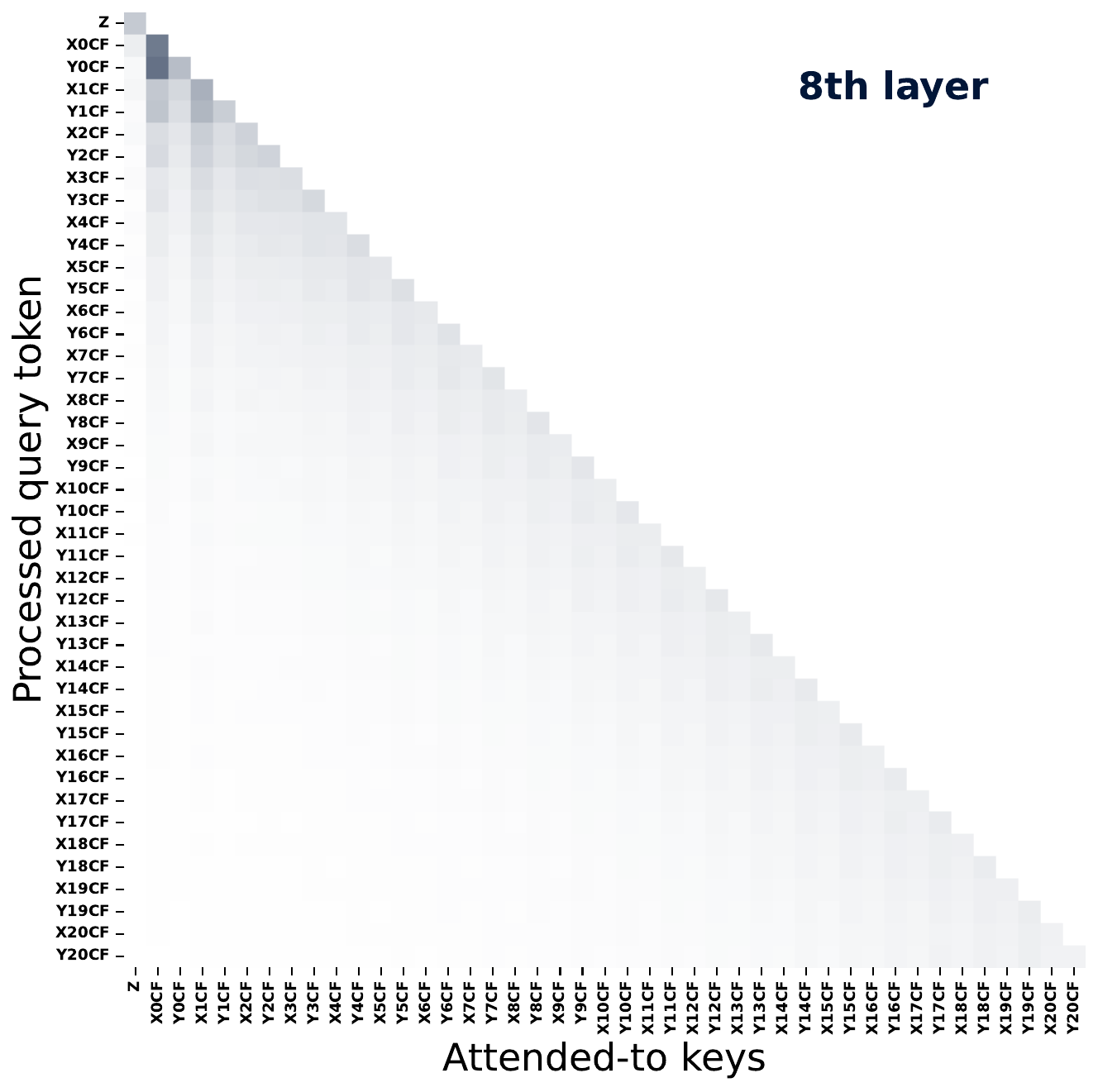}
    \end{subfigure}
    \hfill
    \begin{subfigure}[b]{\linewidth}
         \centering
         \includegraphics[width=\textwidth]{visualizations/attention_maps/custom_colorbar.pdf}
    \end{subfigure}
    \caption{\textbf{Cyclic causal relationship.} The $8$-layer \textbf{AO} Transformer, trained on counterfactual reasoning in Lotka-Volterra SDEs, includes two dedicated copying heads. At layers $5$ and $7$, induction heads shift residual stream information from the previous token forward one position. Earlier layers $1, 2$ and $4$ focus on the initial condition $\xiCFdot{0}$ and implement decaying attention over prior positions.}
    \label{fig:cyclicattentions}
\end{figure}

\autoref{fig:cyclicattentions} shows that the $8$-layer, $1$-head \textbf{AO} Transformer trained on $N$ equal subintervals performs counterfactual reasoning by repeatedly moving information forward. Averaging over $8$ test sequences, we observe that the tokens of the counterfactual sequence attend to the position of the delimiter token $\mathbf{z}$ and the initial condition $\xiCFdot{0}$ at the first layer. This is intensified by layers $2$ and $4$ with attention to the initial $\xiCFdot{0}$. Note that we only plot attention values between tokens of the counterfactual sequence. The Transformer at the fifth layer implements an induction head~\citep{olsson2022incontextlearninginductionheads} that copies forward the information one step. Afterwards, attention head $6$ implements decaying attention over prior positions. Maximum weight is here put on the current token, and progressively lower weights on all previous tokens relative to token distance. This design facilitates forward information propagation across tokens. Similar behavior can be observed for the first two layers with attention to tokens $\{\xiCFdot{0}, ..., \xiCFdot{n}\}$. \looseness=-1

In accordance with the hypothesis that induction heads drive in-context learning~\citep{elhage2021transformercircuits,olsson2022incontextlearninginductionheads}, we notice this pattern for larger models, with multiple heads per layer, and including the $\MLP$. For instance, the \textbf{AO} \textsc{standard} Transformer contains five heads across the first six layers which mostly attend to the delimiter token $\mathbf{z}$. At the eighth and ninth layer, the model implements each one induction head and one 2-gram~\citep{akyürek2024incontextlanguagelearningarchitectures} head in parallel. For the \textbf{Full} \textsc{standard} Transformer, we similarly observe five copying heads which move information forward at layers $8$ and $9$. When trained on variable sequence length, the $8$-layer, $1$-head \textbf{AO} Transformer also exhibits this pattern with an induction head at layer $7$. Taken together, this provides more evidence that counterfactual reasoning emerges, at least partly, in the self-attention layer of the Transformer. \looseness=-1

\begin{figure}[h]
    \centering
    \begin{subfigure}[b]{0.49\linewidth}
         \centering
         \includegraphics[width=\textwidth]{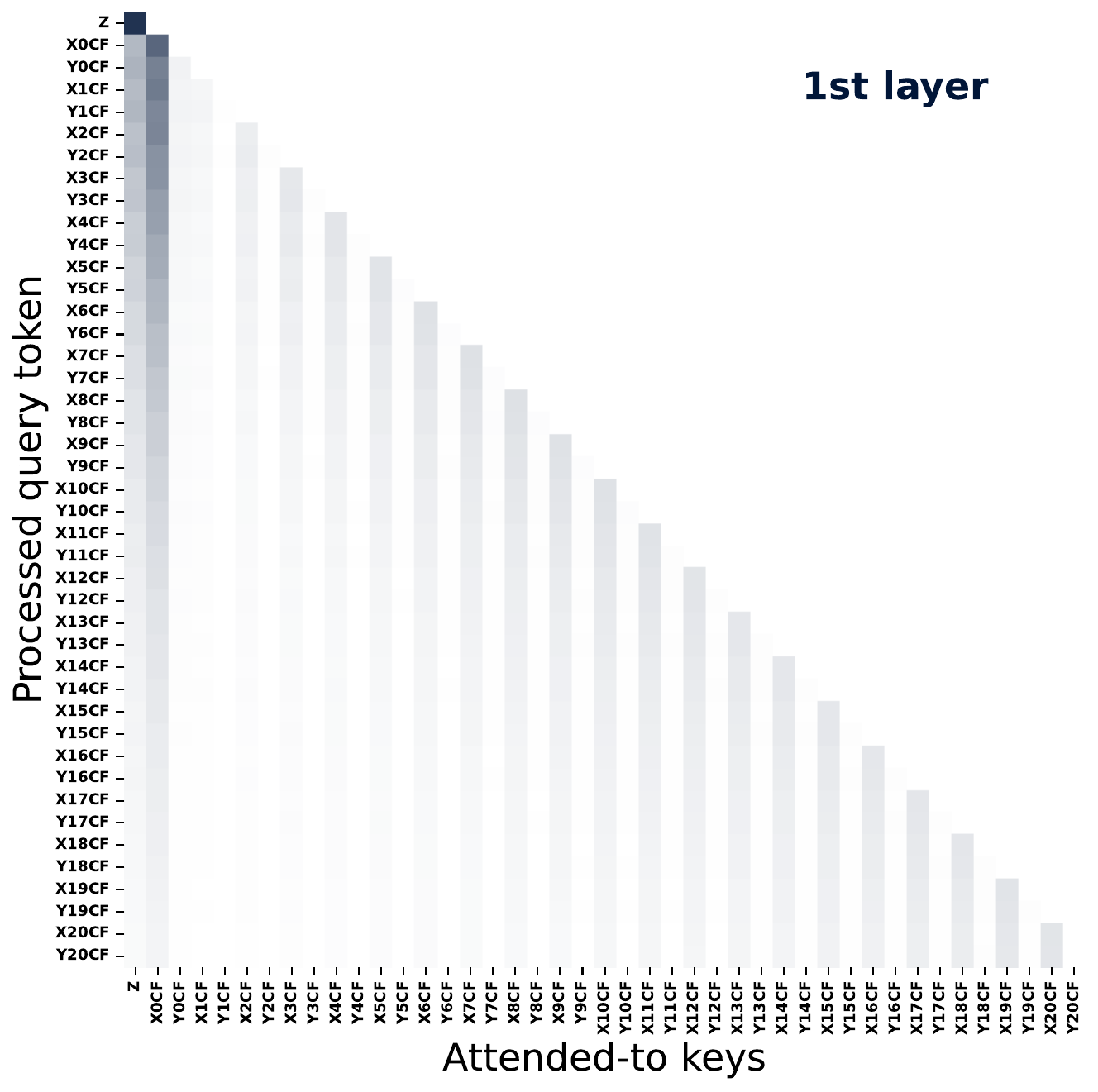}
    \end{subfigure}
    \hfill
    \begin{subfigure}[b]{0.49\linewidth}
         \centering
         \includegraphics[width=\textwidth]{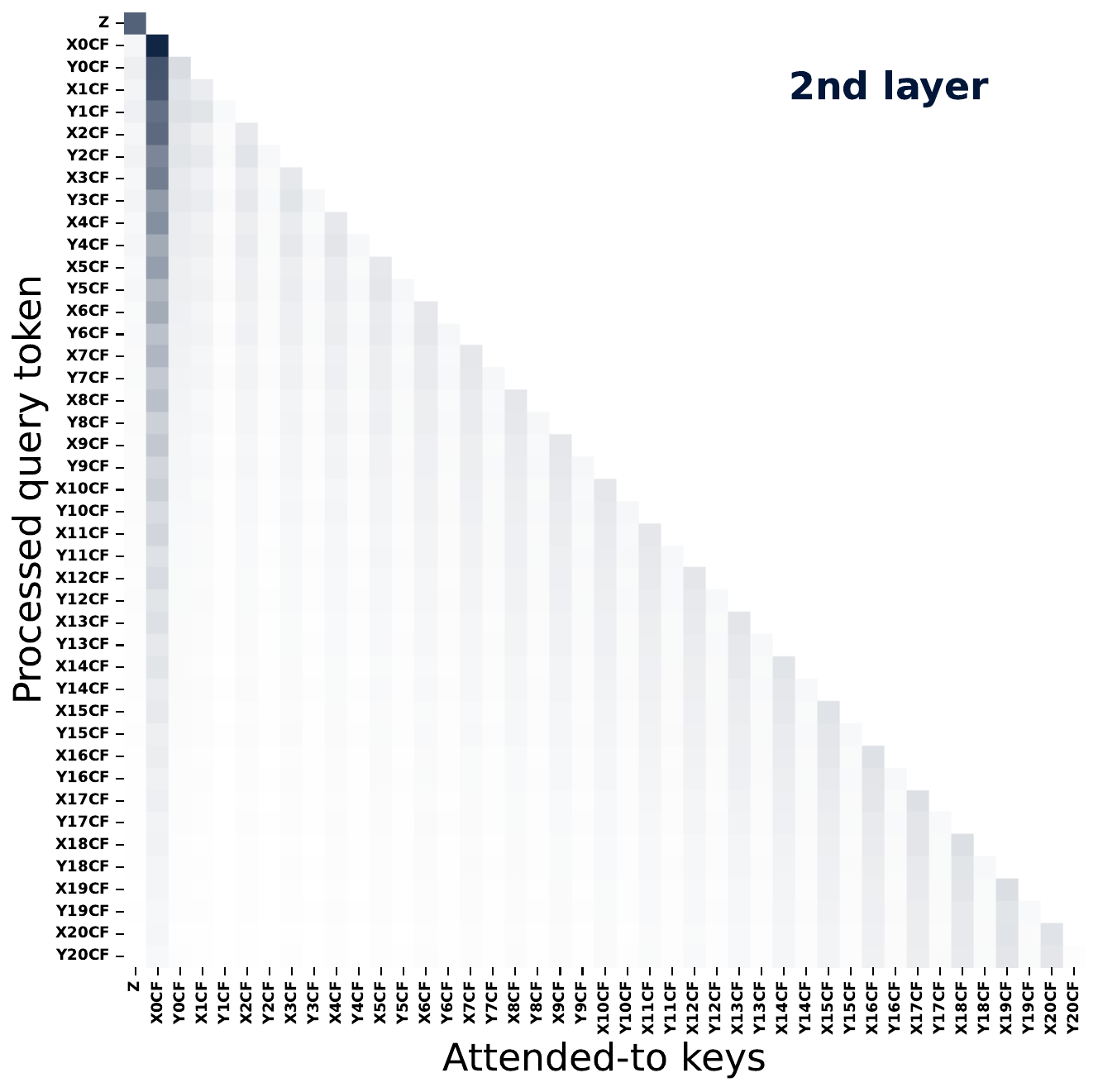}
    \end{subfigure}
    \hfill
    \begin{subfigure}[b]{0.49\linewidth}
         \centering
         \includegraphics[width=\textwidth]{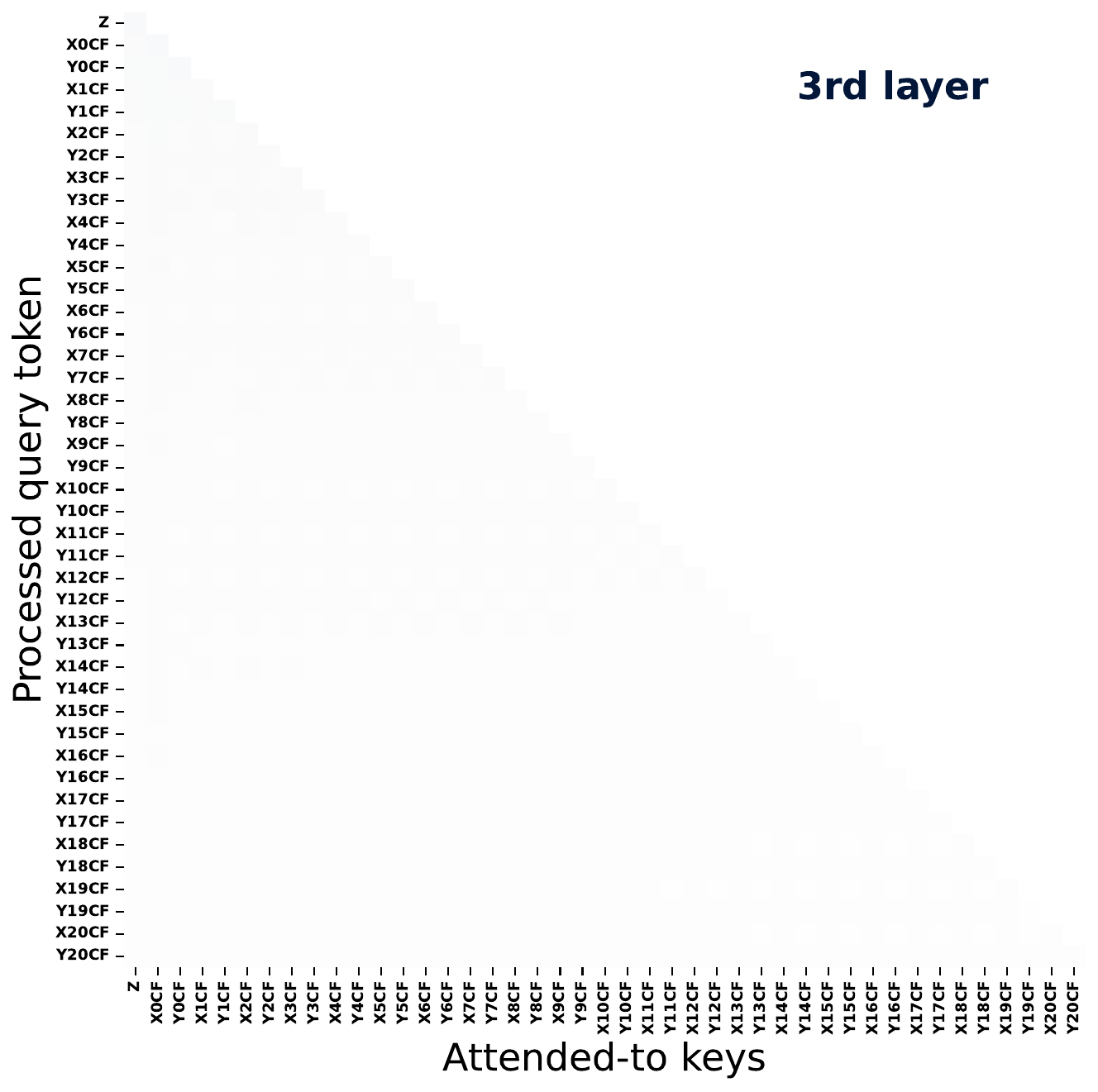}
    \end{subfigure}
    \hfill
    \begin{subfigure}[b]{0.49\linewidth}
         \centering
         \includegraphics[width=\textwidth]{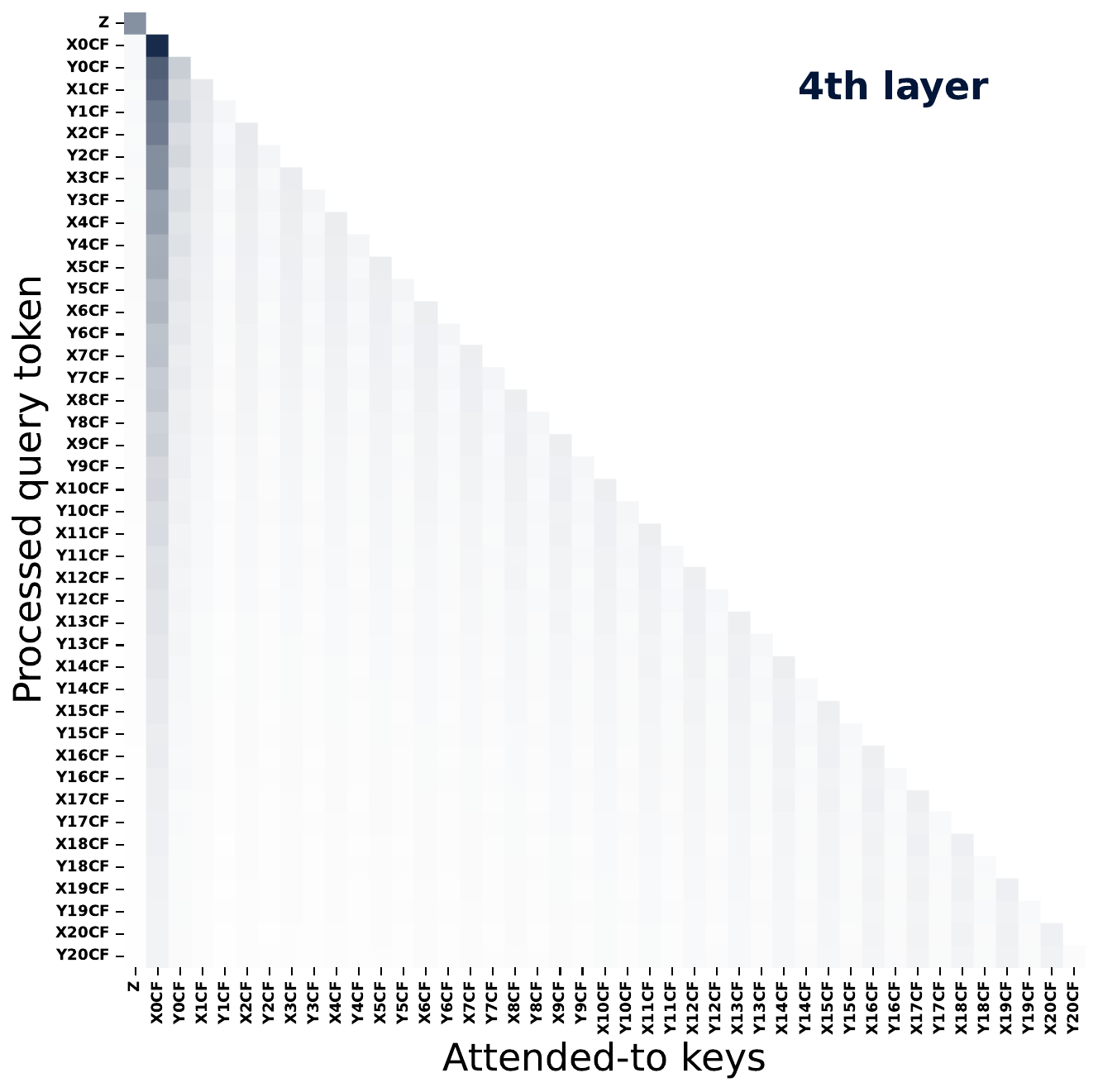}
    \end{subfigure}
    \hfill
    \begin{subfigure}[b]{0.49\linewidth}
         \centering
         \includegraphics[width=\textwidth]{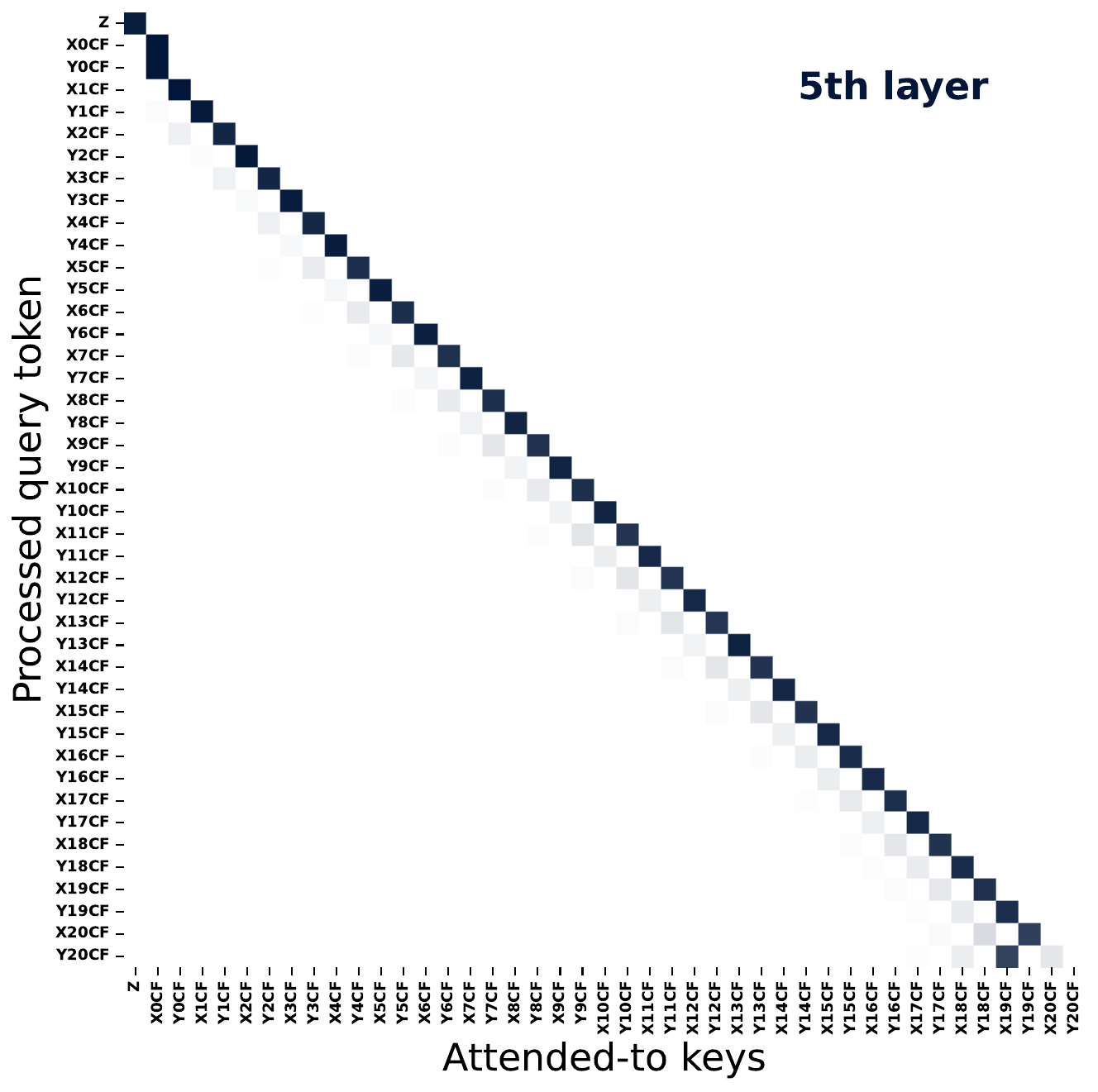}
    \end{subfigure}
    \hfill
    \begin{subfigure}[b]{0.49\linewidth}
         \centering
         \includegraphics[width=\textwidth]{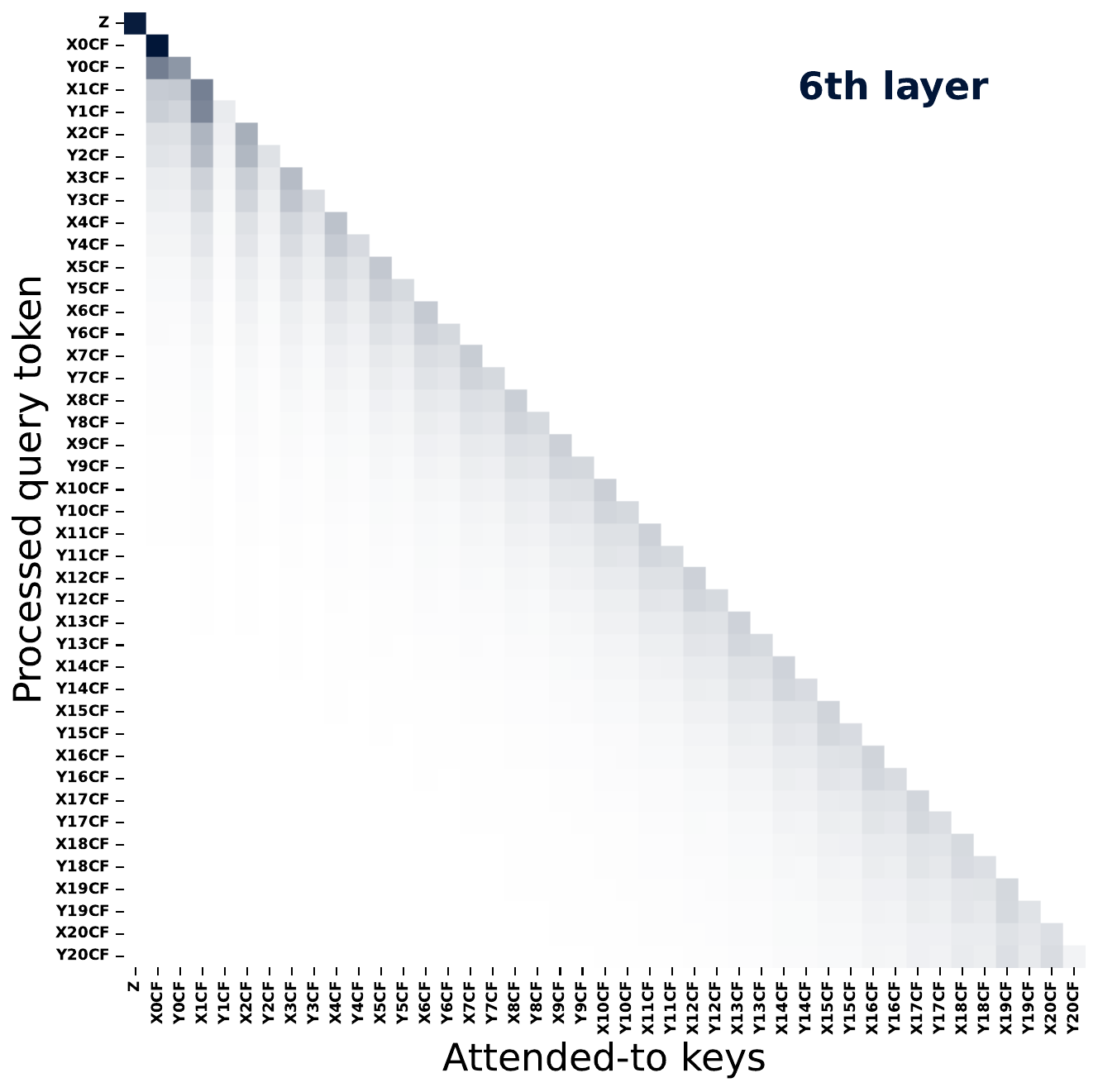}
    \end{subfigure}
\end{figure}

%% file: chapters/appendix-cyclic-training-details.tex
We train for $N = 50'000$ training steps at batch size $B = 8$ and numerically approximate the solutions to the SDEs in~\eqref{eqn:sdes} using the Euler–Maruyama~\citep{maruyama1955euler} scheme. 
The counterfactual sequence is generated analogously but by recycling the Brownian motion used for the observational sequence. 
The manifestation of this can be seen in~\autoref{fig:prediction_lv} for \textcolor{ourmagenta}{\textit{observational}} and \textcolor{ouryellow}{\textit{counterfactual}} $y$.
We inherit the \texttt{torchsde} package from \citet{li2020torchsde} and follow~\citet{lvpython2018readthedocs} for the implementation of the Lotka-Volterra equations. All other specifications are similar to the regression setup described in~\autoref{app:trainingdetails}.
For fast approximation of the SDEs, we generate $20$ batches of size $2500 \cdot 8$ at once. Across one batch, $\{t_1,...,t_n\}$ is constant. At $N = 50'000$ training steps and a resulting dataset sample size of $50'000 \cdot 8$, we hence train on $\frac{50'000}{2500} = 20$ distinct realizations of the time sequence $\{t_1,...,t_n\}$. As this can be seen as curriculum learning, we do not fully adhere to the aforementioned notion of exchangeability. Due to the complexity of approximating SDEs, however, we give priority to an efficient implementation and acknowledge the resulting shortcoming.
Last, the Euler-Maruyama method is an SDE approximation on $N$ \textit{equal} subintervals of width $\Delta t > 0$. Therefore, we also train models with equidistant time steps $\{0, t_1, ..., t_n\}$ for constant $n = 20$. For this setup, the observational sequence starts at $t_0 = 0$,
\begin{align*}
    \left( \mathbf{x}_{0},\mathbf{y}_{0},...,\mathbf{x}_{t_{n}},\mathbf{y}_{t_{n}},\mathbf{z},\xiCFdot{0},\yiCFdot{0}\right)
\end{align*}
with completion $\comp := (\xiCFdot{t_1},\yiCFdot{t_1},...,\xiCFdot{t_{n}},\yiCFdot{t_{n}})$ and evaluation on $n \cdot 2$ tokens. Throughout our experiments, we state the differences between these two setups.

%% file: chapters/appendix-natural-language.tex
\section{Connection between synthetic setup and natural language}
\label{app:naturallanguage}
\subsection{Discrete regression setup}
Due to the synthetic nature of our setup, the connection to natural language is not immediately apparent. To illustrate how unobserved noise $ u $ can be interpreted linguistically, we return to the introductory example. Each pair $(x_j, y_j)$ can be thought of as a pair of sentences describing a patient's treatment and effect. For $j \in [n]$, $x_j$ captures the treatment patient $j$ receives, and $y_j$ provides information on the effect of that medication. After presenting the model with $n$ such in-context examples, we prompt it to predict the counterfactual effect had patient $j = z$ received a different treatment. The model therefore needs to disentangle the subject-specific noise $u_j$ from the observational pair $(x_j, y_j)$. \autoref{fig:mainfigure} visualizes this process in natural language. 

In addition to the one-prompt understanding, we can interpret the exchangeable setup such that each sequence resembles a different health issue. While we analyze breast cancer above, this enables training on a diverse set of treatment-effect pairs in order to perform inference over an unseen clinical picture. Establishing such a principled machine can assist informed decisions in personalized medicine. \looseness=-1

\subsection{Continuous SDE setup}
The cyclic causal dependencies modeled by the SDEs in~\eqref{eqn:sdes} align with the sequential structure of natural language. Each query can be interpreted as a factual narrative of length $t_N$. Given a hypothetical initial condition $(\xzeroCF, \yzeroCF)$, we task the model with counterfactually completing the story while keeping the noise fixed. This noise may represent semantic variability as well as narrative-internal uncertainty. By doing so, the model can generate plausible counterfactual continuations conditioned on the observed prompt. This approach resonates with the work of~\citet{qin2019counterfactualstory}, who explore how language models can reason about alternative story outcomes through counterfactual perturbations to character actions or events.

%% file: neurips.bbl
\begin{thebibliography}{69}
\providecommand{\natexlab}[1]{#1}
\providecommand{\url}[1]{\texttt{#1}}
\expandafter\ifx\csname urlstyle\endcsname\relax
  \providecommand{\doi}[1]{doi: #1}\else
  \providecommand{\doi}{doi: \begingroup \urlstyle{rm}\Url}\fi

\bibitem[Achiam et~al.(2023)Achiam, Adler, Agarwal, Ahmad, Akkaya, Aleman, Almeida, Altenschmidt, Altman, Anadkat, et~al.]{achiam2023gpt}
J.~Achiam, S.~Adler, S.~Agarwal, L.~Ahmad, I.~Akkaya, F.~L. Aleman, D.~Almeida, J.~Altenschmidt, S.~Altman, S.~Anadkat, et~al.
\newblock Gpt-4 technical report.
\newblock \emph{arXiv preprint arXiv:2303.08774}, 2023.
\newblock URL \url{https://arxiv.org/abs/2303.08774}.

\bibitem[Aky{\"u}rek et~al.(2023)Aky{\"u}rek, Schuurmans, Andreas, Ma, and Zhou]{akyürek2023what}
E.~Aky{\"u}rek, D.~Schuurmans, J.~Andreas, T.~Ma, and D.~Zhou.
\newblock What learning algorithm is in-context learning? investigations with linear models.
\newblock In \emph{The Eleventh International Conference on Learning Representations}, 2023.
\newblock URL \url{https://openreview.net/forum?id=0g0X4H8yN4I}.

\bibitem[Akyürek et~al.(2024)Akyürek, Wang, Kim, and Andreas]{akyürek2024incontextlanguagelearningarchitectures}
E.~Akyürek, B.~Wang, Y.~Kim, and J.~Andreas.
\newblock In-context language learning: Architectures and algorithms.
\newblock In \emph{Forty-first International Conference on Machine Learning}, 2024.
\newblock URL \url{https://openreview.net/forum?id=3Z9CRr5srL}.

\bibitem[Arjovsky et~al.(2016)Arjovsky, Shah, and Bengio]{arjovsky2015unitary}
M.~Arjovsky, A.~Shah, and Y.~Bengio.
\newblock Unitary evolution recurrent neural networks.
\newblock In \emph{Proceedings of the 33rd International Conference on International Conference on Machine Learning - Volume 48}, ICML'16, page 1120–1128. JMLR.org, 2016.

\bibitem[Bottou(2014)]{bottou2011reasoning}
L.~Bottou.
\newblock From machine learning to machine reasoning.
\newblock \emph{Machine Learning}, 94\penalty0 (2):\penalty0 133--149, 2014.
\newblock \doi{10.1007/s10994-013-5335-x}.
\newblock URL \url{https://doi.org/10.1007/s10994-013-5335-x}.

\bibitem[Bottou and Sch{\"o}lkopf(2025)]{bottou2025fiction}
L.~Bottou and B.~Sch{\"o}lkopf.
\newblock The fiction machine, April 2025.
\newblock URL \url{https://www.siam.org/publications/siam-news/articles/the-fiction-machine/}.

\bibitem[Brown et~al.(2020)Brown, Mann, Ryder, Subbiah, Kaplan, Dhariwal, Neelakantan, Shyam, Sastry, Askell, Agarwal, Herbert-Voss, Krueger, Henighan, Child, Ramesh, Ziegler, Wu, Winter, Hesse, Chen, Sigler, Litwin, Gray, Chess, Clark, Berner, McCandlish, Radford, Sutskever, and Amodei]{brown2020fewshot}
T.~Brown, B.~Mann, N.~Ryder, M.~Subbiah, J.~D. Kaplan, P.~Dhariwal, A.~Neelakantan, P.~Shyam, G.~Sastry, A.~Askell, S.~Agarwal, A.~Herbert-Voss, G.~Krueger, T.~Henighan, R.~Child, A.~Ramesh, D.~Ziegler, J.~Wu, C.~Winter, C.~Hesse, M.~Chen, E.~Sigler, M.~Litwin, S.~Gray, B.~Chess, J.~Clark, C.~Berner, S.~McCandlish, A.~Radford, I.~Sutskever, and D.~Amodei.
\newblock Language models are few-shot learners.
\newblock In H.~Larochelle, M.~Ranzato, R.~Hadsell, M.~Balcan, and H.~Lin, editors, \emph{Advances in Neural Information Processing Systems}, volume~33, pages 1877--1901. Curran Associates, Inc., 2020.
\newblock URL \url{https://proceedings.neurips.cc/paper_files/paper/2020/file/1457c0d6bfcb4967418bfb8ac142f64a-Paper.pdf}.

\bibitem[Charleux et~al.(2018)Charleux, Roux, Goyallon, Feverati, and Nagorny]{lvpython2018readthedocs}
L.~Charleux, E.~Roux, T.~Goyallon, G.~Feverati, and P.~Nagorny.
\newblock Lotka-volterra equations, 2018.
\newblock URL \url{https://scientific-python.readthedocs.io/en/latest/notebooks_rst/3_Ordinary_Differential_Equations/02_Examples/Lotka_Volterra_model.html}.

\bibitem[Cho et~al.(2014)Cho, van Merri{\"e}nboer, Gulcehre, Bahdanau, Bougares, Schwenk, and Bengio]{cho2014learning}
K.~Cho, B.~van Merri{\"e}nboer, C.~Gulcehre, D.~Bahdanau, F.~Bougares, H.~Schwenk, and Y.~Bengio.
\newblock Learning phrase representations using {RNN} encoder{--}decoder for statistical machine translation.
\newblock In A.~Moschitti, B.~Pang, and W.~Daelemans, editors, \emph{Proceedings of the 2014 Conference on Empirical Methods in Natural Language Processing ({EMNLP})}, pages 1724--1734, Doha, Qatar, Oct. 2014. Association for Computational Linguistics.
\newblock \doi{10.3115/v1/D14-1179}.
\newblock URL \url{https://aclanthology.org/D14-1179/}.

\bibitem[Dai et~al.(2023)Dai, Sun, Dong, Hao, Ma, Sui, and Wei]{dai2023why}
D.~Dai, Y.~Sun, L.~Dong, Y.~Hao, S.~Ma, Z.~Sui, and F.~Wei.
\newblock Why can {GPT} learn in-context? language models implicitly perform gradient descent as meta-optimizers.
\newblock In \emph{ICLR 2023 Workshop on Mathematical and Empirical Understanding of Foundation Models}, 2023.
\newblock URL \url{https://openreview.net/forum?id=fzbHRjAd8U}.

\bibitem[de~Finetti(1931)]{deFinetti1931FunzioneAleatorio}
B.~de~Finetti.
\newblock {Funzione caratteristica di un fenomeno aleatorio.}
\newblock \emph{Atti della R. Academia Nazionale dei Lincei, Serie 6. Memorie, Classe di Scienze Fisiche, Mathematice e Naturale}, 4:\penalty0 251--299, 1931.
\newblock URL \url{http://www.brunodefinetti.it/Opere/funzioneCaratteristica.pdf}.

\bibitem[Deutch et~al.(2024)Deutch, Magar, Natan, and Dar]{deutch-etal-2024-context}
G.~Deutch, N.~Magar, T.~Natan, and G.~Dar.
\newblock In-context learning and gradient descent revisited.
\newblock In K.~Duh, H.~Gomez, and S.~Bethard, editors, \emph{Proceedings of the 2024 Conference of the North American Chapter of the Association for Computational Linguistics: Human Language Technologies (Volume 1: Long Papers)}, pages 1017--1028, Mexico City, Mexico, June 2024. Association for Computational Linguistics.
\newblock \doi{10.18653/v1/2024.naacl-long.58}.
\newblock URL \url{https://aclanthology.org/2024.naacl-long.58/}.

\bibitem[Efron(1979)]{efron1979basicbootstrap}
B.~Efron.
\newblock Bootstrap methods: Another look at the jackknife.
\newblock \emph{The Annals of Statistics}, 7\penalty0 (1):\penalty0 1--26, 1979.
\newblock ISSN 00905364, 21688966.
\newblock URL \url{http://www.jstor.org/stable/2958830}.

\bibitem[Elhage et~al.(2021)Elhage, Nanda, Olsson, Henighan, Joseph, Mann, Askell, Bai, Chen, Conerly, DasSarma, Drain, Ganguli, Hatfield-Dodds, Hernandez, Jones, Kernion, Lovitt, Ndousse, Amodei, Brown, Clark, Kaplan, McCandlish, and Olah]{elhage2021transformercircuits}
N.~Elhage, N.~Nanda, C.~Olsson, T.~Henighan, N.~Joseph, B.~Mann, A.~Askell, Y.~Bai, A.~Chen, T.~Conerly, N.~DasSarma, D.~Drain, D.~Ganguli, Z.~Hatfield-Dodds, D.~Hernandez, A.~Jones, J.~Kernion, L.~Lovitt, K.~Ndousse, D.~Amodei, T.~Brown, J.~Clark, J.~Kaplan, S.~McCandlish, and C.~Olah.
\newblock A mathematical framework for transformer circuits.
\newblock \emph{Transformer Circuits Thread}, 2021.
\newblock URL \url{https://transformer-circuits.pub/2021/framework/index.html}.

\bibitem[Elman(1990)]{elman1990finding}
J.~L. Elman.
\newblock Finding structure in time.
\newblock \emph{Cognitive Science}, 14\penalty0 (2):\penalty0 179--211, 1990.
\newblock URL \url{https://onlinelibrary.wiley.com/doi/abs/10.1207/s15516709cog1402_1}.

\bibitem[Falck et~al.(2024)Falck, Wang, and Holmes]{pmlr-v235-falck24a}
F.~Falck, Z.~Wang, and C.~C. Holmes.
\newblock Is in-context learning in large language models bayesian? {A} martingale perspective.
\newblock In R.~Salakhutdinov, Z.~Kolter, K.~Heller, A.~Weller, N.~Oliver, J.~Scarlett, and F.~Berkenkamp, editors, \emph{Proceedings of the 41st International Conference on Machine Learning}, volume 235 of \emph{Proceedings of Machine Learning Research}, pages 12784--12805. PMLR, 21--27 Jul 2024.
\newblock URL \url{https://proceedings.mlr.press/v235/falck24a.html}.

\bibitem[Feng et~al.(2024)Feng, Tung, Ahmed, Bengio, and Hajimirsadeghi]{feng2024rnnsneeded}
L.~Feng, F.~Tung, M.~O. Ahmed, Y.~Bengio, and H.~Hajimirsadeghi.
\newblock Were rnns all we needed?, 2024.
\newblock URL \url{https://arxiv.org/abs/2410.01201}.

\bibitem[Garg et~al.(2022)Garg, Tsipras, Liang, and Valiant]{linreg}
S.~Garg, D.~Tsipras, P.~Liang, and G.~Valiant.
\newblock What can transformers learn in-context? a case study of simple function classes.
\newblock In \emph{Proceedings of the 36th International Conference on Neural Information Processing Systems}, NIPS '22, Red Hook, NY, USA, 2022. Curran Associates Inc.
\newblock ISBN 9781713871088.
\newblock URL \url{https://openreview.net/pdf?id=flNZJ2eOet}.

\bibitem[Grendar(2006)]{grendar2006effectivesupportsize}
M.~Grendar.
\newblock Entropy and effective support size.
\newblock \emph{Entropy}, 8\penalty0 (3):\penalty0 169--174, 2006.
\newblock ISSN 1099-4300.
\newblock \doi{10.3390/e8030169}.
\newblock URL \url{https://www.mdpi.com/1099-4300/8/3/169}.

\bibitem[Gu and Dao(2024)]{gu2024mambalineartimesequencemodeling}
A.~Gu and T.~Dao.
\newblock Mamba: Linear-time sequence modeling with selective state spaces, 2024.
\newblock URL \url{https://arxiv.org/abs/2312.00752}.

\bibitem[Guo et~al.(2023)Guo, Wildberger, and Sch{\"o}lkopf]{guoout}
S.~Guo, J.~B. Wildberger, and B.~Sch{\"o}lkopf.
\newblock Out-of-variable generalisation for discriminative models.
\newblock In \emph{The Twelfth International Conference on Learning Representations}, 2023.
\newblock URL \url{https://openreview.net/pdf?id=zwMfg9PfPs}.

\bibitem[Guo et~al.(2024{\natexlab{a}})Guo, Tóth, Schölkopf, and Huszár]{causaldefinetti}
S.~Guo, V.~Tóth, B.~Schölkopf, and F.~Huszár.
\newblock Causal de finetti: On the identification of invariant causal structure in exchangeable data, 2024{\natexlab{a}}.
\newblock URL \url{https://arxiv.org/abs/2203.15756}.

\bibitem[Guo et~al.(2024{\natexlab{b}})Guo, Zhang, Mohan, Huszár, and Schölkopf]{dofinetti}
S.~Guo, C.~Zhang, K.~Mohan, F.~Huszár, and B.~Schölkopf.
\newblock Do finetti: On causal effects for exchangeable data, 2024{\natexlab{b}}.
\newblock URL \url{https://arxiv.org/abs/2405.18836}.

\bibitem[Henaff et~al.(2016)Henaff, Szlam, and LeCun]{henaff2016recurrent}
M.~Henaff, A.~Szlam, and Y.~LeCun.
\newblock Recurrent orthogonal networks and long-memory tasks.
\newblock In M.~F. Balcan and K.~Q. Weinberger, editors, \emph{Proceedings of The 33rd International Conference on Machine Learning}, volume~48 of \emph{Proceedings of Machine Learning Research}, pages 2034--2042, New York, New York, USA, 20--22 Jun 2016. PMLR.
\newblock URL \url{https://proceedings.mlr.press/v48/henaff16.html}.

\bibitem[Hernan(2024)]{Hernan2024-HERCIW}
M.~A. Hernan.
\newblock \emph{Causal Inference: What If}.
\newblock Taylor \& Francis, Boca Raton, 2024.
\newblock URL \url{https://miguelhernan.org/whatifbook}.

\bibitem[Hochreiter and Schmidhuber(1997)]{hochreiter1997lstm}
S.~Hochreiter and J.~Schmidhuber.
\newblock Long short-term memory.
\newblock \emph{Neural Computation}, 9\penalty0 (8):\penalty0 1735--1780, 1997.
\newblock URL \url{https://www.bioinf.jku.at/publications/older/2604.pdf}.

\bibitem[Kingma and Ba(2015)]{kingma2015adam}
D.~P. Kingma and J.~Ba.
\newblock Adam: {A} method for stochastic optimization.
\newblock In Y.~Bengio and Y.~LeCun, editors, \emph{3rd International Conference on Learning Representations, {ICLR} 2015, San Diego, CA, USA, May 7-9, 2015, Conference Track Proceedings}, 2015.
\newblock URL \url{http://arxiv.org/abs/1412.6980}.

\bibitem[Klenke(2008)]{klenke2008}
A.~Klenke.
\newblock \emph{Probability Theory: A Comprehensive Course}.
\newblock Springer, 2008.
\newblock URL \url{https://link.springer.com/book/10.1007/978-1-84800-048-3}.

\bibitem[Kusner et~al.(2017)Kusner, Loftus, Russell, and Silva]{kusner2017counterfactual}
M.~J. Kusner, J.~Loftus, C.~Russell, and R.~Silva.
\newblock Counterfactual fairness.
\newblock \emph{Advances in neural information processing systems}, 30, 2017.
\newblock URL \url{https://proceedings.neurips.cc/paper_files/paper/2017/file/a486cd07e4ac3d270571622f4f316ec5-Paper.pdf}.

\bibitem[Li et~al.(2023)Li, Yu, and Ettinger]{li2023counterfactualreasoning}
J.~Li, L.~Yu, and A.~Ettinger.
\newblock Counterfactual reasoning: Testing language models' understanding of hypothetical scenarios.
\newblock In A.~Rogers, J.~Boyd-Graber, and N.~Okazaki, editors, \emph{Proceedings of the 61st Annual Meeting of the Association for Computational Linguistics (Volume 2: Short Papers)}, pages 804--815, Toronto, Canada, July 2023. Association for Computational Linguistics.
\newblock \doi{10.18653/v1/2023.acl-short.70}.
\newblock URL \url{https://aclanthology.org/2023.acl-short.70/}.

\bibitem[Li et~al.(2020)Li, Wong, Chen, and Duvenaud]{li2020torchsde}
X.~Li, T.-K.~L. Wong, R.~T.~Q. Chen, and D.~Duvenaud.
\newblock Scalable gradients for stochastic differential equations.
\newblock \emph{International Conference on Artificial Intelligence and Statistics}, 2020.
\newblock URL \url{https://proceedings.mlr.press/v108/li20i/li20i.pdf}.

\bibitem[Lorch et~al.(2024)Lorch, Krause, and Sch\"{o}lkopf]{lorch2024sde}
L.~Lorch, A.~Krause, and B.~Sch\"{o}lkopf.
\newblock Causal modeling with stationary diffusions.
\newblock In S.~Dasgupta, S.~Mandt, and Y.~Li, editors, \emph{Proceedings of The 27th International Conference on Artificial Intelligence and Statistics}, volume 238 of \emph{Proceedings of Machine Learning Research}, pages 1927--1935. PMLR, 02--04 May 2024.
\newblock URL \url{https://proceedings.mlr.press/v238/lorch24a.html}.

\bibitem[Lorenz(1973)]{lorenz1973spiegel}
K.~Lorenz.
\newblock \emph{Die R{\"u}ckseite des Spiegels : Versuch einer Naturgeschichte menschlichen Erkennens}.
\newblock Piper, M{\"u}nchen [u.a, 2. aufl. edition, 1973.
\newblock ISBN 3492020305.

\bibitem[Loshchilov and Hutter(2019)]{loshchilov2018decoupled}
I.~Loshchilov and F.~Hutter.
\newblock Decoupled weight decay regularization.
\newblock In \emph{International Conference on Learning Representations}, 2019.
\newblock URL \url{https://openreview.net/forum?id=Bkg6RiCqY7}.

\bibitem[Lotka(1910)]{lotka1910volterra}
A.~J. Lotka.
\newblock Contribution to the theory of periodic reactions.
\newblock \emph{The Journal of Physical Chemistry}, 14\penalty0 (3):\penalty0 271--274, 03 1910.
\newblock \doi{10.1021/j150111a004}.
\newblock URL \url{https://doi.org/10.1021/j150111a004}.

\bibitem[Lu et~al.(2020)Lu, Huang, Wang, Hernández-Lobato, Zhang, and Sch{\"o}lkopf]{lu2020sample}
C.~Lu, B.~Huang, K.~Wang, J.~M. Hernández-Lobato, K.~Zhang, and B.~Sch{\"o}lkopf.
\newblock Sample-efficient reinforcement learning via counterfactual-based data augmentation.
\newblock In \emph{Offline Reinforcement Learning - Workshop at the 34th Conference on Neural Information Processing Systems (NeurIPS)}, 2020.
\newblock URL \url{https://offline-rl-neurips.github.io/pdf/34.pdf}.

\bibitem[Maruyama(1955)]{maruyama1955euler}
G.~Maruyama.
\newblock Continuous markov processes and stochastic equations.
\newblock \emph{Rendiconti del Circolo Matematico di Palermo}, 4\penalty0 (1):\penalty0 48--90, 1955.
\newblock \doi{10.1007/BF02846028}.
\newblock URL \url{https://doi.org/10.1007/BF02846028}.

\bibitem[Mesnard et~al.(2021)Mesnard, Weber, Viola, Thakoor, Saade, Harutyunyan, Dabney, Stepleton, Heess, Guez, Moulines, Hutter, Buesing, and Munos]{mesnard2020counterfactual}
T.~Mesnard, T.~Weber, F.~Viola, S.~Thakoor, A.~Saade, A.~Harutyunyan, W.~Dabney, T.~S. Stepleton, N.~Heess, A.~Guez, E.~Moulines, M.~Hutter, L.~Buesing, and R.~Munos.
\newblock Counterfactual credit assignment in model-free reinforcement learning.
\newblock In M.~Meila and T.~Zhang, editors, \emph{Proceedings of the 38th International Conference on Machine Learning}, volume 139 of \emph{Proceedings of Machine Learning Research}, pages 7654--7664. PMLR, 18--24 Jul 2021.
\newblock URL \url{https://proceedings.mlr.press/v139/mesnard21a.html}.

\bibitem[Mondorf and Plank(2024)]{mondorf2024beyond}
P.~Mondorf and B.~Plank.
\newblock Beyond accuracy: Evaluating the reasoning behavior of large language models - a survey.
\newblock In \emph{First Conference on Language Modeling}, 2024.
\newblock URL \url{https://openreview.net/forum?id=Lmjgl2n11u}.

\bibitem[Mooij et~al.(2013)Mooij, Janzing, and Sch{\"o}lkopf]{MooJanSch13}
J.~Mooij, D.~Janzing, and B.~Sch{\"o}lkopf.
\newblock From ordinary differential equations to structural causal models: the deterministic case.
\newblock In A.~Nicholson and P.~Smyth, editors, \emph{Proceedings of the Twenty-Ninth Conference Annual Conference on Uncertainty in Artificial Intelligence}, pages 440--448, Corvallis, OR, 2013. AUAI Press.
\newblock URL \url{http://www.is.tuebingen.mpg.de/fileadmin/user_upload/files/publications/2013/MooijJS2013-uai.pdf}.

\bibitem[Nanda et~al.(2023)Nanda, Chan, Lieberum, Smith, and Steinhardt]{nanda2023progress}
N.~Nanda, L.~Chan, T.~Lieberum, J.~Smith, and J.~Steinhardt.
\newblock Progress measures for grokking via mechanistic interpretability.
\newblock In \emph{The Eleventh International Conference on Learning Representations}, 2023.
\newblock URL \url{https://openreview.net/forum?id=9XFSbDPmdW}.

\bibitem[Nasr-Esfahany et~al.(2023)Nasr-Esfahany, Alizadeh, and Shah]{nasr2023counterfactualidentifiability}
A.~Nasr-Esfahany, M.~Alizadeh, and D.~Shah.
\newblock Counterfactual identifiability of bijective causal models.
\newblock In A.~Krause, E.~Brunskill, K.~Cho, B.~Engelhardt, S.~Sabato, and J.~Scarlett, editors, \emph{Proceedings of the 40th International Conference on Machine Learning}, volume 202 of \emph{Proceedings of Machine Learning Research}, pages 25733--25754. PMLR, 23--29 Jul 2023.
\newblock URL \url{https://proceedings.mlr.press/v202/nasr-esfahany23a.html}.

\bibitem[Olsson et~al.(2022)Olsson, Elhage, Nanda, Joseph, DasSarma, Henighan, Mann, Askell, Bai, Chen, Conerly, Drain, Ganguli, Hatfield-Dodds, Hernandez, Johnston, Jones, Kernion, Lovitt, Ndousse, Amodei, Brown, Clark, Kaplan, McCandlish, and Olah]{olsson2022incontextlearninginductionheads}
C.~Olsson, N.~Elhage, N.~Nanda, N.~Joseph, N.~DasSarma, T.~Henighan, B.~Mann, A.~Askell, Y.~Bai, A.~Chen, T.~Conerly, D.~Drain, D.~Ganguli, Z.~Hatfield-Dodds, D.~Hernandez, S.~Johnston, A.~Jones, J.~Kernion, L.~Lovitt, K.~Ndousse, D.~Amodei, T.~Brown, J.~Clark, J.~Kaplan, S.~McCandlish, and C.~Olah.
\newblock In-context learning and induction heads, 2022.
\newblock URL \url{https://arxiv.org/abs/2209.11895}.

\bibitem[Park et~al.(2024)Park, Choe, and Veitch]{park2024lrh}
K.~Park, Y.~J. Choe, and V.~Veitch.
\newblock The linear representation hypothesis and the geometry of large language models.
\newblock In \emph{Proceedings of the 41st International Conference on Machine Learning}, ICML'24. JMLR.org, 2024.

\bibitem[Paszke et~al.(2019)Paszke, Gross, Massa, Lerer, Bradbury, Chanan, Killeen, Lin, Gimelshein, Antiga, Desmaison, K\"{o}pf, Yang, DeVito, Raison, Tejani, Chilamkurthy, Steiner, Fang, Bai, and Chintala]{paszke2019pytorch}
A.~Paszke, S.~Gross, F.~Massa, A.~Lerer, J.~Bradbury, G.~Chanan, T.~Killeen, Z.~Lin, N.~Gimelshein, L.~Antiga, A.~Desmaison, A.~K\"{o}pf, E.~Yang, Z.~DeVito, M.~Raison, A.~Tejani, S.~Chilamkurthy, B.~Steiner, L.~Fang, J.~Bai, and S.~Chintala.
\newblock \emph{PyTorch: an imperative style, high-performance deep learning library}.
\newblock Curran Associates Inc., Red Hook, NY, USA, 2019.
\newblock URL \url{https://proceedings.neurips.cc/paper_files/paper/2019/file/bdbca288fee7f92f2bfa9f7012727740-Paper.pdf}.

\bibitem[Pearl(2009)]{pearl2009}
J.~Pearl.
\newblock \emph{Causality: Models, Reasoning and Inference}.
\newblock Cambridge University Press, 2nd edition, 2009.
\newblock URL \url{https://bayes.cs.ucla.edu/BOOK-2K/}.

\bibitem[Peters et~al.(2011)Peters, Janzing, and Scholkopf]{peters2011causal}
J.~Peters, D.~Janzing, and B.~Scholkopf.
\newblock Causal inference on discrete data using additive noise models.
\newblock \emph{IEEE Transactions on Pattern Analysis and Machine Intelligence}, 33\penalty0 (12):\penalty0 2436--2450, 2011.
\newblock URL \url{https://ieeexplore.ieee.org/document/5740928}.

\bibitem[Peters et~al.(2014)Peters, Mooij, Janzing, and Sch{\"o}lkopf]{peters2014causal}
J.~Peters, J.~M. Mooij, D.~Janzing, and B.~Sch{\"o}lkopf.
\newblock Causal discovery with continuous additive noise models.
\newblock \emph{The Journal of Machine Learning Research}, 15\penalty0 (1):\penalty0 2009--2053, 2014.
\newblock URL \url{https://jmlr.org/papers/volume15/peters14a/peters14a.pdf}.

\bibitem[Peters et~al.(2017)Peters, Janzing, and Schölkopf]{elements}
J.~Peters, D.~Janzing, and B.~Schölkopf.
\newblock \emph{Elements of Causal Inference: Foundations and Learning Algorithms}.
\newblock The MIT Press, 2017.
\newblock ISBN 0262037319.

\bibitem[Peters et~al.(2022)Peters, Bauer, and Pfister]{peters2020dynamicalsystems}
J.~Peters, S.~Bauer, and N.~Pfister.
\newblock \emph{Causal Models for Dynamical Systems}, page 671–690.
\newblock Association for Computing Machinery, New York, NY, USA, 1 edition, 2022.
\newblock ISBN 9781450395861.
\newblock URL \url{https://doi.org/10.1145/3501714.3501752}.

\bibitem[Petty et~al.(2024)Petty, Steenkiste, Dasgupta, Sha, Garrette, and Linzen]{petty2024impact}
J.~Petty, S.~Steenkiste, I.~Dasgupta, F.~Sha, D.~Garrette, and T.~Linzen.
\newblock The impact of depth on compositional generalization in transformer language models.
\newblock In K.~Duh, H.~Gomez, and S.~Bethard, editors, \emph{Proceedings of the 2024 Conference of the North American Chapter of the Association for Computational Linguistics: Human Language Technologies (Volume 1: Long Papers)}, pages 7239--7252, Mexico City, Mexico, June 2024. Association for Computational Linguistics.
\newblock \doi{10.18653/v1/2024.naacl-long.402}.
\newblock URL \url{https://aclanthology.org/2024.naacl-long.402/}.

\bibitem[Qin et~al.(2019)Qin, Bosselut, Holtzman, Bhagavatula, Clark, and Choi]{qin2019counterfactualstory}
L.~Qin, A.~Bosselut, A.~Holtzman, C.~Bhagavatula, E.~Clark, and Y.~Choi.
\newblock Counterfactual story reasoning and generation.
\newblock In K.~Inui, J.~Jiang, V.~Ng, and X.~Wan, editors, \emph{Proceedings of the 2019 Conference on Empirical Methods in Natural Language Processing and the 9th International Joint Conference on Natural Language Processing (EMNLP-IJCNLP)}, pages 5043--5053, Hong Kong, China, Nov. 2019. Association for Computational Linguistics.
\newblock \doi{10.18653/v1/D19-1509}.
\newblock URL \url{https://aclanthology.org/D19-1509/}.

\bibitem[Radford et~al.(2019)Radford, Wu, Child, Luan, Amodei, and Sutskever]{radford2019language}
A.~Radford, J.~Wu, R.~Child, D.~Luan, D.~Amodei, and I.~Sutskever.
\newblock Language models are unsupervised multitask learners.
\newblock \emph{arXiv}, 2019.
\newblock URL \url{https://cdn.openai.com/better-language-models/language_models_are_unsupervised_multitask_learners.pdf}.
\newblock OpenAI Technical Report.

\bibitem[Ravfogel et~al.(2025)Ravfogel, Svete, Sn{\ae}bjarnarson, and Cotterell]{ravfogel2025gumbel}
S.~Ravfogel, A.~Svete, V.~Sn{\ae}bjarnarson, and R.~Cotterell.
\newblock Gumbel counterfactual generation from language models.
\newblock In \emph{The Thirteenth International Conference on Learning Representations}, 2025.
\newblock URL \url{https://openreview.net/forum?id=TUC0ZT2zIQ}.

\bibitem[Reizinger et~al.(2025)Reizinger, Guo, Husz{\'a}r, Sch{\"o}lkopf, and Brendel]{reizinger2024identifiable}
P.~Reizinger, S.~Guo, F.~Husz{\'a}r, B.~Sch{\"o}lkopf, and W.~Brendel.
\newblock Identifiable exchangeable mechanisms for causal structure and representation learning.
\newblock In \emph{The Thirteenth International Conference on Learning Representations}, 2025.
\newblock URL \url{https://openreview.net/forum?id=k03mB41vyM}.

\bibitem[Sauer and Geiger(2021)]{sauer2021counterfactual}
A.~Sauer and A.~Geiger.
\newblock Counterfactual generative networks.
\newblock In \emph{International Conference on Learning Representations}, 2021.
\newblock URL \url{https://openreview.net/forum?id=BXewfAYMmJw}.

\bibitem[Sch{\"o}lkopf(2022)]{scholkopf2022causality}
B.~Sch{\"o}lkopf.
\newblock Causality for machine learning.
\newblock In \emph{Probabilistic and causal inference: The works of Judea Pearl}, pages 765--804. Association for Computing Machinery, New York, NY, United States, 2022.
\newblock URL \url{https://dl.acm.org/doi/10.1145/3501714.3501755}.

\bibitem[Shannon(1948)]{shannon1948entropy}
C.~E. Shannon.
\newblock A mathematical theory of communication.
\newblock \emph{The Bell System Technical Journal}, 27\penalty0 (3):\penalty0 379--423, 1948.
\newblock URL \url{https://people.math.harvard.edu/~ctm/home/text/others/shannon/entropy/entropy.pdf}.

\bibitem[Southgate and Vernetti(2014)]{southgate2014belief}
V.~Southgate and A.~Vernetti.
\newblock Belief-based action prediction in preverbal infants.
\newblock \emph{Cognition}, 130\penalty0 (1):\penalty0 1--10, 2014.
\newblock URL \url{https://www.sciencedirect.com/science/article/pii/S0010027713001650?via%3Dihub}.

\bibitem[Tandon et~al.(2019)Tandon, Dalvi, Sakaguchi, Clark, and Bosselut]{tandon2019wiqa}
N.~Tandon, B.~Dalvi, K.~Sakaguchi, P.~Clark, and A.~Bosselut.
\newblock {WIQA}: A dataset for {\textquotedblleft}what if...{\textquotedblright} reasoning over procedural text.
\newblock In K.~Inui, J.~Jiang, V.~Ng, and X.~Wan, editors, \emph{Proceedings of the 2019 Conference on Empirical Methods in Natural Language Processing and the 9th International Joint Conference on Natural Language Processing (EMNLP-IJCNLP)}, pages 6076--6085, Hong Kong, China, Nov. 2019. Association for Computational Linguistics.
\newblock \doi{10.18653/v1/D19-1629}.
\newblock URL \url{https://aclanthology.org/D19-1629/}.

\bibitem[Vaswani et~al.(2017)Vaswani, Shazeer, Parmar, Uszkoreit, Jones, Gomez, Kaiser, and Polosukhin]{transformer}
A.~Vaswani, N.~Shazeer, N.~Parmar, J.~Uszkoreit, L.~Jones, A.~N. Gomez, L.~Kaiser, and I.~Polosukhin.
\newblock Attention is all you need.
\newblock In \emph{Proceedings of the 31st International Conference on Neural Information Processing Systems}, NIPS'17, page 6000–6010, Red Hook, NY, USA, 2017. Curran Associates Inc.
\newblock ISBN 9781510860964.
\newblock URL \url{https://proceedings.neurips.cc/paper_files/paper/2017/file/3f5ee243547dee91fbd053c1c4a845aa-Paper.pdf}.

\bibitem[von Oswald et~al.(2023)von Oswald, Niklasson, Randazzo, Sacramento, Mordvintsev, Zhmoginov, and Vladymyrov]{vonoswald2023transformerslearn}
J.~von Oswald, E.~Niklasson, E.~Randazzo, J.~a. Sacramento, A.~Mordvintsev, A.~Zhmoginov, and M.~Vladymyrov.
\newblock Transformers learn in-context by gradient descent.
\newblock In \emph{Proceedings of the 40th International Conference on Machine Learning}, ICML'23. JMLR.org, 2023.
\newblock URL \url{https://proceedings.mlr.press/v202/von-oswald23a/von-oswald23a.pdf}.

\bibitem[Wachter et~al.(2017)Wachter, Mittelstadt, and Russell]{wachter2017counterfactual}
S.~Wachter, B.~Mittelstadt, and C.~Russell.
\newblock Counterfactual explanations without opening the black box: Automated decisions and the gdpr.
\newblock \emph{Harv. JL \& Tech.}, 31:\penalty0 841, 2017.
\newblock URL \url{https://arxiv.org/pdf/1711.00399}.

\bibitem[Wei et~al.(2022)Wei, Tay, Bommasani, Raffel, Zoph, Borgeaud, Yogatama, Bosma, Zhou, Metzler, Chi, Hashimoto, Vinyals, Liang, Dean, and Fedus]{wei2022emergent}
J.~Wei, Y.~Tay, R.~Bommasani, C.~Raffel, B.~Zoph, S.~Borgeaud, D.~Yogatama, M.~Bosma, D.~Zhou, D.~Metzler, E.~H. Chi, T.~Hashimoto, O.~Vinyals, P.~Liang, J.~Dean, and W.~Fedus.
\newblock Emergent abilities of large language models.
\newblock \emph{Transactions on Machine Learning Research}, 2022.
\newblock ISSN 2835-8856.
\newblock URL \url{https://openreview.net/forum?id=yzkSU5zdwD}.
\newblock Survey Certification.

\bibitem[Xiao et~al.(2025)Xiao, Zhao, and Huang]{xiao2025rolediversityincontextlearning}
W.~Xiao, H.~Zhao, and L.~Huang.
\newblock The role of diversity in in-context learning for large language models, 2025.
\newblock URL \url{https://arxiv.org/abs/2505.19426}.

\bibitem[Xie et~al.(2022)Xie, Raghunathan, Liang, and Ma]{xie2022explanationincontextlearningimplicit}
S.~M. Xie, A.~Raghunathan, P.~Liang, and T.~Ma.
\newblock An explanation of in-context learning as implicit bayesian inference, 2022.
\newblock URL \url{https://arxiv.org/abs/2111.02080}.

\bibitem[Yan et~al.(2023)Yan, Kong, Gui, Chi, Xing, He, and Zhang]{yan2023counterfactual}
H.~Yan, L.~Kong, L.~Gui, Y.~Chi, E.~Xing, Y.~He, and K.~Zhang.
\newblock Counterfactual generation with identifiability guarantees.
\newblock In \emph{Thirty-seventh Conference on Neural Information Processing Systems}, 2023.
\newblock URL \url{https://openreview.net/forum?id=cslnCXE9XA}.

\bibitem[Ye and Namkoong(2024)]{ye2024exchangeablesequencemodelsquantify}
N.~Ye and H.~Namkoong.
\newblock Exchangeable sequence models quantify uncertainty over latent concepts, 2024.
\newblock URL \url{https://arxiv.org/abs/2408.03307}.

\bibitem[Zhang et~al.(2023)Zhang, McCoy, Sumers, Zhu, and Griffiths]{zhang2023deepfinettirecoveringtopic}
L.~Zhang, R.~T. McCoy, T.~R. Sumers, J.-Q. Zhu, and T.~L. Griffiths.
\newblock Deep de finetti: Recovering topic distributions from large language models, 2023.
\newblock URL \url{https://arxiv.org/abs/2312.14226}.

\end{thebibliography}
